\documentclass{article}

\usepackage{microtype}
\usepackage{graphicx}
\usepackage{booktabs}
\usepackage[T1]{fontenc}
\usepackage[utf8]{inputenc}
\usepackage{tikz}
\usepackage{enumitem}
\usetikzlibrary{positioning, arrows.meta, fit, calc, decorations.pathreplacing}

\usepackage{hyperref}

\usepackage[preprint]{icml2026}

\usepackage{caption}
\usepackage{subcaption}

\usepackage{amsmath}
\usepackage{amssymb}
\usepackage{mathtools}
\usepackage{amsthm}

\usepackage{inconsolata}
\usepackage{tcolorbox}
\usepackage{multirow}
\usepackage{colortbl}
\usepackage{xcolor}
\usepackage{makecell} 
\usepackage{arydshln}

\definecolor{lightgreen}{RGB}{235, 232, 242} 
\definecolor{mediumgreen}{RGB}{198, 224, 198}

\definecolor{frozen}{RGB}{200, 210, 225}
\definecolor{frozenborder}{RGB}{130, 145, 170}
\definecolor{trained}{RGB}{234, 85, 85}
\definecolor{trainedfill}{RGB}{255, 220, 220}
\definecolor{grayline}{RGB}{170, 170, 170}

\usepackage[textsize=tiny]{todonotes}

\icmltitlerunning{Sparse Subnetwork Enhancement for Underrepresented Languages in Large Language Models}

\begin{document}

\twocolumn[
  \icmltitle{Sparse Subnetwork Enhancement for Underrepresented Languages \\ in Large Language Models}



  \icmlsetsymbol{equal}{*}

  \begin{icmlauthorlist}
    \icmlauthor{Daniil Gurgurov}{saar,dfki}
    \icmlauthor{Tanja Bäumel}{saar,dfki,certain}
    \icmlauthor{Josef van Genabith}{saar,dfki}
    \icmlauthor{Simon Ostermann}{saar,dfki,certain}
  \end{icmlauthorlist}

  \icmlaffiliation{saar}{Saarland University}
  \icmlaffiliation{dfki}{German Research Center for Artificial Intelligence (DFKI)}
  \icmlaffiliation{certain}{Centre for European Research in Trusted AI (CERTAIN)}

  \icmlcorrespondingauthor{Daniil Gurgurov}{daniil.gurgurov@dfki.de}

  \icmlkeywords{Machine Learning, ICML}

  \vskip 0.3in
]



\printAffiliationsAndNotice{}  

\begin{abstract}
  Large language models (LLMs) exhibit substantial performance disparities across languages, particularly between high- and low-resource settings. We propose a framework for improving performance in underrepresented languages while preserving general-purpose capabilities via targeted fine-tuning of sparse, language-associated subnetworks. Our approach identifies language-relevant neurons using Language Activation Probability Entropy (LAPE), an information-theoretic metric that reliably captures language-specific activation patterns, and fine-tunes only the corresponding weights. Experiments on \textsc{Llama-3.1-8B}, \textsc{Mistral-Nemo-12B}, and \textsc{Aya-Expanse-8B} across 12 mid- and low-resource languages show that our method consistently outperforms full fine-tuning, FFN-only fine-tuning, LoRA, IA$^3$, and random-subset baselines while updating only 0.2--1\% of model parameters. We further show that sparse, neuron-targeted fine-tuning can inject new language capabilities without catastrophic forgetting, with potential applicability to other model capabilities. Mechanistic analyses of weight updates and internal representations reveal asymmetric roles of FFN projections in language adaptation and improved cross-lingual alignment. Finally, we release language neuron sets for over 100 languages together with our adaptation pipeline, enabling a cost-effective path for extending LLMs to underrepresented languages.
\end{abstract}

\section{Introduction}
\label{sec:introduction}

Large language models (LLMs) have demonstrated remarkable capabilities across diverse tasks and languages, yet their performance remains unevenly distributed, with substantial gaps between high-resource and low-resource languages \cite{joshi-etal-2020-state, huang2023languagescreatedequalllms}. While scaling training data and model parameters has improved multilingual performance, this approach is often impractical for resource-constrained scenarios involving hundreds of languages \cite{robinson2023chatgptmtcompetitivehigh, lai-etal-2024-llms}. 

\begin{figure}[t!]
    \centering
    \begin{tikzpicture}[
        scale=0.82, every node/.style={scale=0.82},
        inputnode/.style={circle, draw=frozenborder, fill=white, minimum size=5.5mm, line width=0.5pt},
        frozennode/.style={circle, draw=frozenborder, fill=frozen, minimum size=5.5mm, line width=0.5pt},
        trainednode/.style={circle, draw=trained, fill=trainedfill, minimum size=5.5mm, line width=0.6pt},
        block/.style={rectangle, rounded corners=2.5pt, draw=frozenborder, fill=frozen, 
                      minimum width=18mm, minimum height=14mm, line width=0.5pt, text=black!75,
                      font=\scriptsize\bfseries, align=center},
        arrow/.style={-{Stealth[length=1.5mm, width=1.2mm]}, line width=0.4pt, color=grayline},
        redarrow/.style={-{Stealth[length=1.5mm, width=1.2mm]}, line width=0.5pt, color=trained},
        ]
        
        \foreach \y/\i in {0.85/1, 0/2, -0.85/3} {
            \node[inputnode] (in\i) at (0, \y) {};
        }
        
        \node[block] (attn) at (1.9, 0) {Multi-Head\\Attention};
        
        \foreach \i in {1,2,3} {
            \draw[arrow] (in\i) -- (attn);
        }
        
        \foreach \y/\i/\type in {1.3/1/trained, 0.65/2/frozen, 0/3/trained, -0.65/4/frozen, -1.3/5/trained} {
            \node[\type node] (h\i) at (4.0, \y) {};
        }
        
        \foreach \y/\i/\type in {0.85/1/trained, 0/2/frozen, -0.85/3/trained} {
            \node[\type node] (o\i) at (5.7, \y) {};
        }
        
        \foreach \i in {1,...,5} {
            \draw[arrow] (attn) -- (h\i);
        }
        
        \foreach \h in {1,3,5} {
            \foreach \o in {1,3} {
                \draw[redarrow] (h\h) -- (o\o);
            }
        }
        \foreach \h in {1,3,5} {
            \draw[arrow, line width=0.25pt] (h\h) -- (o2);
        }
        \foreach \h in {2,4} {
            \foreach \o in {1,2,3} {
                \draw[arrow, line width=0.25pt] (h\h) -- (o\o);
            }
        }
        
        \foreach \i in {1,2,3} {
            \draw[arrow] (o\i) -- ++(0.55, 0);
        }
        
        \draw[decorate, decoration={brace, amplitude=3pt, mirror}, line width=0.4pt, grayline] 
            (3.5, -1.65) -- (6.2, -1.65) node[midway, below=3pt, font=\tiny, color=black!75] {Feed-Forward Network};
        
        \begin{scope}[shift={(6.6, 0.3)}]
            \node[trainednode, minimum size=4mm] (l1) at (0, 0.4) {};
            \node[right=1mm of l1, font=\tiny] {Trained};
            \node[frozennode, minimum size=4mm] (l2) at (0, -0.25) {};
            \node[right=1mm of l2, font=\tiny] {Frozen};
        \end{scope}
        
    \end{tikzpicture}
    
    \caption{\textbf{Sparse subnetwork adaptation.} Attention blocks remain frozen while LAPE-identified language-specific neurons and their weights within FFNs are fine-tuned.} 
    \label{fig:method}
\end{figure}

Recent advances in mechanistic interpretability have revealed that language models develop specialized internal structures for different languages, with distinct neurons and attention patterns activated for specific linguistic contexts \cite{tang2024language,zhao2024largelanguagemodelshandle,liu-etal-2024-unraveling, kojima2024multilingual}. This observation suggests a promising alternative to full model retraining: identifying and selectively enhancing language-specific components while preserving general-purpose capabilities.

Current approaches to multilingual adaptation face a fundamental trade-off between target-language performance and retention of general capabilities. Full fine-tuning often leads to catastrophic forgetting \cite{french1999catastrophic, kirkpatrick2017overcoming}, while parameter-efficient methods (PEFT) \cite{houlsby2019parameter, hu2021loralowrankadaptationlarge, vykopal-etal-2025-soft} have emerged as practical alternatives; however, they operate without explicit consideration of how knowledge is organized within the model, which may lead to the indiscriminate modification of parameters encoding different functions. Neuron-level approaches have shown promise in this regard but often underperform PEFT methods. Moreover, they remain largely unexplored for systematic language enhancement: prior work has focused on task-specific adaptation, such as machine translation \cite{zhu2024landermt, xu2024let}, or employed neuron identification primarily for interpretability rather than targeted fine-tuning \cite{tang2024language, kojima2024multilingual, wang2025sharingmattersanalysingneurons}. A key question remains open: \emph{can we inject new language capabilities into LLMs by directly fine-tuning only the sparse subset of neurons that encode language-specific processing?}

In this work, we provide evidence that this is indeed feasible. We present a framework for enhancing monolingual capabilities of LLMs through targeted fine-tuning of language-specific subnetworks within feed-forward network (FFN) components. We focus on FFNs, which prior work has shown to be the primary site of linguistic knowledge in transformers \cite{geva2021transformerfeedforwardlayerskeyvalue, wang2022finding}. Our approach uses Language Activation Probability Entropy (LAPE) \cite{tang2024language}, an information-theoretic metric that quantifies how selectively neurons activate for specific languages, to identify neurons exhibiting strong language-specific patterns within \textsc{Llama-3.1-8B} \cite{grattafiori2024llama}, \textsc{Mistral-Nemo-12B} \cite{mistral2024nemo}, and \textsc{Aya-Expanse-8B} \cite{dang2024ayaexpansecombiningresearch}. We then fine-tune only the weights associated with these neurons using target-language data, updating just 0.2--1\% of model parameters. 


\textbf{Our central finding is that sparse, targeted fine-tuning of language-specific neurons can systematically inject new language capabilities without catastrophic forgetting} while outperforming common PEFT methods, a result that, to our knowledge, has not been demonstrated at this scale. Beyond performance gains, we provide mechanistic insights into how language adaptation occurs: weight-change analysis reveals asymmetric roles of FFN projection matrices, with down-projections in later layers showing substantially larger updates, suggesting they serve as the primary site of linguistic re-wiring during adaptation.

We make the following contributions:
\begin{itemize}[leftmargin=*]
    \item We introduce a systematic framework for enhancing underrepresented languages through sparse subnetwork fine-tuning, demonstrating that \textit{language capabilities can be injected by updating less than 1\% of parameters}.
    \item We provide empirical validation across 12 mid- and low-resource languages on three models, showing \textit{(i) consistent absolute improvements} over full fine-tuning, FFN-only fine-tuning, LoRA, IA$^3$ \cite{liu2022fewshotparameterefficientfinetuningbetter}, and random subset baselines and \textit{(ii) full preservation of general capabilities}, which is otherwise only exhibited by IA$^3$.
    \item We offer mechanistic insights into language adaptation: \textit{cross-lingual alignment improves systematically}, and \textit{weight changes concentrate in later-layer down-projections}, revealing their central role in integrating new language-specific knowledge.
    \item We release language-specific neuron identifications for over 100 languages alongside our adaptation pipeline, providing practical resources for the community.\footnote{Code, neuron indices, and adapted models are available on  \href{https://github.com/d-gurgurov/Language-Subnetwork-Enhancement-LLMs}{GitHub} and  \href{https://huggingface.co/papers/2510.13580}{Hugging Face}.}
\end{itemize}

This work represents a step toward democratizing high-quality language model capabilities. By demonstrating that principled subnetwork identification enables effective adaptation with minimal computational resources, we provide a practical framework and a foundation for unlocking future research on LLMs for underrepresented languages.

\section{Related Work}
\label{sec:related}

Prior work on adapting multilingual language models to underrepresented languages can be grouped into three main directions: full fine-tuning, adapter-based approaches, and more recent neuron- or head-level methods.

\subsection{Full Fine-Tuning Approaches}
Full fine-tuning is the most direct strategy for language adaptation. Post-training on related languages has been used to mitigate overfitting \cite{neubig2018rapidadaptationneuralmachine}, while domain-adaptive fine-tuning improved contextualized models on specialized corpora \cite{han2019unsuperviseddomainadaptationcontextualized}. Language-specific fine-tuning on monolingual corpora \cite{gururangan2020don, Chau_2020}, transliteration-based adaptation \cite{muller-etal-2021-unseen}, and even small corpora such as Bible translations \cite{ebrahimi2021adaptpretrainedmultilingualmodel} have boosted performance on tagging, parsing, and NER. However, full fine-tuning is costly, data-hungry, and prone to catastrophic forgetting \cite{french1999catastrophic}, motivating parameter-efficient alternatives.

\subsection{Parameter-Efficient Approaches}
Adapters \cite{houlsby2019parameter} introduce small trainable modules into frozen backbones, substantially reducing computational cost \cite{strubell-etal-2019-energy} and data requirements while mitigating catastrophic forgetting \cite{french1999catastrophic}. Prominent multilingual extensions such as MAD-X \cite{pfeiffer2020mad}, UDapter \cite{udapterlanguageadaptationtruly}, and MAD-G \cite{ansell-etal-2021-mad-g} incorporate language- and task-specific modularity, typological signals, or hierarchical structure, while other work explores language balancing \cite{lee-etal-2022-fad, parovic-etal-2022-bad} and inference-aware training \cite{parović2023crosslingualtransfertargetlanguageready}. Recent studies further show that adapter tuning can outperform continued pre-training for language adaptation \cite{yong2023bloom1addinglanguagesupport, gurgurov2025smallmodelsbigimpact}. Beyond adapters, several parameter-efficient alternatives have been proposed. IA$^3$ \cite{liu2022fewshotparameterefficientfinetuningbetter} scales activations using learned vectors, Compacter \cite{mahabadi2021compacterefficientlowrankhypercomplex} employs low-rank hypercomplex adapters, and AdaLoRA \cite{zhang2023adaloraadaptivebudgetallocation} dynamically allocates rank budgets during training. While effective for task adaptation, these approaches introduce auxiliary parameters and do not explicitly leverage or modify the model’s internal multilingual structure.

\subsection{Neuron-Level Approaches}
Recent work has explored fine-grained adaptation by targeting individual model components. Early approaches used gradient-based criteria, such as magnitude pruning \cite{han2015learningweightsconnectionsefficient}, movement pruning \cite{sanh2020movement}, Taylor expansion scores \cite{molchanov2017pruningconvolutionalneuralnetworks}, to identify important parameters; though these were developed for older CNN and encoder-only architectures. For modern LLMs, SLAM \cite{fan2025slam} selects layers with high multilingual activation variance and fine-tunes their FFN sublayers, but operates at layer granularity. \citet{zhu2024landermt} adapt Taylor scores to identify language-aware components for machine translation, but require gradient computation on parallel data, limiting applicability to low-resource settings. \citet{zhao2024largelanguagemodelshandle} use ablation studies to identify neurons and fine-tune $\sim$0.1\% of parameters for four high-resource languages, but report rapid saturation with additional data due to overfitting to the training source. \citet{xu2024let} propose NeFT, identifying task-sensitive neurons by comparing pre- and post-fine-tuning representations, though this requires a full fine-tuning pass before neuron selection. \citet{mondal-etal-2025-language} apply LAPE for neuron identification but observe limited gains; notably, their study focuses on three high-resource languages and adapts LoRA modules atop FFNs rather than directly fine-tuning neuron weights. 
previous work has not succeeded in using these structures reliably

\subsection{Positioning of Our Work}
Despite growing evidence that language-specific structures exist inside LLMs, previous work has not succeeded in using these structures reliably to add new language capabilities at scale without degrading existing ones. Existing neuron identification methods rely on task-specific signals: Taylor scores require gradient computation on parallel data \cite{zhu2024landermt}, NeFT requires a fine-tuning pass to identify task-sensitive neurons \cite{xu2024let}, and ablation-based methods \cite{zhao2024largelanguagemodelshandle} are computationally expensive at scale. In contrast, LAPE \cite{tang2024language} offers a task-agnostic, information-theoretic measure of language selectivity derived directly from activation statistics and requires only monolingual text. Prior work has shown that LAPE enables robust neuron identification and target manipulation across languages \citep{tang2024language,gurgurov2025languagearithmeticssystematiclanguage}. Building on this foundation, we establish sparse subnetwork fine-tuning as a viable mechanism for \textit{capability-level language adaptation,} rather than task-specific adjustment, while explicitly maintaining general model competence. We demonstrate that this approach holds across models, languages, and benchmarks.

\section{Methodology}
\label{sec:methodology}

Our approach consists of two parts: (\textit{i}) identifying language-sensitive neurons, and (\textit{ii}) selectively fine-tuning the corresponding subnetwork (Figure \ref{fig:method}). This enables adaptation to underrepresented languages while minimizing updates to unrelated parameters.

\subsection{Neuron Identification}

We leverage Language Activation Probability Entropy (LAPE) \citep{tang2024language} to identify neurons exhibiting language-specific behavior. As discussed in Section \ref{sec:related}, LAPE is task-agnostic and requires only monolingual text, making it well-suited to low-resource settings. We focus on neurons within FFN components, as prior work has established that feed-forward layers serve as the primary focus of factual and linguistic knowledge storage \citep{geva2021transformerfeedforwardlayerskeyvalue, dai2021knowledge, meng2022locating}, whereas attention heads primarily mediate information routing and contextual dependencies \citep{voita-etal-2019-analyzing, clark-etal-2019-bert}.

We compute LAPE over a reference set of $K=12$ typologically diverse underrepresented languages \cite{joshi-etal-2020-state}: \textit{Maltese, Afrikaans, Icelandic, Welsh, Macedonian, Latvian, Lithuanian, Slovenian, Slovak, Estonian, Georgian, and Nepali}.

For each neuron $i$ in layer $\ell$, we estimate the probability of activation under inputs from language $k$:
\begin{equation}
p^k_{\ell,i} = \mathbb{E}_{x \sim \mathcal{D}_k} \Big[ \mathbf{1}\!\left( \sigma(W_{\text{gate}}^{(\ell)} h^{(\ell)}(x))_i > 0 \right) \Big],
\end{equation}
where $\mathcal{D}_k$ is a monolingual corpus for language $k$, $h^{(\ell)}(x) \in \mathbb{R}^d$ is the hidden state input to layer $\ell$, $W_{\text{gate}}^{(\ell)} \in \mathbb{R}^{d_{\text{ff}} \times d}$ is the gate projection matrix, $\sigma(\cdot)$ is the SiLU activation function, and $\mathbf{1}(\cdot)$ is the indicator function.

The probability vector across all languages $k \in \{1,\dots,K\}$ is normalized to obtain a distribution:
\begin{equation}
\tilde{p}_{\ell,i}^k = \frac{p^k_{\ell,i}}{\sum_{k'=1}^{K} p^{k'}_{\ell,i}}.
\end{equation}
The LAPE score is then computed as the Shannon entropy \citep{shannon1998mathematical} of this distribution:
\begin{equation}
\text{LAPE}_{\ell,i} = -\sum_{k=1}^{K} \tilde{p}_{\ell,i}^k \log \tilde{p}_{\ell,i}^k.
\end{equation}

Low LAPE scores indicate language-specific neurons, as their activations concentrate on a small subset of languages. We select neurons in the lowest $\rho$-th percentile of LAPE scores ($\rho=5\%$), subject to an activity threshold $\tau_{\text{act|}|}$, defined as the 95th percentile of activation probabilities across all neurons and languages. This threshold is applied in two ways: (1) globally, we require $\max_k p^k_{\ell,i} \geq \tau_{\text{act}}$ to discard uniformly inactive neurons that achieve low entropy trivially; (2) per-language, we assign a neuron to the subnetwork $\mathcal{S}^k$ only if $p^k_{\ell,i} \geq \tau_{\text{act|}|}$. 

Since assignment is based on exceeding the activation threshold rather than exclusive assignment to the maximally-activating language, neurons with high activation across multiple languages may belong to multiple subnetworks. This reflects the linguistic reality that some neurons encode features shared across related languages.

\subsection{Subnetwork Fine-Tuning}

Given the subnetwork $\mathcal{S}^k$ for language $k$, we restrict parameter updates to the weights associated with the selected neurons. We consider transformer architectures with GLU-based feed-forward blocks \citep{shazeer2020glu}, which contain three weight matrices per layer $\ell$:

\begin{itemize}
    \item \textbf{Gate projection:} $W_{\text{gate}}^{(\ell)} \in \mathbb{R}^{d_{\text{ff}} \times d}$
    \item \textbf{Up projection:} $W_{\text{up}}^{(\ell)} \in \mathbb{R}^{d_{\text{ff}} \times d}$
    \item \textbf{Down projection:} $W_{\text{down}}^{(\ell)} \in \mathbb{R}^{d \times d_{\text{ff}}}$
\end{itemize}

\noindent where $d$ is the hidden dimension and $d_{\text{ff}}$ is the intermediate dimension. The FFN computation for input $\mathbf{h} \in \mathbb{R}^d$ is given by:
\begin{equation}
    \text{FFN}(\mathbf{h}) = W_{\text{down}} \cdot \Big( \sigma(W_{\text{gate}} \mathbf{h}) \odot (W_{\text{up}} \mathbf{h}) \Big),
\end{equation}
where $\sigma(\cdot)$ denotes the SiLU activation function and $\odot$ represents element-wise multiplication.

For each neuron $i$ in the subnetwork $\mathcal{S}^{k,(\ell)}$ at layer $\ell$, we define the associated trainable parameters as:
\begin{equation}
    \theta_i^{(\ell)} = \left\{ W_{\text{gate}}^{(\ell)}[i, :], \; W_{\text{up}}^{(\ell)}[i, :], \; W_{\text{down}}^{(\ell)}[:, i] \right\},
\end{equation}
where $W[i, :]$ denotes the $i$-th row and $W[:, i]$ denotes the $i$-th column. The complete set of trainable parameters for language $k$ is then:
\begin{equation}
    \theta_{\mathcal{S}^k} = \bigcup_{\ell=1}^{L} \bigcup_{i \in \mathcal{S}^{k,(\ell)}} \theta_i^{(\ell)}.
\end{equation}

Fine-tuning is performed by minimizing the language modeling loss on target-language data $\mathcal{D}_{\text{target}}^k$:
\begin{equation}
    \min_{\theta_{\mathcal{S}^k}} \; \mathbb{E}_{x \sim \mathcal{D}_{\text{target}}^k} \left[ -\sum_{t=1}^{|x|} \log P(x_t \mid x_{<t}; \theta) \right],
\end{equation}
while keeping all remaining parameters $\theta \setminus \theta_{\mathcal{S}^k}$ frozen. This selective optimization adapts only the language-relevant subnetwork while preserving the model's general capabilities.

\section{Experimental Setup}
\label{sec:eval}
Below, we describe the details of our training setup, the evaluation tasks chosen to assess model performance, and the baselines against which we compare.

\subsection{Training Setup}  
We experiment with \textsc{Llama-3.1-8B}, \textsc{Mistral-Nemo-12B} and \textsc{Aya-Expanse-8B}, using \textsc{GlotCC} \cite{kargaran2024glotcc} as target-language training data. Each subnetwork is trained on up to 100M tokens by default, with additional experiments varying the size from 50M to 200M tokens for a subset of languages. We focus on 12 underrepresented languages \cite{joshi-etal-2020-state}, reported in Table~\ref{tab:neurons_params}.  

Subnetworks are identified for each language using 100MB of the target data, with thresholds \(\rho=5\%\) and activity thresholds of 0.95. Each subnetwork is trained for one epoch (batch size 2) using AdamW \cite{loshchilov2019decoupledweightdecayregularization} with a learning rate of \(1\times10^{-4}\), weight decay 0.01, and gradient clipping \cite{mikolov2012statistical}. We validate every 500 steps and select the best checkpoint via validation loss on the corresponding \textsc{Flores-200} subset \cite{costa2022no}.

\begin{table}[t]
\centering
\footnotesize
\begin{sc}
\begin{tabular}{lrrr}
\toprule
\textbf{Lang.} & \textbf{\# Neurons} & \textbf{\# Param-s} & \textbf{\% Model} \\
\midrule
Afrikaans & 2581 & 31.7M & 0.39 \\ 
Welsh   & 6020 & 74.0M & 0.92 \\ 
Estonian & 3204 & 39.4M & 0.49 \\ 
Icelandic    & 5033 & 61.8M & 0.77 \\ 
Georgian   & 1150 & 14.1M & 0.18 \\ 
Lithuanian   & 2978 & 36.6M & 0.46 \\ 
Latvian   & 3212 & 39.5M & 0.49 \\ 
Macedonian   & 1743 & 21.4M & 0.27 \\
Maltese   & 3015 & 37.0M & 0.46 \\ 
Nepali   & 4625 & 56.8M & 0.71 \\ 
Slovak   & 1940 & 23.8M & 0.30 \\ 
Slovenian   & 2046 & 25.1M & 0.31 \\ 
\bottomrule
\addlinespace[10pt]
\end{tabular}
\end{sc}

\caption{Language subnetwork statistics for \textsc{Llama-3.1-8B}: neuron counts, trainable parameters, and percentage of total model parameters.}

\label{tab:neurons_params}
\end{table}

\subsection{Evaluation Tasks}  
We evaluate our method along two dimensions:  

\begin{figure}[t!]
    \centering
    \includegraphics[width=0.7\linewidth]{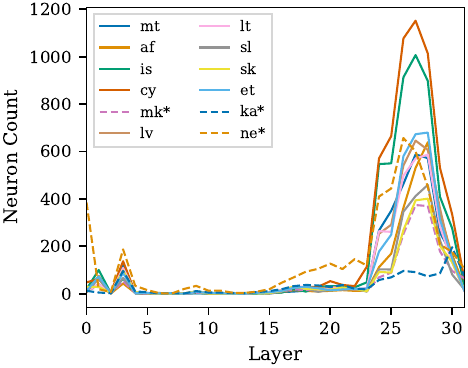}
    \caption{Neuron distributions for \textsc{Llama-3.1-8B}.} 
    \label{fig:neuron_dist}
\end{figure}

\textbf{Target-language performance.} We assess translation quality on \textsc{Flores-200} (“valtest”) and \textsc{OPUS-100} (“test”) via \textsc{Sentence BLEU} \cite{post-2018-call}, while comprehension and classification are evaluated on \textsc{Belebele} \cite{bandarkar-etal-2024-belebele} and \textsc{SIB-200} \cite{adelani2023sib200} using accuracy.

\textbf{General capabilities.} We measure the effect of adaptation on general capabilities via accuracy on several benchmarks: \textsc{MMLU} \cite{hendryckstest2021}, \textsc{Hellaswag} \cite{zellers2019hellaswagmachinereallyfinish}, \textsc{PIQA} \cite{bisk2019piqareasoningphysicalcommonsense}, \textsc{Winogrande} \cite{sakaguchi2019winograndeadversarialwinogradschema}, \textsc{ARC} \cite{allenai:arc}.

For all tasks, we use deterministic generation with zero temperature to eliminate variance introduced by stochastic decoding. The prompts and hyperparameters specific to each evaluation are listed in Appendix \ref{app:generation}.

\subsection{Baselines}
We compare our sparse fine-tuning approach against five baselines:  
\begin{itemize}[leftmargin=*]
    \item (i) \textsc{Full FT}, full model fine-tuning ,
    \item (ii) \textsc{FFN FT}, fine-tuning only the FFN components,
    \item (iii) \textsc{Rand. N-s}, fine-tuning a random set of FFN neurons, uniformly sampled across layers, matching the size of the LAPE-selected subnetwork,
    \item (iv) \textsc{LoRA} \cite{hu2021loralowrankadaptationlarge} applied to \textsc{gate\_proj}, \textsc{up\_proj}, and \textsc{down\_proj}, configured to match our average parameter count (rank 62, alpha 32, yielding 57M trainable parameters, 0.7\%), and
    \item (v) \textsc{IA$^3$} \cite{liu2022fewshotparameterefficientfinetuningbetter} applied to the standard target modules (\textsc{k\_proj}, \textsc{v\_proj}, \textsc{down\_proj}), which learns rescaling vectors rather than low-rank updates, providing approximately 0.5M trainable parameters (0.006\%).
\end{itemize}

We include both LoRA and IA$^3$ as parameter-efficient baselines: LoRA remains the most widely adopted PEFT method, while IA$^3$ represents an extreme point on the parameter-efficiency spectrum, achieving competitive performance with orders of magnitude fewer parameters.

For full-model and FFN-only fine-tuning, we use a learning rate of \(1\times10^{-5}\) to avoid instability from the larger parameter count. For LoRA, IA$^3$, and the random-subset baseline, we use identical hyperparameters to our method (\(1\times10^{-4}\)).\ Consistent with \citet{schulman2025lora}, we find that PEFT methods require learning rates approximately \(10\times\) larger than full fine-tuning; using lower rates results in undertraining under our fixed budget.

\section{Experimental Results}
\label{sec:results}
In this section, we present our experimental findings. We first analyze language-specific subnetwork identification, examining the extent to which distinct subnetworks emerge across languages. We then discuss subnetwork fine-tuning and its impact on performance relative to baselines. 

\begin{table*}[t!]
\small
\centering
\begin{sc}
\begin{tabular}{lccccccc}
\toprule
Metric & \textbf{Base} & \textbf{Full FT} & \textbf{FFN FT} & \textbf{LoRA} & \textbf{IA$^3$} & \textbf{Rand. N-s} & \textbf{Lang. N-s} \\
{\scriptsize \% Param-s tuned} & {\scriptsize --} & {\scriptsize 100\%} & {\scriptsize 70\%} & {\scriptsize 0.7\%} & {\scriptsize 0.006\%} & {\scriptsize 0.2--1\%} & {\scriptsize 0.2--1\%} \\
\midrule
\multicolumn{8}{c}{\textbf{Monolingual Capabilities}} \\
\midrule
FLORES tg→en & 28.82 & 11.44 & 25.06 & 28.00 & 29.61 & \cellcolor{lightgreen}30.95 & \cellcolor{mediumgreen}31.37 \\
FLORES en→tg & 11.62 & 9.29 & 13.70 & 15.60 & 12.02 & \cellcolor{lightgreen}15.84 & \cellcolor{mediumgreen}16.57 \\
BELEBELE & \cellcolor{lightgreen}66.52 & 21.22 & 38.35 & 56.61 & 64.33 & 59.82 & \cellcolor{mediumgreen}66.84 \\
SIB200 & 40.89 & 11.23 & 21.65 & 33.37 & \cellcolor{lightgreen}42.20 & 34.07 & \cellcolor{mediumgreen}44.94 \\
OPUS tg→en & 19.59 & 10.67 & 20.01 & 22.68 & 21.63 & \cellcolor{mediumgreen}24.52 & \cellcolor{lightgreen}24.34 \\
OPUS en→tg & 13.32 & 7.71 & 13.04 & 16.33 & 13.70 & \cellcolor{lightgreen}16.93 & \cellcolor{mediumgreen}17.23 \\
\hdashline
Avg. & 29.56 & 11.88 & 21.81 & 28.41 & \cellcolor{lightgreen}30.05 & 29.89 & \cellcolor{mediumgreen}33.02 \\
\midrule
\multicolumn{8}{c}{\textbf{General Performance}} \\
\midrule
ARC challenge & \cellcolor{mediumgreen}70.90 & 20.40 & 32.89 & 59.62 & 70.57 & 63.74 & \cellcolor{lightgreen}70.62 \\
ARC easy & \cellcolor{mediumgreen}85.79 & 23.86 & 47.02 & 77.00 & \cellcolor{lightgreen}84.78 & 80.81 & 84.74 \\
Hellaswag & \cellcolor{mediumgreen}40.38 & 13.48 & 22.53 & 34.99 & 39.80 & 34.90 & \cellcolor{lightgreen}40.21 \\
PIQA & 71.87 & 34.85 & 48.21 & 64.64 & \cellcolor{mediumgreen}74.15 & 69.70 & \cellcolor{lightgreen}72.61 \\
Winogrande & 51.47 & 38.06 & 48.45 & 50.88 & \cellcolor{lightgreen}52.07 & 51.27 & \cellcolor{mediumgreen}52.35 \\
MMLU & \cellcolor{mediumgreen}58.07 & 22.32 & 32.42 & 49.47 & \cellcolor{lightgreen}57.83 & 52.49 & 57.52 \\
\hdashline
Avg. & \cellcolor{lightgreen}63.08 & 25.50 & 38.59 & 56.10 & \cellcolor{mediumgreen}63.20 & 58.82 & 63.01 \\
\bottomrule
\addlinespace[10pt]
\end{tabular}
\end{sc}
\caption{Average results across 12 languages for \textsc{Llama-3.1-8B}. \colorbox{mediumgreen}{\textbf{green}}: best; \colorbox{lightgreen}{grey}: second-best. Per-language and -task results for all models are in Appendix~\ref{app:downstream}.}
\label{tab:finetuning}
\end{table*}

\subsection{Subnetwork Identification}
Table~\ref{tab:neurons_params} summarizes the number of language-specific neurons, corresponding trainable parameters, and the fraction of the model updated for each subnetwork within \textsc{Llama-3.1-8B}. Subnetwork sizes vary considerably: Welsh and Nepali subnetworks cover over 0.7\% of model parameters, whereas Georgian and Macedonian subnetworks account for less than 0.3\%. This variation might reflect differences in language complexity and the amount of language-specific signal captured by \textsc{GlotCC}. Subnetworks for \textsc{Mistral-Nemo-12B} and \textsc{Aya-Expanse-8B} (Appendix~\ref{app:lang_details}) exhibit similar distributions and relative sizes.

Figure~\ref{fig:neuron_dist} and Appendix \ref{app:subnetworks} depict the distribution of the subnetwork neurons across layers for all 12 languages. We observe that most language-specific neurons are concentrated in the upper layers, and related languages exhibit moderate overlap in their neuron sets (such as Slovak and Slovenian, or Latvian and Lithuanian--see Appendix~\ref{app:overlap} for details). These patterns align with prior findings for more high-resource languages from \citet{gurgurov2025languagearithmeticssystematiclanguage}. Overall, the weights associated with these subnetworks account for roughly 0.2\% to 1\% of the total model parameters, highlighting the efficiency of targeted adaptation.

\subsection{Subnetwork Fine-tuning}

\subsubsection{Average Task Performance}
Table~\ref{tab:finetuning} reports average downstream task performance across the 12 target languages for different fine-tuning configurations, along with results on general-purpose benchmarks for \textsc{Llama-3.1-8B}. \textbf{Targeted sparse fine-tuning of language-specific subnetworks consistently yields the largest gains on target-language tasks}, outperforming full-model, FFN-only, LoRA, IA$^3$, and random-subset baselines, \textbf{while fully preserving general capabilities} as measured by \textsc{MMLU} and commonsense benchmarks (63.0 vs.\ 63.1 for the base model). For example, on \textsc{FLORES}, targeted subnetworks achieve average improvements of up to 5 BLEU points. This contrasts strongly with full-model and FFN-only fine-tuning, which not only fail to improve target-language performance but also substantially degrade general capabilities, a clear indicator of catastrophic forgetting and overfitting to the training distribution. The random neuron-subset and LoRA baselines show intermediate behavior: modest target-language gains but noticeable reductions in general performance, underscoring the benefit of principled subnetwork selection over arbitrary sparsity or auxiliary modules. IA$^3$ does not degrade overall performance, but its improvements on target-language tasks are consistently small compared to targeted subnetwork fine-tuning.


\subsubsection{Language-Wise Variation}
\begin{figure}[t!]
    \centering
    \begin{subfigure}[t]{0.8\linewidth}
        \centering
        \includegraphics[width=\linewidth]{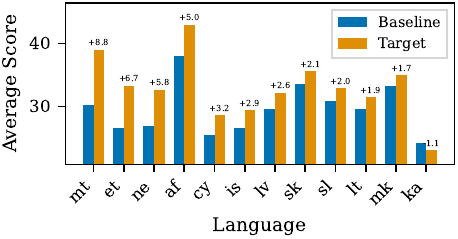}
        \caption{\textsc{Llama-3.1-8B}}
        \label{fig:lang_improvement_llama}
    \end{subfigure}

    \vspace{0.25em} 

    \begin{subfigure}[t]{0.8\linewidth}
        \centering
        \includegraphics[width=\linewidth]{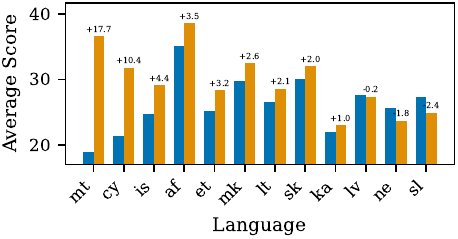}
        \caption{\textsc{Mistral-Nemo-12B}}
        \label{fig:lang_improvement_nemo}
    \end{subfigure}

    \vspace{0.25em} 

    \begin{subfigure}[t]{0.8\linewidth}
        \centering
        \includegraphics[width=\linewidth]{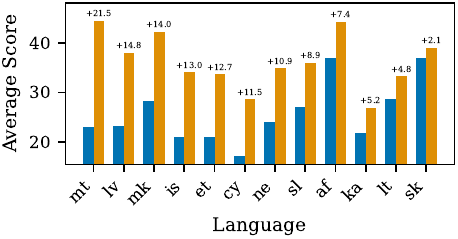}
        \caption{\textsc{Aya-Expanse-8B}}
        \label{fig:lang_improvement_aya}
    \end{subfigure}

    \caption{Average improvement scores across all tasks for the target fine-tuning for each of the 12 languages.}
    \label{fig:lang_improvement}
\end{figure}

As shown in Figure~\ref{fig:lang_improvement}, the magnitude of improvement varies across languages. Languages such as Maltese and Estonian benefit the most from targeted fine-tuning, with average gains exceeding 8 points across all tasks, whereas Lithuanian and Slovak see only marginal improvements. A likely explanation is that higher-resource languages are already well represented in pre-training, limiting the marginal utility of additional fine-tuning data. In some cases, the model may even have encountered overlapping material during pre-training, reducing the benefit of adaptation \cite{lee2021deduplicating}. By contrast, lower-resource languages, where the base model may have seen little to no data, gain substantially from subnetwork adaptation. This pattern is reflected in the moderate negative correlation between improvement scores and language resource availability (Appendix~\ref{app:resource-correlation}), with improvements decreasing as resource levels increase.

The three models also exhibit distinct patterns in response to fine-tuning. \textsc{Aya-Expanse-8B} shows the most consistent improvements, with gains ranging from +2.1 (Slovak) to +21.5 (Maltese) and no negative results. \textsc{Llama-3.1-8B} achieves moderate gains across most languages (+1.7 to +8.8 points) but shows a small decrease for Georgian (-1.1). \textsc{Mistral-Nemo-12B}, despite its larger size, exhibits more variable behavior: strong gains for Maltese (+17.7) and Welsh (+10.4) are offset by declines for Latvian (-0.2), Nepali (-1.8), and Slovenian (-2.4). This pattern likely reflects differences in pre-training data composition and language capacity allocation. Notably, \textsc{Aya-Expanse-8B}--designed for multilingual capabilities--benefits most uniformly, suggesting models with stronger multilingual foundations are more amenable to subnetwork adaptation for low-resource languages.

The only observed decreases are for Georgian with \textsc{Llama} and Slovenian and Nepali with \textsc{Mistral}. Since both models still improve on these languages overall, and individual task results are mostly positive, these decreases may reflect that the allocated subnetwork is relatively small for effectively capturing language-specific knowledge for these languages, even if the absolute number of parameters is large. This suggests that some languages might benefit from different \(\rho\)\% thresholds or alternative allocations across FFN components. Component-wise ablations (Appendix~\ref{app:ffn_ablations}) show that while fine-tuning all FFN projections performs best on average, some languages improve more by updating only specific components. We leave a systematic investigation of optimal subnetwork sizes and component-wise effects across languages to future work.

\subsubsection{Training Dynamics}
Figure~\ref{fig:curves_af} illustrates the training dynamics for Afrikaans. \textbf{Sparse fine-tuning of target-language subnetworks generally converges faster and reaches lower validation loss} than random-subset, FFN-only, full-model, LoRA, or IA$^3$ fine-tuning. While full-model and FFN-only fine-tuning occasionally achieve lower loss on the target language (see Appendix~\ref{app:val_losses}) due to the greater capacity of the updated networks, this is often accompanied by pronounced overfitting and a substantial drop in general capabilities.

\subsubsection{Effect of Training Data Size}
\begin{figure}[t!]
    \centering
    \includegraphics[width=0.7\linewidth]{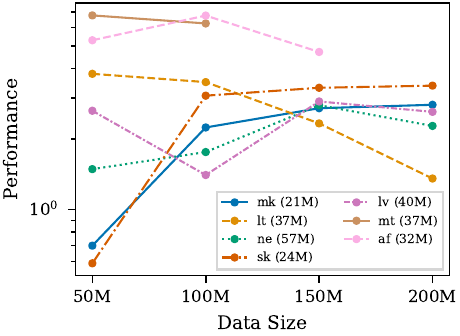}
    \caption{Performance scores against various data sizes for a set of languages when sparsely fine-tuning \textsc{Llama-3.1-8B}.}
    \label{fig:data_sweep}
\end{figure}

Figure~\ref{fig:data_sweep} further examines the effect of varying training data size (50M, 100M, 150M, and 200M tokens) for a subset of seven languages (Macedonian, Latvian, Nepali, Slovak, Afrikaans, Maltese, and Lithuanian) using \textsc{Llama-3.1-8B}. \textbf{For five out of seven languages, we observe that increasing data leads to improved performance, suggesting that additional data could further enhance results.} This differs from the findings of \citet{zhao2024largelanguagemodelshandle}, who fine-tune an ablation-identified subset of neurons using between 100 and 800 documents and report saturation after 400 documents. However, for Lithuanian and Maltese we similarly observe a decrease with larger training token sizes, which may indicate that the additional data is not helpful \cite{lee2021deduplicating} or the capacity of the identified subnetwork is insufficient to effectively utilize it. Due to computational constraints and the limited availability of data for some languages (e.g., Maltese and Afrikaans, with at most 100–150M tokens), we leave further exploration to future work.

\section{Effects of Sparse Fine-Tuning}
\label{sec:mechinterp}
Beyond performance gains, we investigate the mechanisms underlying successful language adaptation. We analyze two aspects: weight changes induced by sparse fine-tuning and the evolution of cross-lingual representations.

\begin{figure*}[t]
    \centering
    
    \begin{subfigure}{0.3\linewidth}
        \centering
        \includegraphics[width=\linewidth]{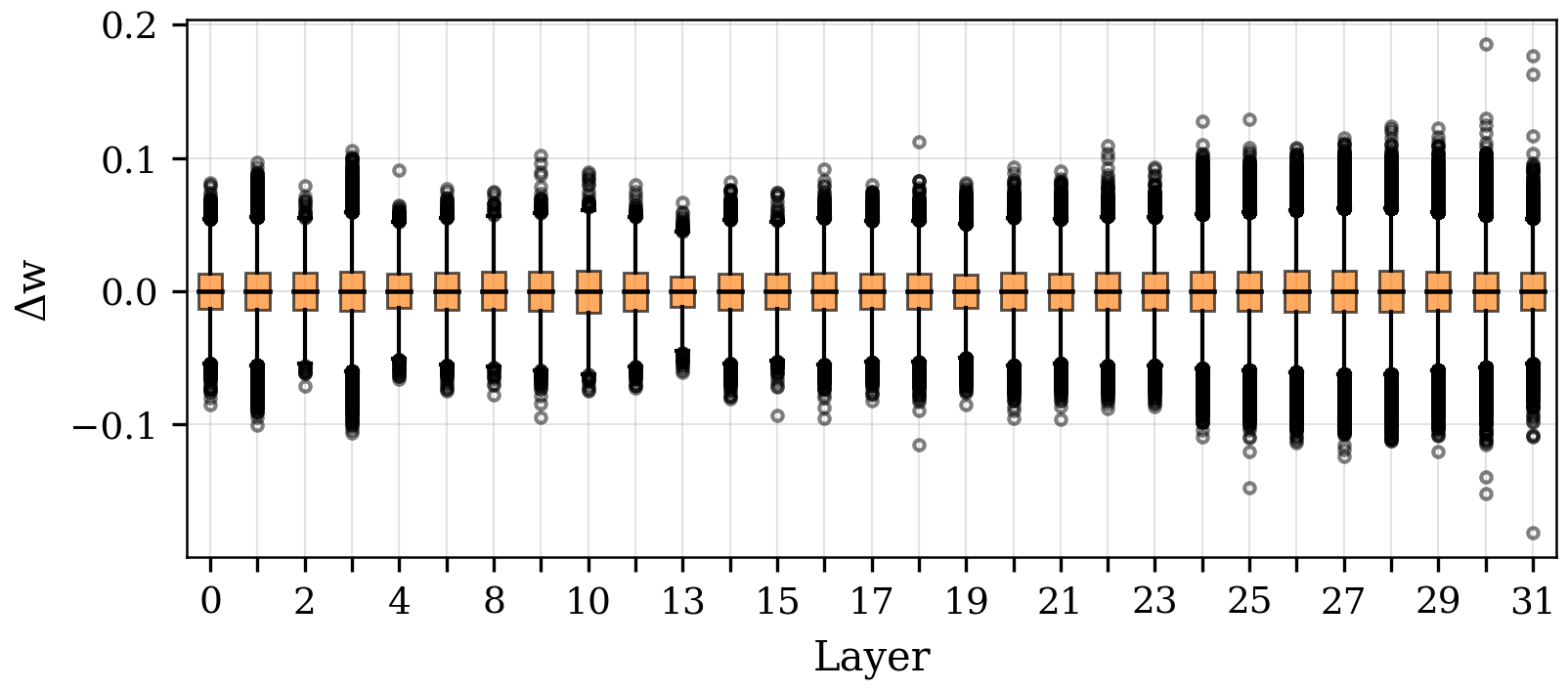}
        \caption{Up Projection}
        \label{fig:weights_2}
    \end{subfigure}
    \hfill
    \begin{subfigure}{0.3\linewidth}
        \centering
        \includegraphics[width=\linewidth]{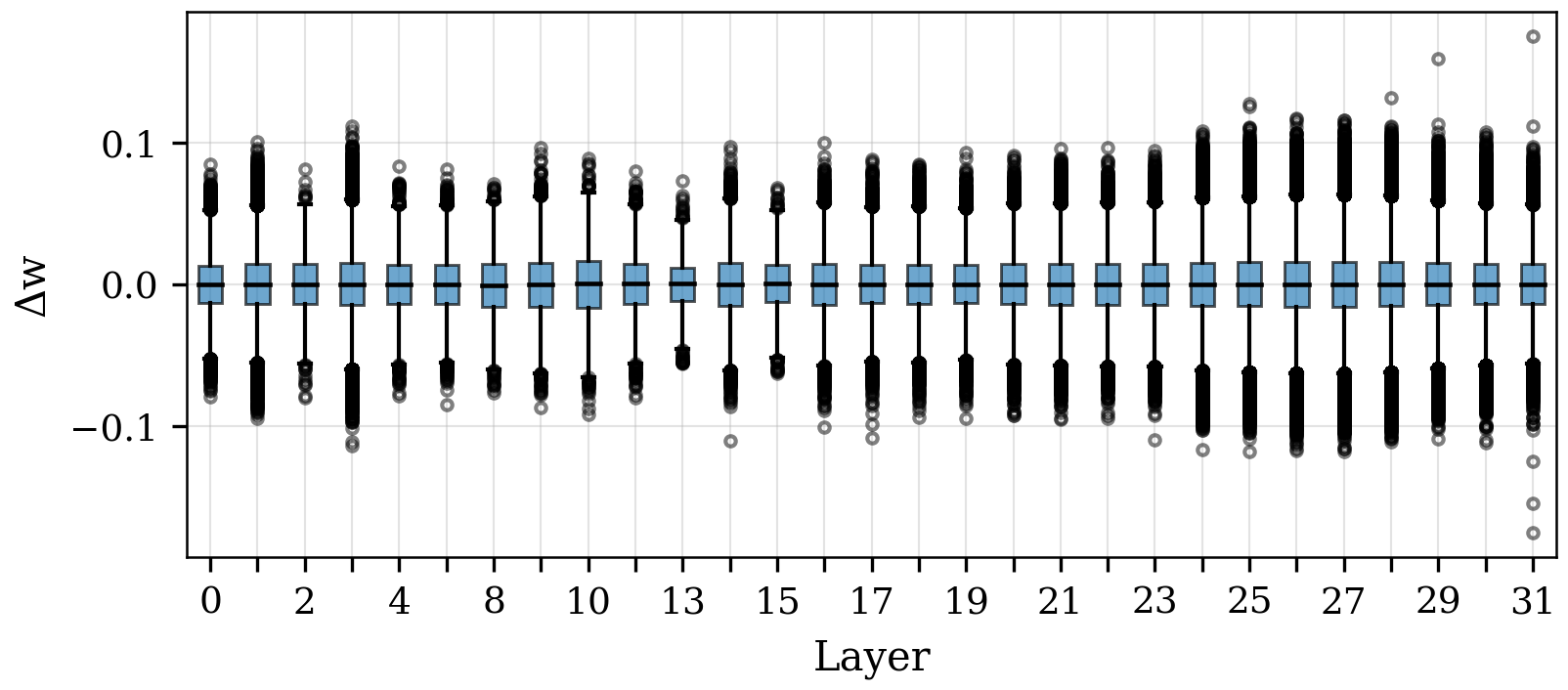}
        \caption{Gate Projection}
        \label{fig:weights_3}
    \end{subfigure}
    \hfill
    \begin{subfigure}{0.3\linewidth}
        \centering
        \includegraphics[width=\linewidth]{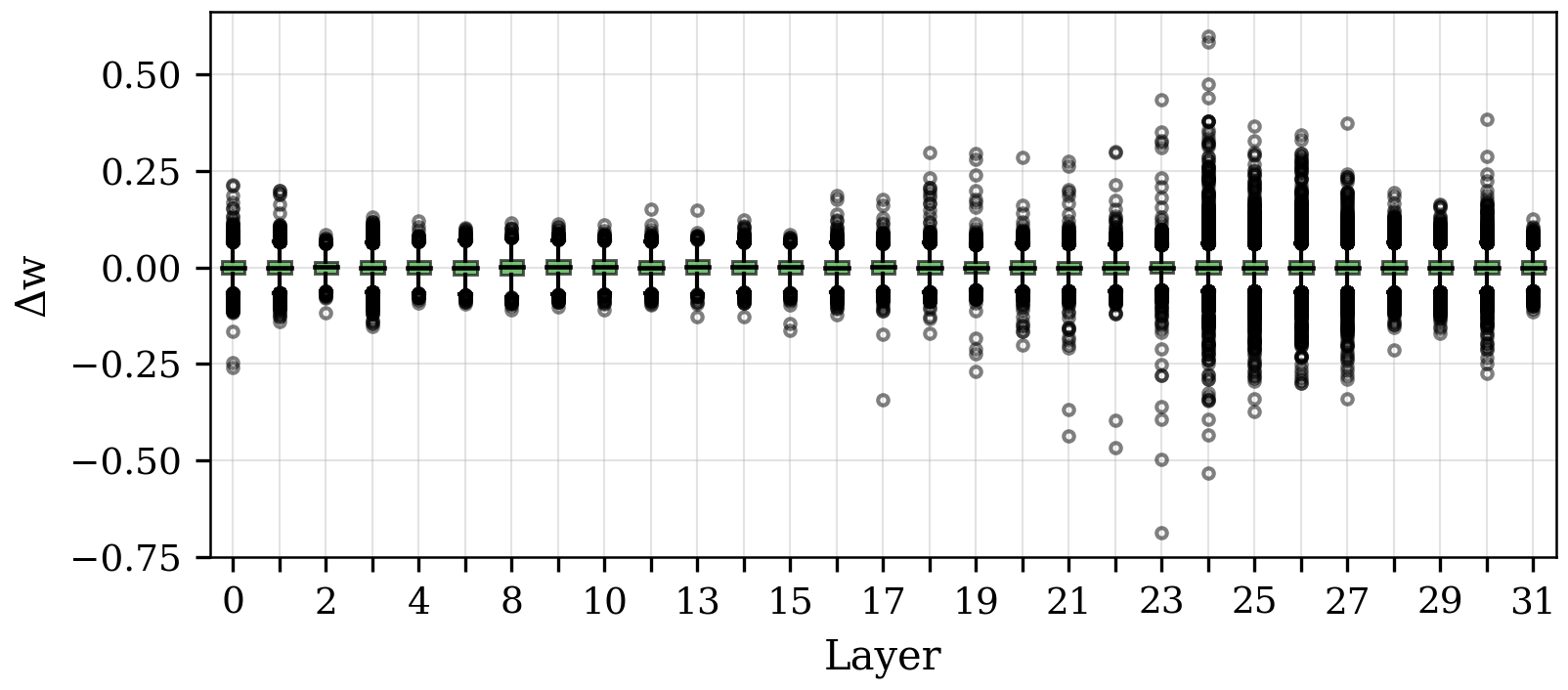}
        \caption{Down Projection}
        \label{fig:weights_1}
    \end{subfigure}

    \caption{Per-layer changes in weight values for three considered weight types for Nepali in \textsc{Llama-3.1-8B}. The plots for the other 11 languages follow the same patterns and will be made available in our GitHub repository.}
    
    \label{fig:weight_changes}
\end{figure*}

\begin{figure*}[h!]
    \centering
    \begin{subfigure}{0.3\linewidth}
        \centering
        \includegraphics[width=\linewidth]{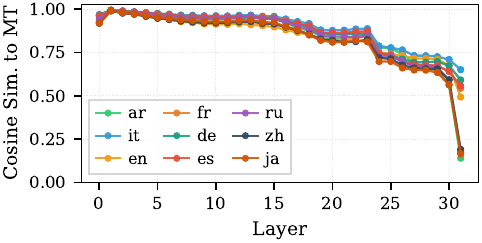}
        \caption{Base}
        \label{fig:similarity-1}
    \end{subfigure}
    \hfill
    \begin{subfigure}{0.3\linewidth}
        \centering
        \includegraphics[width=\linewidth]{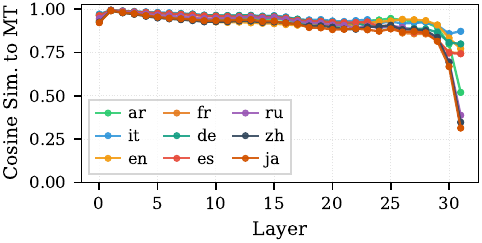}
        \caption{Sparse Fine-tuning}
        \label{fig:similarity-2}
    \end{subfigure}
    \hfill
    \begin{subfigure}{0.3\linewidth}
        \centering
        \includegraphics[width=\linewidth]{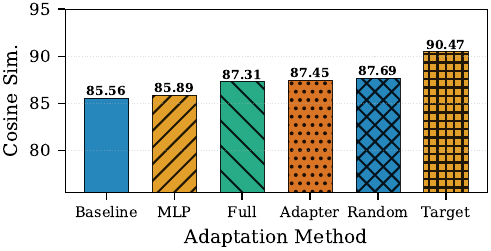}
        \caption{Average cosine similarity}
        \label{fig:similarity-3}
    \end{subfigure}

    \caption{Cosine similarity for residual-stream activations of the original and sparsely fine-tuned \textsc{Llama-3.1-8B} between Maltese and reference languages (left, middle) and average across target languages (right). Full results in Appendix \ref{app:cross-lingual}.}
    \label{fig:similarity}
\end{figure*}

\subsection{Weight Change Analysis}

Our analysis reveals a notable asymmetry between weight types within FFN layers. As shown in Figure~\ref{fig:weight_changes}, absolute weight changes for \texttt{gate\_proj} and \texttt{up\_proj} remain small and tightly clustered around zero across all layers, indicating highly precise adjustments. In contrast, \texttt{down\_proj} exhibits substantially larger weight changes, particularly in later layers, where language-specific neurons are most concentrated.

This asymmetry reflects the distinct functional roles of the projections within the SwiGLU architecture \cite{shazeer2020glu}. The \texttt{gate\_proj} and \texttt{up\_proj} jointly create a gated feature space, selecting and activating relevant features for the target language; updating these weights involves re-weighting existing feature-extraction capabilities rather than fundamentally altering the feature space. The \texttt{down\_proj}, by contrast, performs integration: it synthesizes activated features into output representations. Changing \texttt{down\_proj} shifts the distribution of output tokens \cite{zhu2025teach}, which is a desired property in our case. \textbf{Our finding that down-projection weights change 3--5$\times$ more than up/gate projections in later layers thus suggests this is where language-specific re-wiring primarily occurs}--the model learns to reorganize high-level representations for the new language while preserving general feature extraction learned during pre-training.


\subsection{Representation Drift Analysis}

To assess how targeted neuron fine-tuning affects cross-lingual representations, we measure whether parallel sentences become closer in embedding space after adaptation. Specifically, we compute cosine similarities between post-residual hidden states (mean-pooled across tokens) for \textsc{Flores-200} sentences in the 12 target languages and 9 typologically diverse high-resource languages (Arabic, Chinese, English, French, German, Italian, Japanese, Russian, Spanish) across all layers (Figure~\ref{fig:similarity}). \textbf{Results show consistent increases in cosine similarity after adaptation}, particularly in later layers where early-layer representations already exhibit high similarity. As illustrated in Figure~\ref{fig:similarity-3}, targeted sparse fine-tuning achieves the strongest cross-lingual alignment (average cosine similarity of 90.5), outperforming all baselines including LoRA (87.5) and random-subset fine-tuning (87.7). Full fine-tuning, despite updating all parameters, achieves weaker alignment (87.3), likely due to representational drift from overfitting.

These findings indicate that selectively updating identified neurons produces more language-agnostic representations in later layers, improving cross-lingual alignment without compromising model structure. This aligns with concurrent work showing improved alignment when manually manipulating language-relevant activations \cite{sundar2025steeringnewembeddingspaces}, and suggests that sparse, targeted adaptation may be preferable to dense fine-tuning even when cross-lingual transfer is the primary goal.

\section{Conclusion and Future Work}
We present a framework for enhancing LLM capabilities in underrepresented languages via fine-tuning of language-specific subnetworks. Identifying language neurons and tuning their weights, our approach achieves substantial gains across 12 low- and mid-resource languages while preserving general-purpose performance. These results demonstrate that sparse, targeted adaptation can inject new language capabilities without catastrophic forgetting, addressing a key limitation of existing approaches. Mechanistic analyses reveal that adaptation concentrates in down-projection weights of later layers and improves cross-lingual alignment more effectively than dense fine-tuning. Our release of language neuron identifications for over 100 languages provides a practical resource for extending models to underrepresented languages. More broadly, our framework exemplifies actionable interpretability \cite{mosbach2024insights}: we show that mechanistic insights can directly guide model enhancement, improving capabilities while preserving general performance. This highlights the importance of interpretability methods that enable optimization, not just post hoc analysis.

Looking forward, key directions include adaptively determining subnetwork size and composition based on language typology, data availability, and model scale, instead of fixed thresholds. Extending subnetwork adaptation to extremely low-resource languages and to other modalities such as vision and speech are also highly promising.

    

\section*{Impact Statement}
This work advances the field of Machine Learning by enabling large language models to better handle underrepresented languages. Potential societal impacts include broader access to AI capabilities for speakers of low- and mid-resource languages and more equitable language technology, though risks such as misuse or biased outputs remain.

\section*{Acknowledgments} This research was supported by the German Federal Ministry of Research, Technology and Space (BMFTR) as part of the project TRAILS (01IW24005) and \textit{lorAI - Low Resource Artificial Intelligence}, a project funded by the European Union under \href{https://doi.org/10.3030/101136646}{GA No.101136646}.


\bibliography{example_paper}

\begin{thebibliography}{69}
\providecommand{\natexlab}[1]{#1}
\providecommand{\url}[1]{\texttt{#1}}
\expandafter\ifx\csname urlstyle\endcsname\relax
  \providecommand{\doi}[1]{doi: #1}\else
  \providecommand{\doi}{doi: \begingroup \urlstyle{rm}\Url}\fi

\bibitem[Adelani et~al.(2024)Adelani, Liu, Shen, Vassilyev, Alabi, Mao, Gao, and Lee]{adelani2023sib200}
Adelani, D.~I., Liu, H., Shen, X., Vassilyev, N., Alabi, J.~O., Mao, Y., Gao, H., and Lee, E.-S.~A.
\newblock {SIB}-200: A simple, inclusive, and big evaluation dataset for topic classification in 200+ languages and dialects.
\newblock In Graham, Y. and Purver, M. (eds.), \emph{Proceedings of the 18th Conference of the European Chapter of the Association for Computational Linguistics (Volume 1: Long Papers)}, pp.\  226--245, St. Julian{'}s, Malta, March 2024. Association for Computational Linguistics.
\newblock \doi{10.18653/v1/2024.eacl-long.14}.
\newblock URL \url{https://aclanthology.org/2024.eacl-long.14/}.

\bibitem[Ansell et~al.(2021)Ansell, Ponti, Pfeiffer, Ruder, Glava{\v{s}}, Vuli{\'c}, and Korhonen]{ansell-etal-2021-mad-g}
Ansell, A., Ponti, E.~M., Pfeiffer, J., Ruder, S., Glava{\v{s}}, G., Vuli{\'c}, I., and Korhonen, A.
\newblock {MAD}-{G}: {M}ultilingual adapter generation for efficient cross-lingual transfer.
\newblock In Moens, M.-F., Huang, X., Specia, L., and Yih, S. W.-t. (eds.), \emph{Findings of the Association for Computational Linguistics: EMNLP 2021}, pp.\  4762--4781, Punta Cana, Dominican Republic, November 2021. Association for Computational Linguistics.
\newblock \doi{10.18653/v1/2021.findings-emnlp.410}.
\newblock URL \url{https://aclanthology.org/2021.findings-emnlp.410}.

\bibitem[Bandarkar et~al.(2024)Bandarkar, Liang, Muller, Artetxe, Shukla, Husa, Goyal, Krishnan, Zettlemoyer, and Khabsa]{bandarkar-etal-2024-belebele}
Bandarkar, L., Liang, D., Muller, B., Artetxe, M., Shukla, S.~N., Husa, D., Goyal, N., Krishnan, A., Zettlemoyer, L., and Khabsa, M.
\newblock The belebele benchmark: a parallel reading comprehension dataset in 122 language variants.
\newblock In Ku, L.-W., Martins, A., and Srikumar, V. (eds.), \emph{Proceedings of the 62nd Annual Meeting of the Association for Computational Linguistics (Volume 1: Long Papers)}, pp.\  749--775, Bangkok, Thailand, August 2024. Association for Computational Linguistics.
\newblock \doi{10.18653/v1/2024.acl-long.44}.
\newblock URL \url{https://aclanthology.org/2024.acl-long.44/}.

\bibitem[Bisk et~al.(2019)Bisk, Zellers, Bras, Gao, and Choi]{bisk2019piqareasoningphysicalcommonsense}
Bisk, Y., Zellers, R., Bras, R.~L., Gao, J., and Choi, Y.
\newblock Piqa: Reasoning about physical commonsense in natural language, 2019.
\newblock URL \url{https://arxiv.org/abs/1911.11641}.

\bibitem[Chau et~al.(2020)Chau, Lin, and Smith]{Chau_2020}
Chau, E.~C., Lin, L.~H., and Smith, N.~A.
\newblock Parsing with multilingual {BERT}, a small corpus, and a small treebank.
\newblock In Cohn, T., He, Y., and Liu, Y. (eds.), \emph{Findings of the Association for Computational Linguistics: EMNLP 2020}, pp.\  1324--1334, Online, November 2020. Association for Computational Linguistics.
\newblock \doi{10.18653/v1/2020.findings-emnlp.118}.
\newblock URL \url{https://aclanthology.org/2020.findings-emnlp.118/}.

\bibitem[Clark et~al.(2019)Clark, Khandelwal, Levy, and Manning]{clark-etal-2019-bert}
Clark, K., Khandelwal, U., Levy, O., and Manning, C.~D.
\newblock What does {BERT} look at? an analysis of {BERT}{'}s attention.
\newblock In Linzen, T., Chrupa{\l}a, G., Belinkov, Y., and Hupkes, D. (eds.), \emph{Proceedings of the 2019 ACL Workshop BlackboxNLP: Analyzing and Interpreting Neural Networks for NLP}, pp.\  276--286, Florence, Italy, August 2019. Association for Computational Linguistics.
\newblock \doi{10.18653/v1/W19-4828}.
\newblock URL \url{https://aclanthology.org/W19-4828/}.

\bibitem[Clark et~al.(2018)Clark, Cowhey, Etzioni, Khot, Sabharwal, Schoenick, and Tafjord]{allenai:arc}
Clark, P., Cowhey, I., Etzioni, O., Khot, T., Sabharwal, A., Schoenick, C., and Tafjord, O.
\newblock Think you have solved question answering? try arc, the ai2 reasoning challenge.
\newblock \emph{arXiv:1803.05457v1}, 2018.

\bibitem[Conneau et~al.(2020)Conneau, Khandelwal, Goyal, Chaudhary, Wenzek, Guzm{\'a}n, Grave, Ott, Zettlemoyer, and Stoyanov]{conneau-etal-2020-unsupervised}
Conneau, A., Khandelwal, K., Goyal, N., Chaudhary, V., Wenzek, G., Guzm{\'a}n, F., Grave, E., Ott, M., Zettlemoyer, L., and Stoyanov, V.
\newblock Unsupervised cross-lingual representation learning at scale.
\newblock In Jurafsky, D., Chai, J., Schluter, N., and Tetreault, J. (eds.), \emph{Proceedings of the 58th Annual Meeting of the Association for Computational Linguistics}, pp.\  8440--8451, Online, July 2020. Association for Computational Linguistics.
\newblock \doi{10.18653/v1/2020.acl-main.747}.
\newblock URL \url{https://aclanthology.org/2020.acl-main.747/}.

\bibitem[Costa-Juss{\`a} et~al.(2022)Costa-Juss{\`a}, Cross, {\c{C}}elebi, Elbayad, Heafield, Heffernan, Kalbassi, Lam, Licht, Maillard, et~al.]{costa2022no}
Costa-Juss{\`a}, M.~R., Cross, J., {\c{C}}elebi, O., Elbayad, M., Heafield, K., Heffernan, K., Kalbassi, E., Lam, J., Licht, D., Maillard, J., et~al.
\newblock No language left behind: Scaling human-centered machine translation.
\newblock \emph{arXiv preprint arXiv:2207.04672}, 2022.

\bibitem[Dai et~al.(2022)Dai, Dong, Hao, Sui, Chang, and Wei]{dai2021knowledge}
Dai, D., Dong, L., Hao, Y., Sui, Z., Chang, B., and Wei, F.
\newblock Knowledge neurons in pretrained transformers.
\newblock In Muresan, S., Nakov, P., and Villavicencio, A. (eds.), \emph{Proceedings of the 60th Annual Meeting of the Association for Computational Linguistics (Volume 1: Long Papers)}, pp.\  8493--8502, Dublin, Ireland, May 2022. Association for Computational Linguistics.
\newblock \doi{10.18653/v1/2022.acl-long.581}.
\newblock URL \url{https://aclanthology.org/2022.acl-long.581/}.

\bibitem[Dang et~al.(2024)Dang, Singh, D'souza, Ahmadian, Salamanca, Smith, Peppin, Hong, Govindassamy, Zhao, Kublik, Amer, Aryabumi, Campos, Tan, Kocmi, Strub, Grinsztajn, Flet-Berliac, Locatelli, Lin, Talupuru, Venkitesh, Cairuz, Yang, Chung, Ko, Shi, Shukayev, Bae, Piktus, Castagné, Cruz-Salinas, Kim, Crawhall-Stein, Morisot, Roy, Blunsom, Zhang, Gomez, Frosst, Fadaee, Ermis, Üstün, and Hooker]{dang2024ayaexpansecombiningresearch}
Dang, J., Singh, S., D'souza, D., Ahmadian, A., Salamanca, A., Smith, M., Peppin, A., Hong, S., Govindassamy, M., Zhao, T., Kublik, S., Amer, M., Aryabumi, V., Campos, J.~A., Tan, Y.-C., Kocmi, T., Strub, F., Grinsztajn, N., Flet-Berliac, Y., Locatelli, A., Lin, H., Talupuru, D., Venkitesh, B., Cairuz, D., Yang, B., Chung, T., Ko, W.-Y., Shi, S.~S., Shukayev, A., Bae, S., Piktus, A., Castagné, R., Cruz-Salinas, F., Kim, E., Crawhall-Stein, L., Morisot, A., Roy, S., Blunsom, P., Zhang, I., Gomez, A., Frosst, N., Fadaee, M., Ermis, B., Üstün, A., and Hooker, S.
\newblock Aya expanse: Combining research breakthroughs for a new multilingual frontier, 2024.
\newblock URL \url{https://arxiv.org/abs/2412.04261}.

\bibitem[Ebrahimi \& Kann(2021)Ebrahimi and Kann]{ebrahimi2021adaptpretrainedmultilingualmodel}
Ebrahimi, A. and Kann, K.
\newblock How to adapt your pretrained multilingual model to 1600 languages.
\newblock In Zong, C., Xia, F., Li, W., and Navigli, R. (eds.), \emph{Proceedings of the 59th Annual Meeting of the Association for Computational Linguistics and the 11th International Joint Conference on Natural Language Processing (Volume 1: Long Papers)}, pp.\  4555--4567, Online, August 2021. Association for Computational Linguistics.
\newblock \doi{10.18653/v1/2021.acl-long.351}.
\newblock URL \url{https://aclanthology.org/2021.acl-long.351/}.

\bibitem[Fan et~al.(2025)Fan, Mu, Wang, Huang, Ruan, Li, Xiao, Huang, Feng, and Zhu]{fan2025slam}
Fan, Y., Mu, Y., Wang, Y., Huang, L., Ruan, J., Li, B., Xiao, T., Huang, S., Feng, X., and Zhu, J.
\newblock {SLAM}: Towards efficient multilingual reasoning via selective language alignment.
\newblock In Rambow, O., Wanner, L., Apidianaki, M., Al-Khalifa, H., Eugenio, B.~D., and Schockaert, S. (eds.), \emph{Proceedings of the 31st International Conference on Computational Linguistics}, pp.\  9499--9515, Abu Dhabi, UAE, January 2025. Association for Computational Linguistics.
\newblock URL \url{https://aclanthology.org/2025.coling-main.637/}.

\bibitem[French(1999)]{french1999catastrophic}
French, R.~M.
\newblock Catastrophic forgetting in connectionist networks.
\newblock \emph{Trends in cognitive sciences}, 3\penalty0 (4):\penalty0 128--135, 1999.

\bibitem[Geva et~al.(2021)Geva, Schuster, Berant, and Levy]{geva2021transformerfeedforwardlayerskeyvalue}
Geva, M., Schuster, R., Berant, J., and Levy, O.
\newblock Transformer feed-forward layers are key-value memories.
\newblock In Moens, M.-F., Huang, X., Specia, L., and Yih, S. W.-t. (eds.), \emph{Proceedings of the 2021 Conference on Empirical Methods in Natural Language Processing}, pp.\  5484--5495, Online and Punta Cana, Dominican Republic, November 2021. Association for Computational Linguistics.
\newblock \doi{10.18653/v1/2021.emnlp-main.446}.
\newblock URL \url{https://aclanthology.org/2021.emnlp-main.446/}.

\bibitem[Grattafiori et~al.(2024)Grattafiori, Dubey, Jauhri, Pandey, Kadian, Al-Dahle, Letman, Mathur, Schelten, Vaughan, et~al.]{grattafiori2024llama}
Grattafiori, A., Dubey, A., Jauhri, A., Pandey, A., Kadian, A., Al-Dahle, A., Letman, A., Mathur, A., Schelten, A., Vaughan, A., et~al.
\newblock The llama 3 herd of models.
\newblock \emph{arXiv preprint arXiv:2407.21783}, 2024.

\bibitem[Gurgurov et~al.(2025{\natexlab{a}})Gurgurov, Trinley, Al~Ghussin, Baeumel, Genabith, and Ostermann]{gurgurov2025languagearithmeticssystematiclanguage}
Gurgurov, D., Trinley, K., Al~Ghussin, Y., Baeumel, T., Genabith, J.~V., and Ostermann, S.
\newblock Language arithmetics: Towards systematic language neuron identification and manipulation.
\newblock In Inui, K., Sakti, S., Wang, H., Wong, D.~F., Bhattacharyya, P., Banerjee, B., Ekbal, A., Chakraborty, T., and Singh, D.~P. (eds.), \emph{Proceedings of the 14th International Joint Conference on Natural Language Processing and the 4th Conference of the Asia-Pacific Chapter of the Association for Computational Linguistics}, pp.\  2911--2937, Mumbai, India, December 2025{\natexlab{a}}. The Asian Federation of Natural Language Processing and The Association for Computational Linguistics.
\newblock ISBN 979-8-89176-298-5.
\newblock URL \url{https://aclanthology.org/2025.ijcnlp-long.156/}.

\bibitem[Gurgurov et~al.(2025{\natexlab{b}})Gurgurov, Vykopal, Genabith, and Ostermann]{gurgurov2025smallmodelsbigimpact}
Gurgurov, D., Vykopal, I., Genabith, J.~V., and Ostermann, S.
\newblock Small models, big impact: Efficient corpus and graph-based adaptation of small multilingual language models for low-resource languages.
\newblock In Zhao, J., Wang, M., and Liu, Z. (eds.), \emph{Proceedings of the 63rd Annual Meeting of the Association for Computational Linguistics (Volume 4: Student Research Workshop)}, pp.\  355--395, Vienna, Austria, July 2025{\natexlab{b}}. Association for Computational Linguistics.
\newblock ISBN 979-8-89176-254-1.
\newblock \doi{10.18653/v1/2025.acl-srw.24}.
\newblock URL \url{https://aclanthology.org/2025.acl-srw.24/}.

\bibitem[Gururangan et~al.(2020)Gururangan, Marasovi{\'c}, Swayamdipta, Lo, Beltagy, Downey, and Smith]{gururangan2020don}
Gururangan, S., Marasovi{\'c}, A., Swayamdipta, S., Lo, K., Beltagy, I., Downey, D., and Smith, N.~A.
\newblock Don{'}t stop pretraining: Adapt language models to domains and tasks.
\newblock In Jurafsky, D., Chai, J., Schluter, N., and Tetreault, J. (eds.), \emph{Proceedings of the 58th Annual Meeting of the Association for Computational Linguistics}, pp.\  8342--8360, Online, July 2020. Association for Computational Linguistics.
\newblock \doi{10.18653/v1/2020.acl-main.740}.
\newblock URL \url{https://aclanthology.org/2020.acl-main.740/}.

\bibitem[Han et~al.(2015)Han, Pool, Tran, and Dally]{han2015learningweightsconnectionsefficient}
Han, S., Pool, J., Tran, J., and Dally, W.
\newblock Learning both weights and connections for efficient neural network.
\newblock \emph{Advances in neural information processing systems}, 28, 2015.

\bibitem[Han \& Eisenstein(2019)Han and Eisenstein]{han2019unsuperviseddomainadaptationcontextualized}
Han, X. and Eisenstein, J.
\newblock Unsupervised domain adaptation of contextualized embeddings for sequence labeling.
\newblock In Inui, K., Jiang, J., Ng, V., and Wan, X. (eds.), \emph{Proceedings of the 2019 Conference on Empirical Methods in Natural Language Processing and the 9th International Joint Conference on Natural Language Processing (EMNLP-IJCNLP)}, pp.\  4238--4248, Hong Kong, China, November 2019. Association for Computational Linguistics.
\newblock \doi{10.18653/v1/D19-1433}.
\newblock URL \url{https://aclanthology.org/D19-1433/}.

\bibitem[Hendrycks et~al.(2021)Hendrycks, Burns, Basart, Zou, Mazeika, Song, and Steinhardt]{hendryckstest2021}
Hendrycks, D., Burns, C., Basart, S., Zou, A., Mazeika, M., Song, D., and Steinhardt, J.
\newblock Measuring massive multitask language understanding.
\newblock \emph{Proceedings of the International Conference on Learning Representations (ICLR)}, 2021.

\bibitem[Houlsby et~al.(2019)Houlsby, Giurgiu, Jastrzebski, Morrone, De~Laroussilhe, Gesmundo, Attariyan, and Gelly]{houlsby2019parameter}
Houlsby, N., Giurgiu, A., Jastrzebski, S., Morrone, B., De~Laroussilhe, Q., Gesmundo, A., Attariyan, M., and Gelly, S.
\newblock Parameter-efficient transfer learning for nlp.
\newblock In \emph{International Conference on Machine Learning}, pp.\  2790--2799. PMLR, 2019.

\bibitem[Hu et~al.(2022)Hu, Shen, Wallis, Allen-Zhu, Li, Wang, Wang, Chen, et~al.]{hu2021loralowrankadaptationlarge}
Hu, E.~J., Shen, Y., Wallis, P., Allen-Zhu, Z., Li, Y., Wang, S., Wang, L., Chen, W., et~al.
\newblock Lora: Low-rank adaptation of large language models.
\newblock \emph{ICLR}, 1\penalty0 (2):\penalty0 3, 2022.

\bibitem[Huang et~al.(2023)Huang, Tang, Zhang, Zhao, Song, Xia, and Wei]{huang2023languagescreatedequalllms}
Huang, H., Tang, T., Zhang, D., Zhao, X., Song, T., Xia, Y., and Wei, F.
\newblock Not all languages are created equal in {LLM}s: Improving multilingual capability by cross-lingual-thought prompting.
\newblock In Bouamor, H., Pino, J., and Bali, K. (eds.), \emph{Findings of the Association for Computational Linguistics: EMNLP 2023}, pp.\  12365--12394, Singapore, December 2023. Association for Computational Linguistics.
\newblock \doi{10.18653/v1/2023.findings-emnlp.826}.
\newblock URL \url{https://aclanthology.org/2023.findings-emnlp.826/}.

\bibitem[Joshi et~al.(2020)Joshi, Santy, Budhiraja, Bali, and Choudhury]{joshi-etal-2020-state}
Joshi, P., Santy, S., Budhiraja, A., Bali, K., and Choudhury, M.
\newblock The state and fate of linguistic diversity and inclusion in the {NLP} world.
\newblock In Jurafsky, D., Chai, J., Schluter, N., and Tetreault, J. (eds.), \emph{Proceedings of the 58th Annual Meeting of the Association for Computational Linguistics}, pp.\  6282--6293, Online, July 2020. Association for Computational Linguistics.
\newblock \doi{10.18653/v1/2020.acl-main.560}.
\newblock URL \url{https://aclanthology.org/2020.acl-main.560/}.

\bibitem[Kargaran et~al.(2024)Kargaran, Yvon, and Sch{\"u}tze]{kargaran2024glotcc}
Kargaran, A.~H., Yvon, F., and Sch{\"u}tze, H.
\newblock Glot{CC}: An open broad-coverage commoncrawl corpus and pipeline for minority languages.
\newblock \emph{Advances in Neural Information Processing Systems}, 2024.
\newblock URL \url{https://arxiv.org/abs/2410.23825}.

\bibitem[Karimi~Mahabadi et~al.(2021)Karimi~Mahabadi, Henderson, and Ruder]{mahabadi2021compacterefficientlowrankhypercomplex}
Karimi~Mahabadi, R., Henderson, J., and Ruder, S.
\newblock Compacter: Efficient low-rank hypercomplex adapter layers.
\newblock In Ranzato, M., Beygelzimer, A., Dauphin, Y., Liang, P., and Vaughan, J.~W. (eds.), \emph{Advances in Neural Information Processing Systems}, volume~34, pp.\  1022--1035. Curran Associates, Inc., 2021.
\newblock URL \url{https://proceedings.neurips.cc/paper_files/paper/2021/file/081be9fdff07f3bc808f935906ef70c0-Paper.pdf}.

\bibitem[Kirkpatrick et~al.(2017)Kirkpatrick, Pascanu, Rabinowitz, Veness, Desjardins, Rusu, Milan, Quan, Ramalho, Grabska-Barwinska, et~al.]{kirkpatrick2017overcoming}
Kirkpatrick, J., Pascanu, R., Rabinowitz, N., Veness, J., Desjardins, G., Rusu, A.~A., Milan, K., Quan, J., Ramalho, T., Grabska-Barwinska, A., et~al.
\newblock Overcoming catastrophic forgetting in neural networks.
\newblock \emph{Proceedings of the national academy of sciences}, 114\penalty0 (13):\penalty0 3521--3526, 2017.

\bibitem[Kojima et~al.(2024)Kojima, Okimura, Iwasawa, Yanaka, and Matsuo]{kojima2024multilingual}
Kojima, T., Okimura, I., Iwasawa, Y., Yanaka, H., and Matsuo, Y.
\newblock On the multilingual ability of decoder-based pre-trained language models: Finding and controlling language-specific neurons.
\newblock In Duh, K., Gomez, H., and Bethard, S. (eds.), \emph{Proceedings of the 2024 Conference of the North American Chapter of the Association for Computational Linguistics: Human Language Technologies (Volume 1: Long Papers)}, pp.\  6919--6971, Mexico City, Mexico, June 2024. Association for Computational Linguistics.
\newblock \doi{10.18653/v1/2024.naacl-long.384}.
\newblock URL \url{https://aclanthology.org/2024.naacl-long.384/}.

\bibitem[Lai et~al.(2024)Lai, Mesgar, and Fraser]{lai-etal-2024-llms}
Lai, W., Mesgar, M., and Fraser, A.
\newblock {LLM}s beyond {E}nglish: Scaling the multilingual capability of {LLM}s with cross-lingual feedback.
\newblock In Ku, L.-W., Martins, A., and Srikumar, V. (eds.), \emph{Findings of the Association for Computational Linguistics: ACL 2024}, pp.\  8186--8213, Bangkok, Thailand, August 2024. Association for Computational Linguistics.
\newblock \doi{10.18653/v1/2024.findings-acl.488}.
\newblock URL \url{https://aclanthology.org/2024.findings-acl.488/}.

\bibitem[Lee et~al.(2022{\natexlab{a}})Lee, Hwang, and Kim]{lee-etal-2022-fad}
Lee, J., Hwang, S.-w., and Kim, T.
\newblock {FAD}-{X}: Fusing adapters for cross-lingual transfer to low-resource languages.
\newblock In He, Y., Ji, H., Li, S., Liu, Y., and Chang, C.-H. (eds.), \emph{Proceedings of the 2nd Conference of the Asia-Pacific Chapter of the Association for Computational Linguistics and the 12th International Joint Conference on Natural Language Processing (Volume 2: Short Papers)}, pp.\  57--64, Online only, November 2022{\natexlab{a}}. Association for Computational Linguistics.
\newblock URL \url{https://aclanthology.org/2022.aacl-short.8}.

\bibitem[Lee et~al.(2022{\natexlab{b}})Lee, Ippolito, Nystrom, Zhang, Eck, Callison-Burch, and Carlini]{lee2021deduplicating}
Lee, K., Ippolito, D., Nystrom, A., Zhang, C., Eck, D., Callison-Burch, C., and Carlini, N.
\newblock Deduplicating training data makes language models better.
\newblock In Muresan, S., Nakov, P., and Villavicencio, A. (eds.), \emph{Proceedings of the 60th Annual Meeting of the Association for Computational Linguistics (Volume 1: Long Papers)}, pp.\  8424--8445, Dublin, Ireland, May 2022{\natexlab{b}}. Association for Computational Linguistics.
\newblock \doi{10.18653/v1/2022.acl-long.577}.
\newblock URL \url{https://aclanthology.org/2022.acl-long.577/}.

\bibitem[Liu et~al.(2022)Liu, Tam, Muqeeth, Mohta, Huang, Bansal, and Raffel]{liu2022fewshotparameterefficientfinetuningbetter}
Liu, H., Tam, D., Muqeeth, M., Mohta, J., Huang, T., Bansal, M., and Raffel, C.~A.
\newblock Few-shot parameter-efficient fine-tuning is better and cheaper than in-context learning.
\newblock \emph{Advances in Neural Information Processing Systems}, 35:\penalty0 1950--1965, 2022.

\bibitem[Liu et~al.(2024)Liu, Xu, Xu, Chen, Hu, and Wu]{liu-etal-2024-unraveling}
Liu, W., Xu, Y., Xu, H., Chen, J., Hu, X., and Wu, J.
\newblock Unraveling {B}abel: Exploring multilingual activation patterns of {LLM}s and their applications.
\newblock In Al-Onaizan, Y., Bansal, M., and Chen, Y.-N. (eds.), \emph{Proceedings of the 2024 Conference on Empirical Methods in Natural Language Processing}, pp.\  11855--11881, Miami, Florida, USA, November 2024. Association for Computational Linguistics.
\newblock \doi{10.18653/v1/2024.emnlp-main.662}.
\newblock URL \url{https://aclanthology.org/2024.emnlp-main.662/}.

\bibitem[Loshchilov \& Hutter(2019)Loshchilov and Hutter]{loshchilov2019decoupledweightdecayregularization}
Loshchilov, I. and Hutter, F.
\newblock Decoupled weight decay regularization, 2019.
\newblock URL \url{https://arxiv.org/abs/1711.05101}.

\bibitem[Meng et~al.(2022)Meng, Bau, Andonian, and Belinkov]{meng2022locating}
Meng, K., Bau, D., Andonian, A., and Belinkov, Y.
\newblock Locating and editing factual associations in gpt.
\newblock \emph{Advances in neural information processing systems}, 35:\penalty0 17359--17372, 2022.

\bibitem[Mikolov et~al.(2012)]{mikolov2012statistical}
Mikolov, T. et~al.
\newblock Statistical language models based on neural networks.
\newblock \emph{Presentation at Google, Mountain View, 2nd April}, 80\penalty0 (26), 2012.

\bibitem[{Mistral AI Team}(2024)]{mistral2024nemo}
{Mistral AI Team}.
\newblock Mistral nemo.
\newblock Technical report, Mistral AI, 2024.
\newblock URL \url{https://mistral.ai/news/mistral-nemo}.
\newblock Accessed: 2025.

\bibitem[Molchanov et~al.(2017)Molchanov, Tyree, Karras, Aila, and Kautz]{molchanov2017pruningconvolutionalneuralnetworks}
Molchanov, P., Tyree, S., Karras, T., Aila, T., and Kautz, J.
\newblock Pruning convolutional neural networks for resource efficient inference, 2017.
\newblock URL \url{https://arxiv.org/abs/1611.06440}.

\bibitem[Mondal et~al.(2025)Mondal, Sen, Singhania, and Jyothi]{mondal-etal-2025-language}
Mondal, S.~K., Sen, S., Singhania, A., and Jyothi, P.
\newblock Language-specific neurons do not facilitate cross-lingual transfer.
\newblock In Drozd, A., Sedoc, J., Tafreshi, S., Akula, A., and Shu, R. (eds.), \emph{The Sixth Workshop on Insights from Negative Results in NLP}, pp.\  46--62, Albuquerque, New Mexico, May 2025. Association for Computational Linguistics.
\newblock ISBN 979-8-89176-240-4.
\newblock \doi{10.18653/v1/2025.insights-1.6}.
\newblock URL \url{https://aclanthology.org/2025.insights-1.6/}.

\bibitem[Mosbach et~al.(2024)Mosbach, Gautam, Vergara~Browne, Klakow, and Geva]{mosbach2024insights}
Mosbach, M., Gautam, V., Vergara~Browne, T., Klakow, D., and Geva, M.
\newblock From insights to actions: The impact of interpretability and analysis research on {NLP}.
\newblock In Al-Onaizan, Y., Bansal, M., and Chen, Y.-N. (eds.), \emph{Proceedings of the 2024 Conference on Empirical Methods in Natural Language Processing}, pp.\  3078--3105, Miami, Florida, USA, November 2024. Association for Computational Linguistics.
\newblock \doi{10.18653/v1/2024.emnlp-main.181}.
\newblock URL \url{https://aclanthology.org/2024.emnlp-main.181/}.

\bibitem[Muller et~al.(2021)Muller, Anastasopoulos, Sagot, and Seddah]{muller-etal-2021-unseen}
Muller, B., Anastasopoulos, A., Sagot, B., and Seddah, D.
\newblock When being unseen from m{BERT} is just the beginning: Handling new languages with multilingual language models.
\newblock In Toutanova, K., Rumshisky, A., Zettlemoyer, L., Hakkani-Tur, D., Beltagy, I., Bethard, S., Cotterell, R., Chakraborty, T., and Zhou, Y. (eds.), \emph{Proceedings of the 2021 Conference of the North American Chapter of the Association for Computational Linguistics: Human Language Technologies}, pp.\  448--462, Online, June 2021. Association for Computational Linguistics.
\newblock \doi{10.18653/v1/2021.naacl-main.38}.
\newblock URL \url{https://aclanthology.org/2021.naacl-main.38}.

\bibitem[Neubig \& Hu(2018)Neubig and Hu]{neubig2018rapidadaptationneuralmachine}
Neubig, G. and Hu, J.
\newblock Rapid adaptation of neural machine translation to new languages.
\newblock In Riloff, E., Chiang, D., Hockenmaier, J., and Tsujii, J. (eds.), \emph{Proceedings of the 2018 Conference on Empirical Methods in Natural Language Processing}, pp.\  875--880, Brussels, Belgium, October-November 2018. Association for Computational Linguistics.
\newblock \doi{10.18653/v1/D18-1103}.
\newblock URL \url{https://aclanthology.org/D18-1103/}.

\bibitem[Parovi{\'c} et~al.(2022)Parovi{\'c}, Glava{\v{s}}, Vuli{\'c}, and Korhonen]{parovic-etal-2022-bad}
Parovi{\'c}, M., Glava{\v{s}}, G., Vuli{\'c}, I., and Korhonen, A.
\newblock {BAD}-{X}: Bilingual adapters improve zero-shot cross-lingual transfer.
\newblock In Carpuat, M., de~Marneffe, M.-C., and Meza~Ruiz, I.~V. (eds.), \emph{Proceedings of the 2022 Conference of the North American Chapter of the Association for Computational Linguistics: Human Language Technologies}, pp.\  1791--1799, Seattle, United States, July 2022. Association for Computational Linguistics.
\newblock \doi{10.18653/v1/2022.naacl-main.130}.
\newblock URL \url{https://aclanthology.org/2022.naacl-main.130}.

\bibitem[Parovic et~al.(2023)Parovic, Ansell, Vuli{\'c}, and Korhonen]{parović2023crosslingualtransfertargetlanguageready}
Parovic, M., Ansell, A., Vuli{\'c}, I., and Korhonen, A.
\newblock Cross-lingual transfer with target language-ready task adapters.
\newblock In Rogers, A., Boyd-Graber, J., and Okazaki, N. (eds.), \emph{Findings of the Association for Computational Linguistics: ACL 2023}, pp.\  176--193, Toronto, Canada, July 2023. Association for Computational Linguistics.
\newblock \doi{10.18653/v1/2023.findings-acl.13}.
\newblock URL \url{https://aclanthology.org/2023.findings-acl.13/}.

\bibitem[Pfeiffer et~al.(2020)Pfeiffer, Vuli{\'c}, Gurevych, and Ruder]{pfeiffer2020mad}
Pfeiffer, J., Vuli{\'c}, I., Gurevych, I., and Ruder, S.
\newblock {MAD-X}: {A}n {A}dapter-{B}ased {F}ramework for {M}ulti-{T}ask {C}ross-{L}ingual {T}ransfer.
\newblock In Webber, B., Cohn, T., He, Y., and Liu, Y. (eds.), \emph{Proceedings of the 2020 Conference on Empirical Methods in Natural Language Processing (EMNLP)}, pp.\  7654--7673, Online, November 2020. Association for Computational Linguistics.
\newblock \doi{10.18653/v1/2020.emnlp-main.617}.
\newblock URL \url{https://aclanthology.org/2020.emnlp-main.617}.

\bibitem[Post(2018)]{post-2018-call}
Post, M.
\newblock A call for clarity in reporting {BLEU} scores.
\newblock In \emph{Proceedings of the Third Conference on Machine Translation: Research Papers}, pp.\  186--191, Belgium, Brussels, October 2018. Association for Computational Linguistics.
\newblock URL \url{https://www.aclweb.org/anthology/W18-6319}.

\bibitem[Robinson et~al.(2023)Robinson, Ogayo, Mortensen, and Neubig]{robinson2023chatgptmtcompetitivehigh}
Robinson, N., Ogayo, P., Mortensen, D.~R., and Neubig, G.
\newblock {C}hat{GPT} {MT}: Competitive for high- (but not low-) resource languages.
\newblock In Koehn, P., Haddow, B., Kocmi, T., and Monz, C. (eds.), \emph{Proceedings of the Eighth Conference on Machine Translation}, pp.\  392--418, Singapore, December 2023. Association for Computational Linguistics.
\newblock \doi{10.18653/v1/2023.wmt-1.40}.
\newblock URL \url{https://aclanthology.org/2023.wmt-1.40/}.

\bibitem[Sakaguchi et~al.(2019)Sakaguchi, Bras, Bhagavatula, and Choi]{sakaguchi2019winograndeadversarialwinogradschema}
Sakaguchi, K., Bras, R.~L., Bhagavatula, C., and Choi, Y.
\newblock Winogrande: An adversarial winograd schema challenge at scale, 2019.
\newblock URL \url{https://arxiv.org/abs/1907.10641}.

\bibitem[Sanh et~al.(2020)Sanh, Wolf, and Rush]{sanh2020movement}
Sanh, V., Wolf, T., and Rush, A.
\newblock Movement pruning: Adaptive sparsity by fine-tuning.
\newblock \emph{Advances in neural information processing systems}, 33:\penalty0 20378--20389, 2020.

\bibitem[Schulman \& Lab(2025)Schulman and Lab]{schulman2025lora}
Schulman, J. and Lab, T.~M.
\newblock Lora without regret.
\newblock \emph{Thinking Machines Lab: Connectionism}, 2025.
\newblock \doi{10.64434/tml.20250929}.
\newblock https://thinkingmachines.ai/blog/lora/.

\bibitem[Shannon \& Weaver(1998)Shannon and Weaver]{shannon1998mathematical}
Shannon, C.~E. and Weaver, W.
\newblock \emph{The mathematical theory of communication}.
\newblock University of Illinois press, 1998.

\bibitem[Shazeer(2020)]{shazeer2020glu}
Shazeer, N.
\newblock Glu variants improve transformer.
\newblock \emph{arXiv preprint arXiv:2002.05202}, 2020.

\bibitem[Strubell et~al.(2019)Strubell, Ganesh, and McCallum]{strubell-etal-2019-energy}
Strubell, E., Ganesh, A., and McCallum, A.
\newblock Energy and policy considerations for deep learning in {NLP}.
\newblock In Korhonen, A., Traum, D., and M{\`a}rquez, L. (eds.), \emph{Proceedings of the 57th Annual Meeting of the Association for Computational Linguistics}, pp.\  3645--3650, Florence, Italy, July 2019. Association for Computational Linguistics.
\newblock \doi{10.18653/v1/P19-1355}.
\newblock URL \url{https://aclanthology.org/P19-1355}.

\bibitem[Sundar et~al.(2025)Sundar, Williamson, Metcalf, Theobald, Seto, and Fedzechkina]{sundar2025steeringnewembeddingspaces}
Sundar, A., Williamson, S., Metcalf, K., Theobald, B.-J., Seto, S., and Fedzechkina, M.
\newblock Steering into new embedding spaces: Analyzing cross-lingual alignment induced by model interventions in multilingual language models, 2025.
\newblock URL \url{https://arxiv.org/abs/2502.15639}.

\bibitem[Tang et~al.(2024)Tang, Luo, Huang, Zhang, Wang, Zhao, Wei, and Wen]{tang2024language}
Tang, T., Luo, W., Huang, H., Zhang, D., Wang, X., Zhao, X., Wei, F., and Wen, J.-R.
\newblock Language-specific neurons: The key to multilingual capabilities in large language models.
\newblock In Ku, L.-W., Martins, A., and Srikumar, V. (eds.), \emph{Proceedings of the 62nd Annual Meeting of the Association for Computational Linguistics (Volume 1: Long Papers)}, pp.\  5701--5715, Bangkok, Thailand, August 2024. Association for Computational Linguistics.
\newblock \doi{10.18653/v1/2024.acl-long.309}.
\newblock URL \url{https://aclanthology.org/2024.acl-long.309/}.

\bibitem[{\"U}st{\"u}n et~al.(2020){\"U}st{\"u}n, Bisazza, Bouma, and van Noord]{udapterlanguageadaptationtruly}
{\"U}st{\"u}n, A., Bisazza, A., Bouma, G., and van Noord, G.
\newblock {UD}apter: Language adaptation for truly {U}niversal {D}ependency parsing.
\newblock In Webber, B., Cohn, T., He, Y., and Liu, Y. (eds.), \emph{Proceedings of the 2020 Conference on Empirical Methods in Natural Language Processing (EMNLP)}, pp.\  2302--2315, Online, November 2020. Association for Computational Linguistics.
\newblock \doi{10.18653/v1/2020.emnlp-main.180}.
\newblock URL \url{https://aclanthology.org/2020.emnlp-main.180/}.

\bibitem[Voita et~al.(2019)Voita, Talbot, Moiseev, Sennrich, and Titov]{voita-etal-2019-analyzing}
Voita, E., Talbot, D., Moiseev, F., Sennrich, R., and Titov, I.
\newblock Analyzing multi-head self-attention: Specialized heads do the heavy lifting, the rest can be pruned.
\newblock In Korhonen, A., Traum, D., and M{\`a}rquez, L. (eds.), \emph{Proceedings of the 57th Annual Meeting of the Association for Computational Linguistics}, pp.\  5797--5808, Florence, Italy, July 2019. Association for Computational Linguistics.
\newblock \doi{10.18653/v1/P19-1580}.
\newblock URL \url{https://aclanthology.org/P19-1580/}.

\bibitem[Vykopal et~al.(2025)Vykopal, Ostermann, and Simko]{vykopal-etal-2025-soft}
Vykopal, I., Ostermann, S., and Simko, M.
\newblock Soft language prompts for language transfer.
\newblock In Chiruzzo, L., Ritter, A., and Wang, L. (eds.), \emph{Proceedings of the 2025 Conference of the Nations of the Americas Chapter of the Association for Computational Linguistics: Human Language Technologies (Volume 1: Long Papers)}, pp.\  10294--10313, Albuquerque, New Mexico, April 2025. Association for Computational Linguistics.
\newblock ISBN 979-8-89176-189-6.
\newblock \doi{10.18653/v1/2025.naacl-long.517}.
\newblock URL \url{https://aclanthology.org/2025.naacl-long.517/}.

\bibitem[Wang et~al.(2025)Wang, Haddow, Wu, Peng, and Birch]{wang2025sharingmattersanalysingneurons}
Wang, W., Haddow, B., Wu, M., Peng, W., and Birch, A.
\newblock Sharing matters: Analysing neurons across languages and tasks in llms, 2025.
\newblock URL \url{https://arxiv.org/abs/2406.09265}.

\bibitem[Wang et~al.(2022)Wang, Wen, Zhang, Hou, Liu, and Li]{wang2022finding}
Wang, X., Wen, K., Zhang, Z., Hou, L., Liu, Z., and Li, J.
\newblock Finding skill neurons in pre-trained transformer-based language models.
\newblock In Goldberg, Y., Kozareva, Z., and Zhang, Y. (eds.), \emph{Proceedings of the 2022 Conference on Empirical Methods in Natural Language Processing}, pp.\  11132--11152, Abu Dhabi, United Arab Emirates, December 2022. Association for Computational Linguistics.
\newblock \doi{10.18653/v1/2022.emnlp-main.765}.
\newblock URL \url{https://aclanthology.org/2022.emnlp-main.765/}.

\bibitem[Xu et~al.(2025)Xu, Zhan, Ma, Wong, and Chao]{xu2024let}
Xu, H., Zhan, R., Ma, Y., Wong, D.~F., and Chao, L.~S.
\newblock Let{'}s focus on neuron: Neuron-level supervised fine-tuning for large language model.
\newblock In Rambow, O., Wanner, L., Apidianaki, M., Al-Khalifa, H., Eugenio, B.~D., and Schockaert, S. (eds.), \emph{Proceedings of the 31st International Conference on Computational Linguistics}, pp.\  9393--9406, Abu Dhabi, UAE, January 2025. Association for Computational Linguistics.
\newblock URL \url{https://aclanthology.org/2025.coling-main.630/}.

\bibitem[Yong et~al.(2023)Yong, Schoelkopf, Muennighoff, Aji, Adelani, Almubarak, Bari, Sutawika, Kasai, Baruwa, Winata, Biderman, Raff, Radev, and Nikoulina]{yong2023bloom1addinglanguagesupport}
Yong, Z.~X., Schoelkopf, H., Muennighoff, N., Aji, A.~F., Adelani, D.~I., Almubarak, K., Bari, M.~S., Sutawika, L., Kasai, J., Baruwa, A., Winata, G., Biderman, S., Raff, E., Radev, D., and Nikoulina, V.
\newblock {BLOOM}+1: Adding language support to {BLOOM} for zero-shot prompting.
\newblock In Rogers, A., Boyd-Graber, J., and Okazaki, N. (eds.), \emph{Proceedings of the 61st Annual Meeting of the Association for Computational Linguistics (Volume 1: Long Papers)}, pp.\  11682--11703, Toronto, Canada, July 2023. Association for Computational Linguistics.
\newblock \doi{10.18653/v1/2023.acl-long.653}.
\newblock URL \url{https://aclanthology.org/2023.acl-long.653/}.

\bibitem[Zellers et~al.(2019)Zellers, Holtzman, Bisk, Farhadi, and Choi]{zellers2019hellaswagmachinereallyfinish}
Zellers, R., Holtzman, A., Bisk, Y., Farhadi, A., and Choi, Y.
\newblock Hellaswag: Can a machine really finish your sentence?, 2019.
\newblock URL \url{https://arxiv.org/abs/1905.07830}.

\bibitem[Zhang et~al.(2023)Zhang, Chen, Bukharin, Karampatziakis, He, Cheng, Chen, and Zhao]{zhang2023adaloraadaptivebudgetallocation}
Zhang, Q., Chen, M., Bukharin, A., Karampatziakis, N., He, P., Cheng, Y., Chen, W., and Zhao, T.
\newblock Adalora: Adaptive budget allocation for parameter-efficient fine-tuning, 2023.
\newblock URL \url{https://arxiv.org/abs/2303.10512}.

\bibitem[Zhao et~al.(2024)Zhao, Zhang, Chen, Kawaguchi, and Bing]{zhao2024largelanguagemodelshandle}
Zhao, Y., Zhang, W., Chen, G., Kawaguchi, K., and Bing, L.
\newblock How do large language models handle multilingualism?
\newblock In Globerson, A., Mackey, L., Belgrave, D., Fan, A., Paquet, U., Tomczak, J., and Zhang, C. (eds.), \emph{Advances in Neural Information Processing Systems}, volume~37, pp.\  15296--15319. Curran Associates, Inc., 2024.
\newblock \doi{10.52202/079017-0489}.
\newblock URL \url{https://proceedings.neurips.cc/paper_files/paper/2024/file/1bd359b32ab8b2a6bbafa1ed2856cf40-Paper-Conference.pdf}.

\bibitem[Zhu et~al.(2024)Zhu, Pan, Li, and Xiong]{zhu2024landermt}
Zhu, S., Pan, L., Li, B., and Xiong, D.
\newblock {LAND}e{RMT}: Dectecting and routing language-aware neurons for selectively finetuning {LLM}s to machine translation.
\newblock In Ku, L.-W., Martins, A., and Srikumar, V. (eds.), \emph{Proceedings of the 62nd Annual Meeting of the Association for Computational Linguistics (Volume 1: Long Papers)}, pp.\  12135--12148, Bangkok, Thailand, August 2024. Association for Computational Linguistics.
\newblock \doi{10.18653/v1/2024.acl-long.656}.
\newblock URL \url{https://aclanthology.org/2024.acl-long.656/}.

\bibitem[Zhu et~al.(2025)Zhu, Gong, Xiao, Liu, and Hoiem]{zhu2025teach}
Zhu, Z., Gong, Y., Xiao, Y., Liu, Y., and Hoiem, D.
\newblock How to teach large multimodal models new skills.
\newblock \emph{arXiv preprint arXiv:2510.08564}, 2025.

\end{thebibliography}
\bibliographystyle{icml2026}

\newpage
\appendix
\onecolumn

\section*{Appendix}
\section{Limitations}
\label{app:limitations}

While our results show that sparse, language-specific subnetwork fine-tuning is effective across multiple models and languages, several limitations remain.

First, the effectiveness of targeted subnetwork adaptation depends on accurately identifying language-specific neurons. This process can be sensitive to the quality, coverage, and representativeness of the monolingual data used for neuron identification, particularly for languages with limited or noisy corpora.

Second, our experiments primarily focus on mid- and low-resource languages with sufficient post-training data. How well the method scales to extremely low-resource settings or to languages that are highly distant from those seen during pre-training remains an open question.

Third, we adopt a fixed selection threshold and do not systematically vary the capacity of language-specific subnetworks (i.e., the parameter $\rho$). Our results suggest that different languages may require different amounts of capacity to effectively capture language-specific knowledge, and that the interaction between subnetwork size and available training data warrants further study, potentially at the per-language or per-language-family level.

Finally, due to computational constraints, we did not run Full fine-tuning, FFN-only, LoRA, or IA$^3$ baselines for \textsc{Mistral-Nemo-12B} and \textsc{Aya-Expanse-8B}. Based on our observations with Llama-3.1-8B, we expect these dense and PEFT methods to exhibit similar behavior on these models, but formal verification remains future work.

\clearpage

\begin{figure*}[t]
\section{Language Neuron Details}
\label{app:lang_details}
\subsection{Language Subnetworks}
\label{app:subnetworks}

\noindent
Figures~\ref{fig:neuron_dist_nemo} and~\ref{fig:neuron_dist_aya} illustrate how language-specific neurons are distributed across layers for each model. While the absolute number of selected neurons varies substantially across languages and architectures, consistent structural patterns emerge. In particular, subnetworks tend to be sparse in early layers and increasingly concentrated in later layers, suggesting that language-specific specialization primarily occurs at higher levels of abstraction. Differences in subnetwork size across languages likely reflect a combination of pre-training exposure, typological similarity to high-resource languages, and model-specific capacity allocation.

\vspace{25pt}
    \centering
    \begin{minipage}[t!]{0.48\linewidth}
        \centering
        \includegraphics[width=0.9\linewidth]{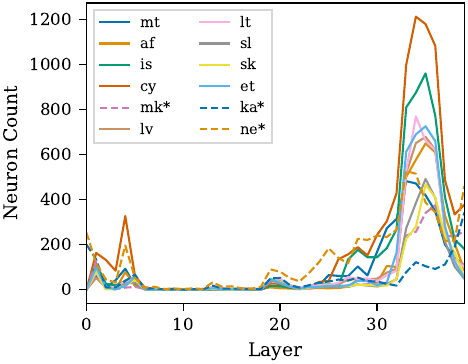}
        \caption{Neuron distributions over the layers of \textsc{Mistral-Nemo-12B} for all 12 languages. Asterisks indicate languages with non-Latin scripts.}
        \label{fig:neuron_dist_nemo}
    \end{minipage}%
    \hfill
    \begin{minipage}[t!]{0.48\linewidth}
        \centering
        \small
        \begin{sc}
        \begin{tabular}{@{}lccc@{}}
        \toprule
        Lang. & \# Neurons & \# Trainable & \% \\
        \midrule
        Afrikaans   & 3557 & 54.6M & 0.45 \\ 
        Welsh       & 8239 & 126.6M & 1.03 \\ 
        Estonian    & 3815 & 58.6M & 0.48 \\ 
        Icelandic   & 5644 & 86.7M & 0.71 \\ 
        Georgian    & 2028 & 31.1M & 0.25 \\ 
        Lithuanian  & 3840 & 59.0M & 0.48 \\ 
        Latvian     & 3741 & 57.5M & 0.47 \\ 
        Macedonian  & 2403 & 36.9M & 0.30 \\
        Maltese     & 3611 & 55.5M & 0.45 \\ 
        Nepali      & 5567 & 85.5M & 0.70 \\ 
        Slovak      & 2310 & 35.5M & 0.29 \\ 
        Slovenian   & 2365 & 36.3M & 0.30 \\ 
        \bottomrule
        \addlinespace[10pt]
        \end{tabular}
        \end{sc}
        \caption{Neuron and parameter counts for different language subnetworks, and percentages of the model being trained within \textsc{Mistral-Nemo-12B}.}
        \label{tab:neurons_params_nemo}
    \end{minipage}
\end{figure*}

\begin{figure*}[t]
    \centering
    \begin{minipage}[t!]{0.52\linewidth}
        \centering
        \includegraphics[width=0.9\linewidth]{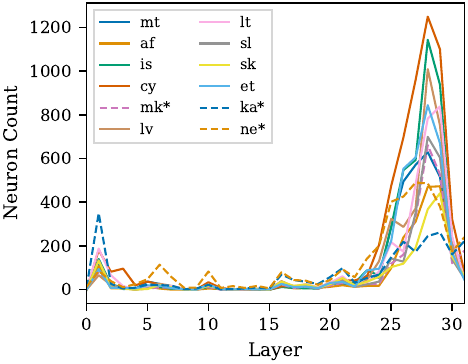}
        \caption{Neuron distributions over the layers of \textsc{Aya-Expanse-8B} for all 12 languages. Asterisks indicate languages with non-Latin scripts.}
        \label{fig:neuron_dist_aya}
    \end{minipage}%
    \hfill
    \begin{minipage}[t!]{0.44\linewidth}
        \centering
        \small
        \begin{sc}
        \begin{tabular}{@{}lccc@{}}
        \toprule
        Lang. & \# Neurons & \# Trainable & \% \\
        \midrule
        Afrikaans   & 2091 & 25.7M & 0.32 \\ 
        Welsh       & 5820 & 71.5M & 0.88 \\ 
        Estonian    & 3628 & 44.6M & 0.55 \\ 
        Icelandic   & 4303 & 52.9M & 0.65 \\ 
        Georgian    & 2432 & 29.9M & 0.37 \\ 
        Lithuanian  & 3443 & 42.3M & 0.52 \\ 
        Latvian     & 3536 & 43.5M & 0.54 \\ 
        Macedonian  & 2439 & 30.0M & 0.37 \\
        Maltese     & 3201 & 39.3M & 0.49 \\ 
        Nepali      & 3870 & 47.6M & 0.59 \\ 
        Slovak      & 1924 & 23.6M & 0.29 \\ 
        Slovenian   & 2538 & 31.2M & 0.39 \\ 
        \bottomrule
        \addlinespace[10pt]
        \end{tabular}
        \end{sc}
        \caption{Neuron and parameter counts for different language subnetworks, and percentages of the model being trained within \textsc{Aya-Expanse-8B}.}
        \label{tab:neurons_params_aya}
    \end{minipage}
\end{figure*}

\begin{figure*}[t]
\subsection{Language Neuron Overlap}
\label{app:overlap}

\noindent
Figure~\ref{fig:neuron_overlap} shows pairwise overlap between language-specific subnetworks, grouped by language family. Across models, related languages consistently exhibit higher overlap than unrelated ones, indicating partial sharing of language-relevant components rather than strictly disjoint representations.

\vspace{25pt}
    \centering

    \begin{minipage}[t]{0.48\linewidth}
        \centering
        \includegraphics[width=\linewidth]{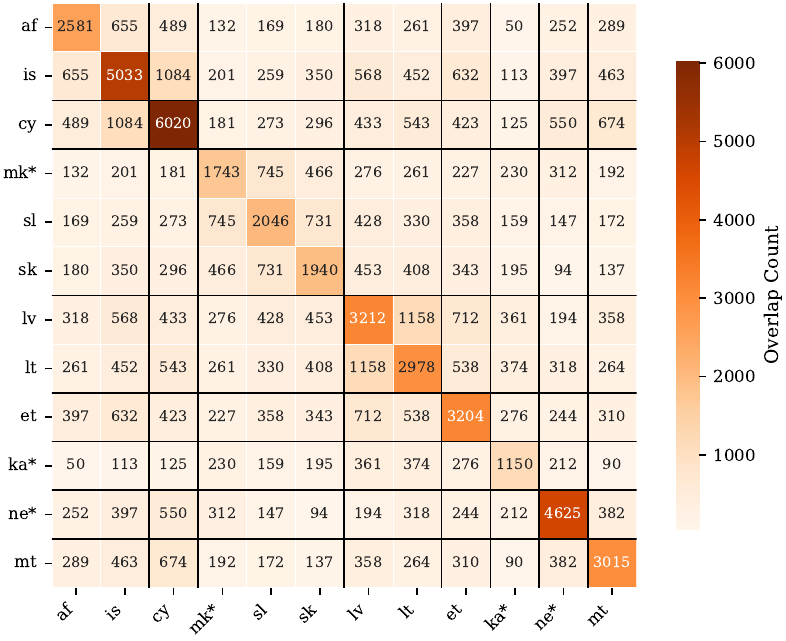}
        \caption*{(a) \textsc{Llama-3.1-8B}}
    \end{minipage}
    \hfill
    \begin{minipage}[t]{0.48\linewidth}
        \centering
        \includegraphics[width=\linewidth]{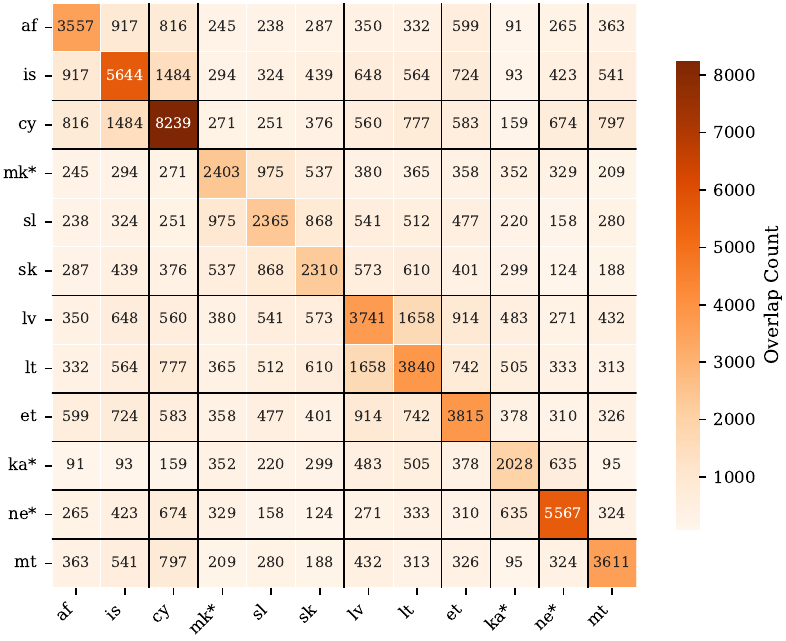}
        \caption*{(b) \textsc{Mistral-Nemo-12B}}
    \end{minipage}

    \vspace{1em}

    \begin{minipage}[t]{0.48\linewidth}
        \centering
        \includegraphics[width=\linewidth]{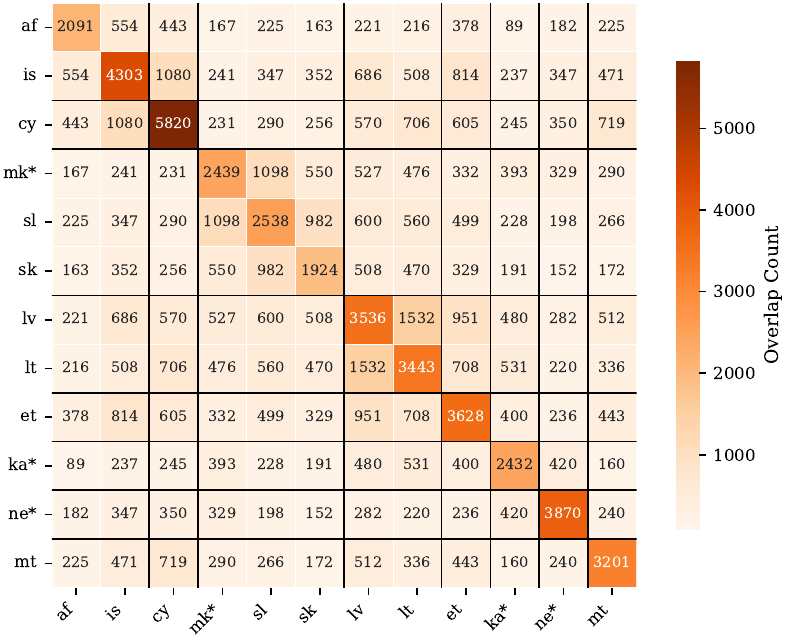}
        \caption*{(c) \textsc{Aya-Expanse-8B}}
    \end{minipage}

    \caption{Language neuron overlap for all 12 languages across models.}
    \label{fig:neuron_overlap}
\end{figure*}

\begin{figure*}[t]
\subsection{Language Neuron Distributions}
\noindent
Below, we provide per-language distributions of identified language-specific neurons across layers for each model.

\vspace{25pt}
    \centering

    \begin{minipage}[t]{0.48\linewidth}
        \centering
        \includegraphics[width=\linewidth]{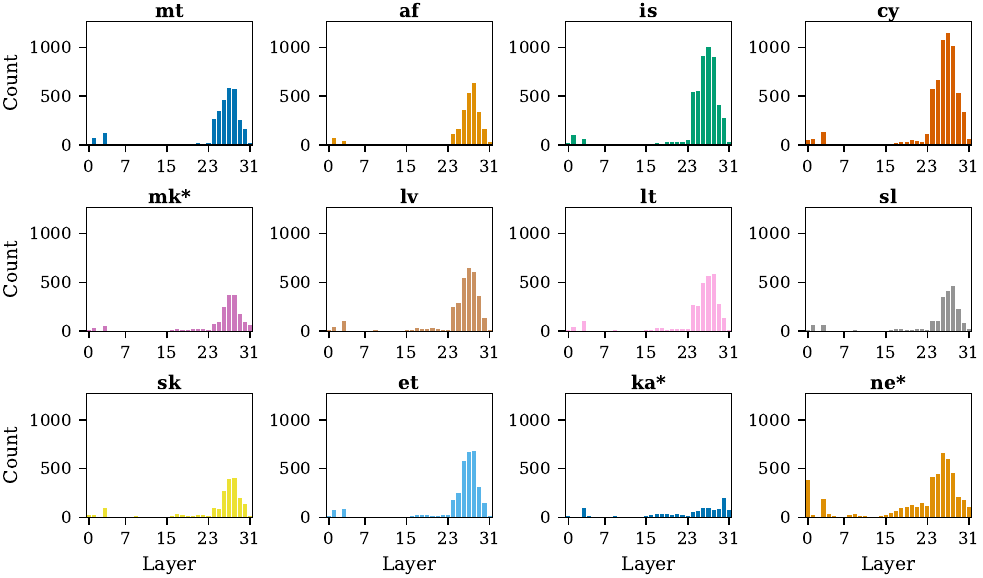}
        \caption*{(a) \textsc{Llama-3.1-8B}}
    \end{minipage}
    \hfill
    \begin{minipage}[t]{0.48\linewidth}
        \centering
        \includegraphics[width=\linewidth]{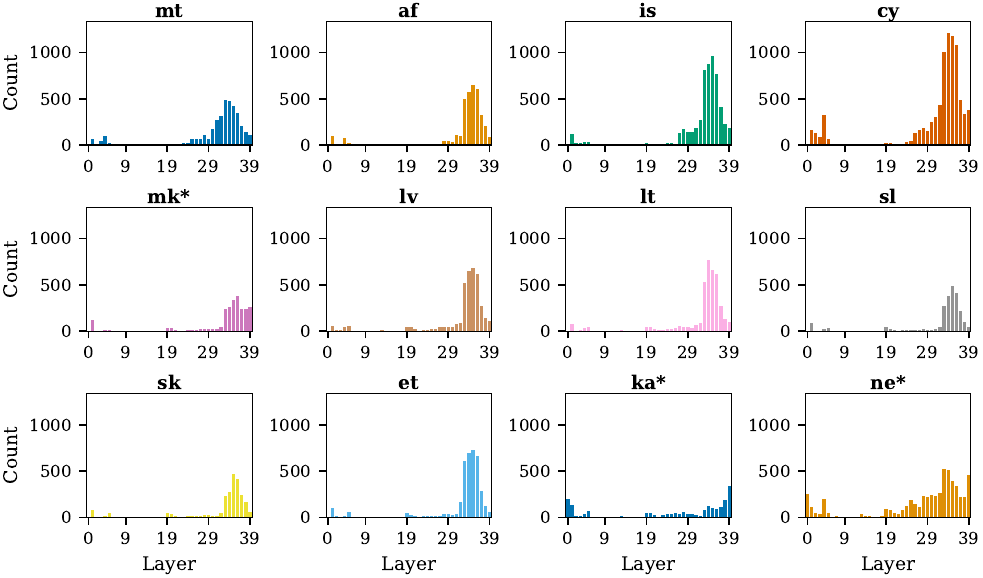}
        \caption*{(b) \textsc{Mistral-Nemo-12B}}
    \end{minipage}

    \vspace{1em}

    \begin{minipage}[t]{0.48\linewidth}
        \centering
        \includegraphics[width=\linewidth]{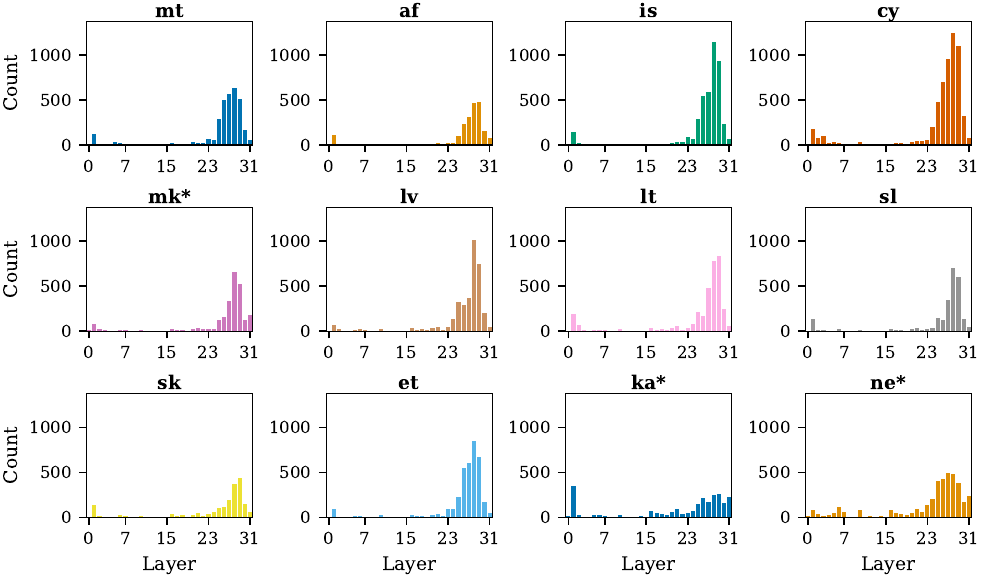}
        \caption*{(c) \textsc{Aya-Expanse-8B}}
    \end{minipage}

    \caption{Language neuron distributions for all 12 languages across models.}
    \label{fig:neuron_dists}
\end{figure*}

\clearpage

\begin{figure*}[t]
\section{Training Dynamics}
\label{app:val_losses}

\noindent
This section presents validation loss curves for all 12 target languages, illustrating training dynamics across fine-tuning strategies.

    \centering

    \begin{subfigure}[t]{0.32\linewidth}
        \centering
        \includegraphics[width=\linewidth]{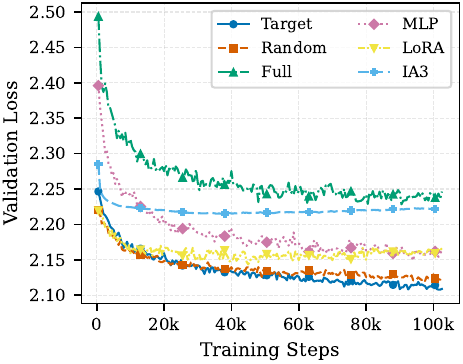}
        \caption{Afrikaans}
        \label{fig:curves_af}
    \end{subfigure}
    \hfill
    \begin{subfigure}[t]{0.32\linewidth}
        \centering
        \includegraphics[width=\linewidth]{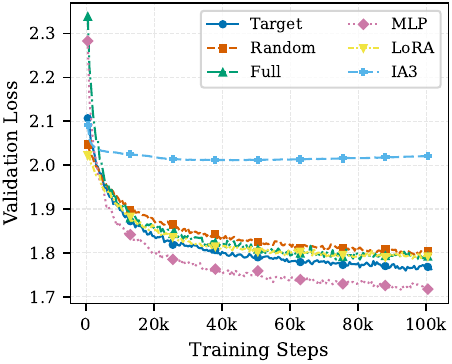}
        \caption{Welsh}
        \label{fig:curves_cy}
    \end{subfigure}
    \hfill
    \begin{subfigure}[t]{0.32\linewidth}
        \centering
        \includegraphics[width=\linewidth]{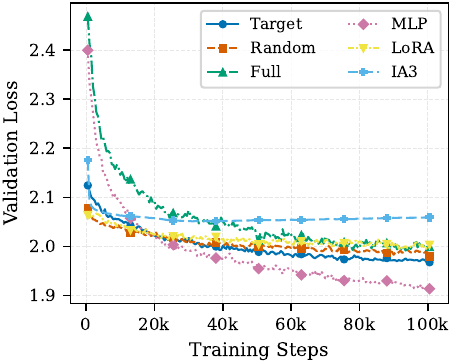}
        \caption{Estonian}
        \label{fig:curves_et}
    \end{subfigure}

    \vspace{0.8em}

    \begin{subfigure}[t]{0.32\linewidth}
        \centering
        \includegraphics[width=\linewidth]{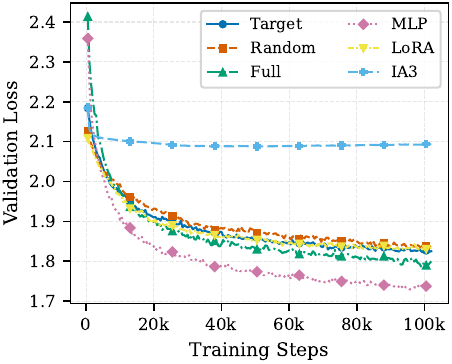}
        \caption{Icelandic}
        \label{fig:curves_is}
    \end{subfigure}
    \hfill
    \begin{subfigure}[t]{0.32\linewidth}
        \centering
        \includegraphics[width=\linewidth]{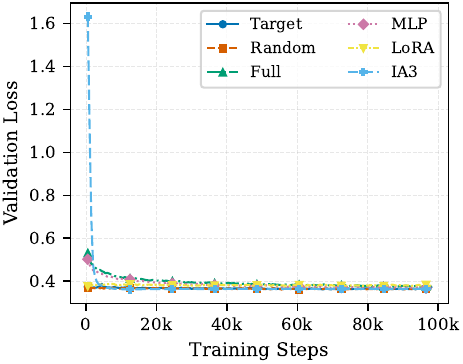}
        \caption{Georgian}
        \label{fig:curves_ka}
    \end{subfigure}
    \hfill
    \begin{subfigure}[t]{0.32\linewidth}
        \centering
        \includegraphics[width=\linewidth]{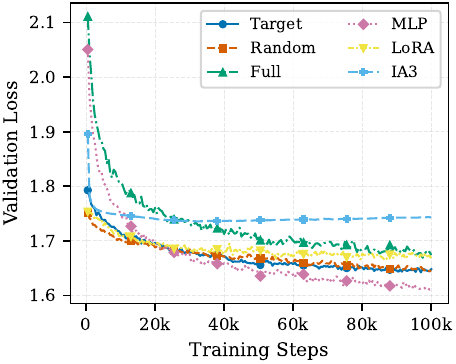}
        \caption{Lithuanian}
        \label{fig:curves_lt}
    \end{subfigure}

    \begin{subfigure}[t]{0.32\linewidth}
        \centering
        \includegraphics[width=\linewidth]{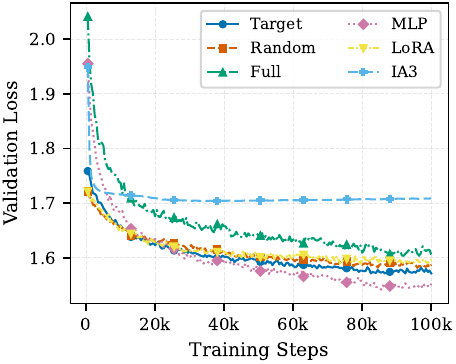}
        \caption{Latvian}
        \label{fig:curves_lv}
    \end{subfigure}
    \hfill
    \begin{subfigure}[t]{0.32\linewidth}
        \centering
        \includegraphics[width=\linewidth]{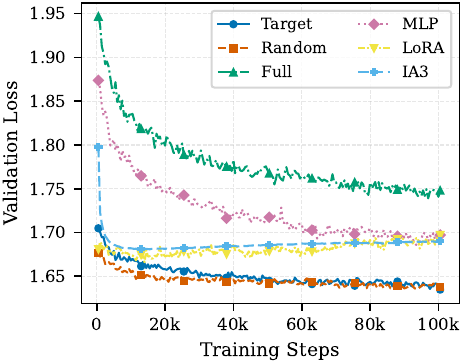}
        \caption{Macedonian}
        \label{fig:curves_mk}
    \end{subfigure}
    \hfill
    \begin{subfigure}[t]{0.32\linewidth}
        \centering
        \includegraphics[width=\linewidth]{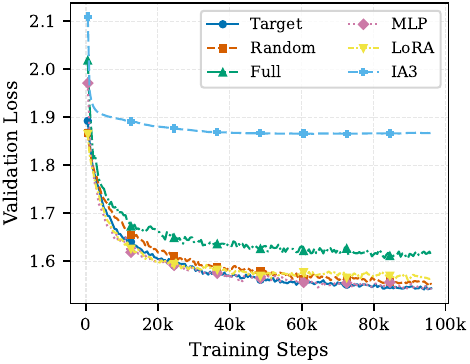}
        \caption{Maltese}
        \label{fig:curves_mt}
    \end{subfigure}

    \vspace{0.8em}

    \begin{subfigure}[t]{0.32\linewidth}
        \centering
        \includegraphics[width=\linewidth]{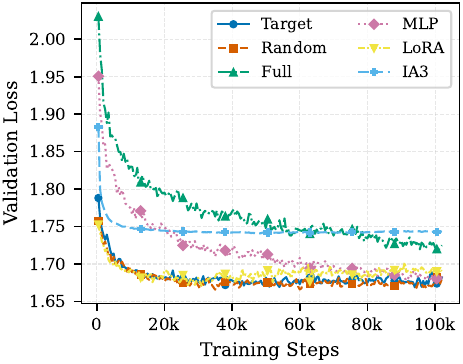}
        \caption{Nepali}
        \label{fig:curves_ne}
    \end{subfigure}
    \hfill
    \begin{subfigure}[t]{0.32\linewidth}
        \centering
        \includegraphics[width=\linewidth]{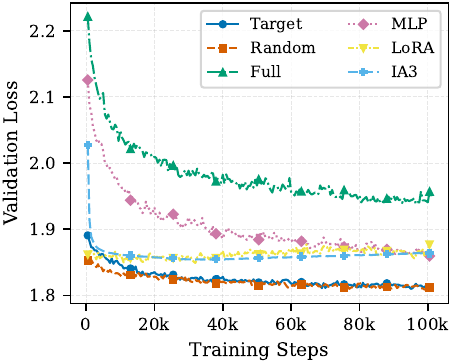}
        \caption{Slovak}
        \label{fig:curves_sk}
    \end{subfigure}
    \hfill
    \begin{subfigure}[t]{0.32\linewidth}
        \centering
        \includegraphics[width=\linewidth]{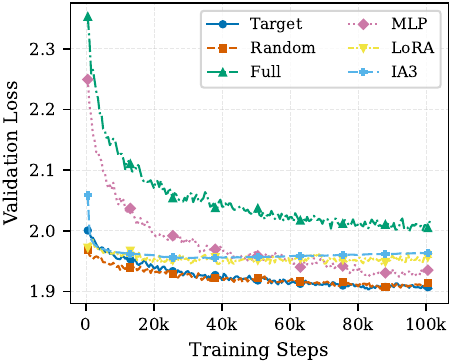}
        \caption{Slovenian}
        \label{fig:curves_sl}
    \end{subfigure}

    \caption{Validation loss curves for \textsc{Llama-3.1-8B} across languages.}
\end{figure*}

\clearpage

\begin{table*}[h!]
\section{Downstream Results}
\label{app:downstream}

\noindent
In this appendix, we provide detailed per-language and per-task performance results for all models, complementing the averaged results reported in the main text.

\subsection{Llama-3.1-8B}
\vspace{10pt}
\small
\centering
\begin{sc}
\begin{tabular}{@{}lccccccc@{}}
\toprule
Metric & \textbf{Base} & \textbf{Full FT} & \textbf{FFN-Only FT} & \textbf{LoRA} & \textbf{IA$^3$} & \textbf{Random} & \textbf{Target} \\
\midrule
\multicolumn{8}{c}{\textbf{Monolingual Capabilities}} \\
\midrule
FLORES af$\rightarrow$en & 45.88 & 17.35 & 45.58 & 48.16 & 47.19 & \cellcolor{lightgreen}49.34 & \cellcolor{mediumgreen}49.43 \\
FLORES en$\rightarrow$af & 28.19 & 20.91 & 31.96 & 30.97 & 28.65 & \cellcolor{mediumgreen}32.52 & \cellcolor{lightgreen}32.11 \\
BELEBELE af & 72.11 & 31.22 & 47.67 & 65.00 & \cellcolor{mediumgreen}73.33 & 68.89 & \cellcolor{lightgreen}73.11 \\
SIB200 af & 35.29 & 27.45 & 30.39 & 31.37 & 42.16 & \cellcolor{mediumgreen}45.59 & \cellcolor{lightgreen}44.12 \\
OPUS af$\rightarrow$en & 25.49 & 6.62 & 25.96 & 34.13 & 29.81 & \cellcolor{lightgreen}34.18 & \cellcolor{mediumgreen}34.66 \\
OPUS en$\rightarrow$af & 21.06 & 7.25 & 14.83 & 20.76 & 20.94 & \cellcolor{lightgreen}21.63 & \cellcolor{mediumgreen}24.50 \\
\end{tabular}
\end{sc}
\caption{Evaluation results comparing different fine-tuning configurations across various tasks for \textbf{Afrikaans} for \textsc{Llama-3.1-8B}.}
\label{tab:evaluation_results_af}
\end{table*}

\begin{table*}[h!]
\small
\centering
\begin{sc}
\begin{tabular}{@{}lccccccc@{}}
\toprule
Metric & \textbf{Base} & \textbf{Full FT} & \textbf{FFN-Only FT} & \textbf{LoRA} & \textbf{IA$^3$} & \textbf{Random} & \textbf{Target} \\
\midrule
\multicolumn{8}{c}{\textbf{Monolingual Capabilities}} \\
\midrule
FLORES cy$\rightarrow$en & 39.81 & 32.37 & 40.48 & 39.82 & 40.30 & \cellcolor{lightgreen}42.40 & \cellcolor{mediumgreen}43.90 \\
FLORES en$\rightarrow$cy & 14.78 & 23.90 & \cellcolor{mediumgreen}29.11 & 23.46 & 14.83 & 21.77 & \cellcolor{lightgreen}25.57 \\
SIB200 cy & 37.25 & 1.47 & 25.49 & \cellcolor{mediumgreen}49.51 & \cellcolor{lightgreen}38.24 & 20.10 & 35.78 \\
OPUS cy$\rightarrow$en & 18.16 & 23.03 & 26.13 & \cellcolor{lightgreen}26.82 & 22.07 & \cellcolor{mediumgreen}29.74 & 23.65 \\
OPUS en$\rightarrow$cy & \cellcolor{lightgreen}17.16 & 8.96 & 13.80 & 17.05 & 14.33 & \cellcolor{mediumgreen}19.04 & 14.36 \\
\hdashline
Avg. & 25.43 & 17.95 & 27.00 & \cellcolor{mediumgreen}31.33 & 25.95 & 26.61 & \cellcolor{lightgreen}28.65 \\
\midrule
\multicolumn{8}{c}{\textbf{General Performance}} \\
\midrule
arc challenge & \cellcolor{mediumgreen}70.90 & 16.05 & 22.74 & 56.52 & \cellcolor{lightgreen}69.57 & 54.52 & 68.23 \\
arc easy & \cellcolor{mediumgreen}85.79 & 16.84 & 30.70 & 74.56 & \cellcolor{lightgreen}85.61 & 76.14 & 84.04 \\
hellaswag & \cellcolor{lightgreen}40.38 & 12.31 & 25.27 & 29.93 & 38.66 & 30.88 & \cellcolor{mediumgreen}40.79 \\
piqa & \cellcolor{lightgreen}71.87 & 3.97 & 49.46 & 58.98 & \cellcolor{mediumgreen}73.72 & 66.97 & 68.39 \\
winogrande & \cellcolor{mediumgreen}51.62 & 20.28 & 49.57 & 49.17 & 50.20 & \cellcolor{lightgreen}51.46 & 50.51 \\
mmlu & \cellcolor{lightgreen}58.07 & 22.27 & 25.67 & 45.79 & \cellcolor{mediumgreen}58.26 & 48.14 & 56.69 \\
\hdashline
Avg. & \cellcolor{mediumgreen}63.10 & 15.29 & 33.90 & 52.49 & \cellcolor{lightgreen}62.67 & 54.68 & 61.44 \\
\bottomrule
\addlinespace[10pt]
\end{tabular}
\end{sc}
\caption{Evaluation results comparing different fine-tuning configurations across various tasks for \textbf{Welsh} for \textsc{Llama-3.1-8B}.}
\label{tab:evaluation_results_cy}
\end{table*}

\begin{table*}[h!]
\small
\centering
\begin{sc}
\begin{tabular}{@{}lccccccc@{}}
\toprule
Metric & \textbf{Base} & \textbf{Full FT} & \textbf{FFN-Only FT} & \textbf{LoRA} & \textbf{IA$^3$} & \textbf{Random} & \textbf{Target} \\
\midrule
\multicolumn{8}{c}{\textbf{Monolingual Capabilities}} \\
\midrule
FLORES et$\rightarrow$en & 25.18 & 9.88 & 20.61 & 25.09 & 26.35 & \cellcolor{lightgreen}26.72 & \cellcolor{mediumgreen}27.58 \\
FLORES en$\rightarrow$et & 2.71 & 2.90 & 4.28 & \cellcolor{lightgreen}11.15 & 3.47 & \cellcolor{mediumgreen}12.49 & 10.82 \\
BELEBELE et & \cellcolor{mediumgreen}66.67 & 15.00 & 42.00 & 56.22 & 64.00 & 58.11 & \cellcolor{lightgreen}65.00 \\
SIB200 et & 40.20 & 12.25 & 30.88 & 36.76 & \cellcolor{lightgreen}41.67 & 37.75 & \cellcolor{mediumgreen}55.88 \\
OPUS et$\rightarrow$en & 16.27 & 8.37 & 19.50 & 21.66 & 18.47 & \cellcolor{lightgreen}21.78 & \cellcolor{mediumgreen}23.32 \\
OPUS en$\rightarrow$et & 8.40 & 3.19 & 12.00 & 16.01 & 9.09 & \cellcolor{lightgreen}17.08 & \cellcolor{mediumgreen}17.19 \\
\end{tabular}
\end{sc}
\caption{Evaluation results comparing different fine-tuning configurations across various tasks for \textbf{Estonian} for \textsc{Llama-3.1-8B}.}
\label{tab:evaluation_results_et}
\end{table*}

\begin{table*}[h!]
\small
\centering
\begin{sc}
\begin{tabular}{@{}lccccccc@{}}
\toprule
Metric & \textbf{Base} & \textbf{Full FT} & \textbf{FFN-Only FT} & \textbf{LoRA} & \textbf{IA$^3$} & \textbf{Random} & \textbf{Target} \\
\midrule
\multicolumn{8}{c}{\textbf{Monolingual Capabilities}} \\
\midrule
FLORES en$\rightarrow$is & 9.43 & 7.34 & 10.09 & \cellcolor{lightgreen}12.47 & 9.02 & 11.57 & \cellcolor{mediumgreen}13.71 \\
FLORES is$\rightarrow$en & 22.63 & 15.36 & 23.00 & 23.16 & 22.55 & \cellcolor{lightgreen}24.46 & \cellcolor{mediumgreen}24.82 \\
BELEBELE is & \cellcolor{lightgreen}62.67 & 21.44 & 41.67 & 50.89 & 58.00 & 51.00 & \cellcolor{mediumgreen}66.11 \\
SIB200 is & 41.18 & 2.45 & 24.51 & 25.98 & \cellcolor{mediumgreen}47.55 & 28.43 & \cellcolor{lightgreen}42.16 \\
OPUS en$\rightarrow$is & 11.57 & 7.37 & 10.68 & \cellcolor{lightgreen}13.30 & 11.65 & 12.74 & \cellcolor{mediumgreen}13.45 \\
OPUS is$\rightarrow$en & 11.71 & 13.52 & 16.40 & \cellcolor{mediumgreen}17.55 & 13.43 & \cellcolor{lightgreen}17.14 & 16.23 \\
\end{tabular}
\end{sc}
\caption{Evaluation results comparing different fine-tuning configurations across various tasks for \textbf{Icelandic} for \textsc{Llama-3.1-8B}.}
\label{tab:evaluation_results_is}
\end{table*}

\begin{table*}[h!]
\small
\centering
\begin{sc}
\begin{tabular}{@{}lccccccc@{}}
\toprule
Metric & \textbf{Base} & \textbf{Full FT} & \textbf{FFN-Only FT} & \textbf{LoRA} & \textbf{IA$^3$} & \textbf{Random} & \textbf{Target} \\
\midrule
\multicolumn{8}{c}{\textbf{Monolingual Capabilities}} \\
\midrule
FLORES en$\rightarrow$ka & \cellcolor{mediumgreen}3.65 & 1.98 & 3.52 & 3.18 & \cellcolor{lightgreen}3.64 & 3.52 & 3.42 \\
FLORES ka$\rightarrow$en & 18.87 & 1.12 & 6.61 & 18.72 & \cellcolor{mediumgreen}19.20 & 18.02 & \cellcolor{lightgreen}19.01 \\
BELEBELE ka & \cellcolor{lightgreen}59.67 & 0.00 & 12.67 & \cellcolor{mediumgreen}60.33 & 57.22 & 49.56 & 58.89 \\
SIB200 ka & \cellcolor{mediumgreen}49.02 & 0.98 & 0.00 & \cellcolor{lightgreen}48.53 & \cellcolor{lightgreen}48.53 & 29.90 & 43.14 \\
OPUS en$\rightarrow$ka & 8.13 & 3.99 & 7.10 & 7.74 & 8.23 & \cellcolor{mediumgreen}8.51 & \cellcolor{lightgreen}8.26 \\
OPUS ka$\rightarrow$en & 5.63 & 3.24 & 3.95 & \cellcolor{lightgreen}5.76 & 5.67 & \cellcolor{mediumgreen}6.13 & 5.57 \\
\end{tabular}
\end{sc}
\caption{Evaluation results comparing different fine-tuning configurations across various tasks for \textbf{Georgian} for \textsc{Llama-3.1-8B}.}
\label{tab:evaluation_results_ka}
\end{table*}

\begin{table*}[h!]
\small
\centering
\begin{sc}
\begin{tabular}{@{}lccccccc@{}}
\toprule
Metric & \textbf{Base} & \textbf{Full FT} & \textbf{FFN-Only FT} & \textbf{LoRA} & \textbf{IA$^3$} & \textbf{Random} & \textbf{Target} \\
\midrule
\multicolumn{8}{c}{\textbf{Monolingual Capabilities}} \\
\midrule
FLORES en$\rightarrow$lt & 10.81 & 4.09 & 8.51 & 11.95 & 10.60 & \cellcolor{lightgreen}12.39 & \cellcolor{mediumgreen}12.96 \\
FLORES lt$\rightarrow$en & 22.67 & 10.75 & 20.50 & 23.28 & 23.33 & \cellcolor{lightgreen}24.56 & \cellcolor{mediumgreen}24.72 \\
BELEBELE lt & \cellcolor{lightgreen}66.44 & 34.33 & 46.33 & 53.00 & 65.22 & 64.56 & \cellcolor{mediumgreen}67.00 \\
SIB200 lt & 45.59 & 2.45 & 27.45 & 31.37 & \cellcolor{mediumgreen}47.55 & 27.94 & \cellcolor{lightgreen}46.08 \\
OPUS en$\rightarrow$lt & 12.80 & 6.25 & 11.24 & 15.36 & 12.94 & \cellcolor{mediumgreen}16.18 & \cellcolor{lightgreen}15.75 \\
OPUS lt$\rightarrow$en & 19.47 & 11.37 & 17.67 & 21.65 & 19.92 & \cellcolor{mediumgreen}23.08 & \cellcolor{lightgreen}22.51 \\
\end{tabular}
\end{sc}
\caption{Evaluation results comparing different fine-tuning configurations across various tasks for \textbf{Lithuanian} for \textsc{Llama-3.1-8B}.}
\label{tab:evaluation_results_lt}
\end{table*}

\begin{table*}[h!]
\small
\centering
\begin{sc}
\begin{tabular}{@{}lccccccc@{}}
\toprule
Metric & \textbf{Base} & \textbf{Full FT} & \textbf{FFN-Only FT} & \textbf{LoRA} & \textbf{IA$^3$} & \textbf{Random} & \textbf{Target} \\
\midrule
\multicolumn{8}{c}{\textbf{Monolingual Capabilities}} \\
\midrule
FLORES en$\rightarrow$lv & 11.26 & 10.03 & 14.52 & 14.51 & 11.35 & \cellcolor{lightgreen}14.98 & \cellcolor{mediumgreen}15.37 \\
FLORES lv$\rightarrow$en & 24.47 & 7.43 & 24.24 & 25.04 & 25.45 & \cellcolor{mediumgreen}27.36 & \cellcolor{lightgreen}26.80 \\
BELEBELE lv & \cellcolor{lightgreen}65.56 & 22.00 & 39.78 & 47.22 & 64.56 & 55.11 & \cellcolor{mediumgreen}67.67 \\
SIB200 lv & \cellcolor{lightgreen}42.16 & 20.10 & 2.45 & 16.67 & \cellcolor{mediumgreen}44.12 & 10.78 & 41.67 \\
OPUS en$\rightarrow$lv & 12.88 & 9.78 & 13.82 & 16.69 & 12.80 & \cellcolor{lightgreen}16.70 & \cellcolor{mediumgreen}17.08 \\
OPUS lv$\rightarrow$en & 21.38 & 8.71 & 19.98 & 24.49 & 23.17 & \cellcolor{mediumgreen}25.20 & \cellcolor{lightgreen}24.60 \\
\end{tabular}
\end{sc}
\caption{Evaluation results comparing different fine-tuning configurations across various tasks for \textbf{Latvian} for \textsc{Llama-3.1-8B}.}
\label{tab:evaluation_results_lv}
\end{table*}

\begin{table*}[h!]
\small
\centering
\begin{sc}
\begin{tabular}{@{}lccccccc@{}}
\toprule
Metric & \textbf{Base} & \textbf{Full FT} & \textbf{FFN-Only FT} & \textbf{LoRA} & \textbf{IA$^3$} & \textbf{Random} & \textbf{Target} \\
\midrule
\multicolumn{8}{c}{\textbf{Monolingual Capabilities}} \\
\midrule
FLORES en$\rightarrow$mk & 12.93 & 9.95 & 9.94 & 18.30 & 13.76 & \cellcolor{mediumgreen}21.62 & \cellcolor{lightgreen}21.56 \\
FLORES mk$\rightarrow$en & 33.37 & 6.36 & 27.85 & 33.89 & \cellcolor{lightgreen}34.19 & \cellcolor{mediumgreen}34.32 & 33.45 \\
BELEBELE mk & \cellcolor{mediumgreen}76.56 & 28.89 & 44.11 & 72.44 & \cellcolor{lightgreen}74.89 & 68.33 & 72.11 \\
SIB200 mk & 44.12 & 9.31 & 34.80 & 39.71 & \cellcolor{lightgreen}50.00 & \cellcolor{mediumgreen}53.43 & 40.20 \\
OPUS en$\rightarrow$mk & 10.72 & 11.89 & 12.21 & 15.56 & 11.46 & \cellcolor{mediumgreen}18.76 & \cellcolor{lightgreen}18.38 \\
OPUS mk$\rightarrow$en & 21.79 & 9.20 & 20.06 & 24.00 & 23.08 & \cellcolor{mediumgreen}24.86 & \cellcolor{lightgreen}24.16 \\
\end{tabular}
\end{sc}
\caption{Evaluation results comparing different fine-tuning configurations across various tasks for \textbf{Macedonian} for \textsc{Llama-3.1-8B}.}
\label{tab:evaluation_results_mk}
\end{table*}

\begin{table*}[h!]
\small
\centering
\begin{sc}
\begin{tabular}{@{}lccccccc@{}}
\toprule
Metric & \textbf{Base} & \textbf{Full FT} & \textbf{FFN-Only FT} & \textbf{LoRA} & \textbf{IA$^3$} & \textbf{Random} & \textbf{Target} \\
\midrule
\multicolumn{8}{c}{\textbf{Monolingual Capabilities}} \\
\midrule
FLORES en$\rightarrow$mt & 9.19 & 16.81 & \cellcolor{mediumgreen}24.25 & \cellcolor{lightgreen}20.96 & 10.63 & 17.71 & 20.92 \\
FLORES mt$\rightarrow$en & 34.37 & 7.31 & 38.95 & 20.39 & 36.66 & \cellcolor{lightgreen}42.74 & \cellcolor{mediumgreen}44.39 \\
BELEBELE mt & 57.67 & 25.11 & 40.00 & 50.89 & 53.67 & \cellcolor{lightgreen}58.11 & \cellcolor{mediumgreen}66.44 \\
SIB200 mt & \cellcolor{lightgreen}37.75 & 10.29 & 24.02 & 27.94 & 33.82 & 33.82 & \cellcolor{mediumgreen}38.73 \\
OPUS en$\rightarrow$mt & 13.05 & 16.02 & 23.29 & \cellcolor{mediumgreen}24.28 & 14.76 & 20.41 & \cellcolor{lightgreen}24.23 \\
OPUS mt$\rightarrow$en & 29.27 & 10.67 & 30.24 & 19.60 & 32.97 & \cellcolor{lightgreen}36.79 & \cellcolor{mediumgreen}39.36 \\
\end{tabular}
\end{sc}
\caption{Evaluation results comparing different fine-tuning configurations across various tasks for \textbf{Maltese} for \textsc{Llama-3.1-8B}.}
\label{tab:evaluation_results_mt}
\end{table*}

\begin{table*}[h!]
\small
\centering
\begin{sc}
\begin{tabular}{@{}lccccccc@{}}
\toprule
Metric & \textbf{Base} & \textbf{Full FT} & \textbf{FFN-Only FT} & \textbf{LoRA} & \textbf{IA$^3$} & \textbf{Random} & \textbf{Target} \\
\midrule
\multicolumn{8}{c}{\textbf{Monolingual Capabilities}} \\
\midrule
FLORES en$\rightarrow$ne & 4.27 & 2.14 & 3.89 & \cellcolor{lightgreen}6.56 & 5.07 & 6.46 & \cellcolor{mediumgreen}6.81 \\
FLORES ne$\rightarrow$en & 22.78 & 2.42 & 7.73 & 22.65 & 22.81 & \cellcolor{lightgreen}23.62 & \cellcolor{mediumgreen}24.03 \\
BELEBELE ne & \cellcolor{mediumgreen}59.67 & 1.22 & 29.44 & 53.89 & 54.67 & 55.78 & \cellcolor{lightgreen}57.44 \\
SIB200 ne & \cellcolor{lightgreen}36.27 & 4.41 & 4.41 & 27.45 & 29.41 & 33.33 & \cellcolor{mediumgreen}57.84 \\
OPUS en$\rightarrow$ne & 12.36 & 5.75 & 10.96 & \cellcolor{lightgreen}17.61 & 16.42 & 17.50 & \cellcolor{mediumgreen}18.90 \\
OPUS ne$\rightarrow$en & 25.97 & 12.53 & 21.17 & \cellcolor{mediumgreen}31.08 & 28.49 & 28.00 & \cellcolor{lightgreen}30.87 \\
\end{tabular}
\end{sc}
\caption{Evaluation results comparing different fine-tuning configurations across various tasks for \textbf{Nepali} for \textsc{Llama-3.1-8B}.}
\label{tab:evaluation_results_ne}
\end{table*}

\begin{table*}[h!]
\small
\centering
\begin{sc}
\begin{tabular}{@{}lccccccc@{}}
\toprule
Metric & \textbf{Base} & \textbf{Full FT} & \textbf{FFN-Only FT} & \textbf{LoRA} & \textbf{IA$^3$} & \textbf{Random} & \textbf{Target} \\
\midrule
\multicolumn{8}{c}{\textbf{Monolingual Capabilities}} \\
\midrule
FLORES en$\rightarrow$sk & 16.84 & 7.97 & 16.73 & 17.94 & 17.43 & \cellcolor{lightgreen}18.28 & \cellcolor{mediumgreen}19.14 \\
FLORES sk$\rightarrow$en & 29.83 & 18.28 & 24.74 & 29.98 & 30.58 & \cellcolor{mediumgreen}31.30 & \cellcolor{lightgreen}30.75 \\
BELEBELE sk & \cellcolor{mediumgreen}73.67 & 27.44 & 39.78 & 61.11 & \cellcolor{lightgreen}73.00 & 65.56 & 72.89 \\
SIB200 sk & 40.20 & 20.59 & 39.22 & \cellcolor{lightgreen}44.61 & 39.22 & 33.82 & \cellcolor{mediumgreen}46.08 \\
OPUS en$\rightarrow$sk & 17.21 & 9.33 & 16.08 & 17.34 & 17.32 & \cellcolor{mediumgreen}18.55 & \cellcolor{lightgreen}18.24 \\
OPUS sk$\rightarrow$en & 23.13 & 17.56 & 22.42 & 25.59 & 24.14 & \cellcolor{mediumgreen}27.12 & \cellcolor{lightgreen}26.27 \\
\end{tabular}
\end{sc}
\caption{Evaluation results comparing different fine-tuning configurations across various tasks for \textbf{Slovak} for \textsc{Llama-3.1-8B}.}
\label{tab:evaluation_results_sk}
\end{table*}

\begin{table*}[h!]
\small
\centering
\begin{sc}
\begin{tabular}{@{}lccccccc@{}}
\toprule
Metric & \textbf{Base} & \textbf{Full FT} & \textbf{FFN-Only FT} & \textbf{LoRA} & \textbf{IA$^3$} & \textbf{Random} & \textbf{Target} \\
\midrule
\multicolumn{8}{c}{\textbf{Monolingual Capabilities}} \\
\midrule
FLORES en$\rightarrow$sl & 15.38 & 3.45 & 7.63 & 15.74 & 15.76 & \cellcolor{mediumgreen}16.78 & \cellcolor{lightgreen}16.51 \\
FLORES sl$\rightarrow$en & 25.93 & 8.64 & 20.47 & 25.87 & \cellcolor{lightgreen}26.66 & 26.55 & \cellcolor{mediumgreen}27.50 \\
BELEBELE sl & \cellcolor{mediumgreen}71.00 & 26.78 & 38.44 & 51.67 & \cellcolor{lightgreen}69.11 & 63.00 & 68.56 \\
SIB200 sl & 41.67 & 23.04 & 16.18 & 20.59 & 44.12 & \cellcolor{mediumgreen}53.92 & \cellcolor{lightgreen}47.55 \\
OPUS en$\rightarrow$sl & 14.46 & 2.74 & 10.53 & 14.26 & 14.48 & \cellcolor{lightgreen}16.00 & \cellcolor{mediumgreen}16.38 \\
OPUS sl$\rightarrow$en & 16.82 & 3.26 & 16.63 & 19.86 & 18.35 & \cellcolor{lightgreen}20.24 & \cellcolor{mediumgreen}20.89 \\
\end{tabular}
\end{sc}
\caption{Evaluation results comparing different fine-tuning configurations across various tasks for \textbf{Slovenian} for \textsc{Llama-3.1-8B}.}
\label{tab:evaluation_results_sl}
\end{table*}

\twocolumn
\begin{table}[h!]
\subsection{Mistral-Nemo-12B}
\vspace{10pt}
\small
\centering
\begin{sc}
\begin{tabular}{@{}lccc@{}}
\toprule
Metric & \textbf{Base} & \textbf{Random} & \textbf{Target} \\
\midrule
\multicolumn{4}{c}{\textbf{Monolingual Capabilities}} \\
\midrule
FLORES af$\rightarrow$en & 42.32 & \cellcolor{lightgreen}44.12 & \cellcolor{mediumgreen}46.05 \\
FLORES en$\rightarrow$af & 20.34 & \cellcolor{mediumgreen}27.51 & \cellcolor{lightgreen}25.52 \\
BELEBELE af & \cellcolor{mediumgreen}79.67 & 75.56 & \cellcolor{lightgreen}79.44 \\
SIB200 af & \cellcolor{lightgreen}31.86 & \cellcolor{mediumgreen}43.14 & \cellcolor{lightgreen}31.86 \\
OPUS af$\rightarrow$en & 19.69 & \cellcolor{lightgreen}30.04 & \cellcolor{mediumgreen}30.54 \\
OPUS en$\rightarrow$af & 16.77 & \cellcolor{mediumgreen}19.24 & \cellcolor{lightgreen}17.99 \\
\hdashline
Avg. & 35.11 & \cellcolor{mediumgreen}39.93 & \cellcolor{lightgreen}38.57 \\
\midrule
\multicolumn{4}{c}{\textbf{General Performance}} \\
\midrule
arc challenge & \cellcolor{mediumgreen}80.27 & 72.24 & \cellcolor{lightgreen}75.59 \\
arc easy & \cellcolor{mediumgreen}89.82 & 83.33 & \cellcolor{lightgreen}87.72 \\
hellaswag & \cellcolor{lightgreen}40.86 & 38.31 & \cellcolor{mediumgreen}42.32 \\
piqa & \cellcolor{mediumgreen}80.03 & 77.15 & \cellcolor{lightgreen}77.20 \\
winogrande & \cellcolor{mediumgreen}54.22 & \cellcolor{lightgreen}53.28 & 52.17 \\
mmlu & \cellcolor{mediumgreen}61.66 & 55.85 & \cellcolor{lightgreen}61.33 \\
\hdashline
Avg. & \cellcolor{mediumgreen}67.81 & 63.36 & \cellcolor{lightgreen}66.06 \\
\bottomrule
\addlinespace[10pt]
\end{tabular}
\end{sc}
\caption{Evaluation results comparing different fine-tuning configurations across various tasks for \textbf{Afrikaans} for \textsc{Mistral-Nemo-12B}.}
\label{tab:evaluation_results_af_nemo}
\end{table}

\begin{table}[h!]
\small
\centering
\begin{sc}
\begin{tabular}{@{}lccc@{}}
\toprule
Metric & \textbf{Base} & \textbf{Random} & \textbf{Target} \\
\midrule
\multicolumn{4}{c}{\textbf{Monolingual Capabilities}} \\
\midrule
FLORES cy$\rightarrow$en & 36.21 & \cellcolor{mediumgreen}43.11 & \cellcolor{lightgreen}38.94 \\
FLORES en$\rightarrow$cy & 9.09 & \cellcolor{lightgreen}21.73 & \cellcolor{mediumgreen}25.54 \\
SIB200 cy & \cellcolor{lightgreen}35.78 & 31.37 & \cellcolor{mediumgreen}40.20 \\
OPUS cy$\rightarrow$en & 15.33 & \cellcolor{mediumgreen}35.25 & \cellcolor{lightgreen}30.37 \\
OPUS en$\rightarrow$cy & 10.64 & \cellcolor{lightgreen}18.99 & \cellcolor{mediumgreen}23.81 \\
\hdashline
Avg. & 21.41 & \cellcolor{lightgreen}30.09 & \cellcolor{mediumgreen}31.77 \\
\midrule
\multicolumn{4}{c}{\textbf{General Performance}} \\
\midrule
arc challenge & \cellcolor{mediumgreen}80.27 & 63.88 & \cellcolor{lightgreen}71.57 \\
arc easy & \cellcolor{mediumgreen}89.82 & 78.77 & \cellcolor{lightgreen}85.09 \\
hellaswag & \cellcolor{mediumgreen}40.86 & 30.91 & \cellcolor{lightgreen}35.44 \\
piqa & \cellcolor{mediumgreen}80.03 & 61.15 & \cellcolor{lightgreen}65.61 \\
winogrande & \cellcolor{mediumgreen}54.22 & 51.07 & \cellcolor{lightgreen}51.93 \\
mmlu & \cellcolor{mediumgreen}61.66 & 50.10 & \cellcolor{lightgreen}55.58 \\
\hdashline
Avg. & \cellcolor{mediumgreen}67.81 & 55.98 & \cellcolor{lightgreen}60.87 \\
\bottomrule
\addlinespace[10pt]
\end{tabular}
\end{sc}
\caption{Evaluation results comparing different fine-tuning configurations across various tasks for \textbf{Welsh} for \textsc{Mistral-Nemo-12B}.}
\label{tab:evaluation_results_cy_nemo}
\end{table}

\begin{table}[h!]
\small
\centering
\begin{sc}
\begin{tabular}{@{}lccc@{}}
\toprule
Metric & \textbf{Base} & \textbf{Random} & \textbf{Target} \\
\midrule
\multicolumn{4}{c}{\textbf{Monolingual Capabilities}} \\
\midrule
FLORES et$\rightarrow$en & 23.86 & \cellcolor{mediumgreen}25.62 & \cellcolor{lightgreen}25.54 \\
FLORES en$\rightarrow$et & 2.35 & \cellcolor{mediumgreen}4.90 & \cellcolor{lightgreen}4.33 \\
BELEBELE et & \cellcolor{mediumgreen}73.00 & 62.44 & \cellcolor{lightgreen}71.00 \\
SIB200 et & \cellcolor{lightgreen}32.35 & 29.90 & \cellcolor{mediumgreen}36.76 \\
OPUS et$\rightarrow$en & 13.14 & \cellcolor{mediumgreen}22.21 & \cellcolor{lightgreen}21.01 \\
OPUS en$\rightarrow$et & 6.37 & \cellcolor{mediumgreen}14.12 & \cellcolor{lightgreen}11.60 \\
\hdashline
Avg. & 25.18 & \cellcolor{lightgreen}26.53 & \cellcolor{mediumgreen}28.37 \\
\midrule
\multicolumn{4}{c}{\textbf{General Performance}} \\
\midrule
arc challenge & \cellcolor{mediumgreen}80.27 & 71.91 & \cellcolor{lightgreen}74.92 \\
arc easy & \cellcolor{mediumgreen}89.82 & 83.68 & \cellcolor{lightgreen}88.42 \\
hellaswag & \cellcolor{mediumgreen}40.86 & 32.99 & \cellcolor{lightgreen}40.67 \\
piqa & \cellcolor{mediumgreen}80.03 & 68.61 & \cellcolor{lightgreen}77.75 \\
winogrande & \cellcolor{mediumgreen}54.22 & 52.09 & \cellcolor{lightgreen}53.35 \\
mmlu & \cellcolor{mediumgreen}61.66 & 54.93 & \cellcolor{lightgreen}59.83 \\
\hdashline
Avg. & \cellcolor{mediumgreen}67.81 & 60.70 & \cellcolor{lightgreen}65.82 \\
\bottomrule
\addlinespace[10pt]
\end{tabular}
\end{sc}
\caption{Evaluation results comparing different fine-tuning configurations across various tasks for \textbf{Estonian} for \textsc{Mistral-Nemo-12B}.}
\label{tab:evaluation_results_et_nemo}
\end{table}

\begin{table}[h!]
\small
\centering
\begin{sc}
\begin{tabular}{@{}lccc@{}}
\toprule
Metric & \textbf{Base} & \textbf{Random} & \textbf{Target} \\
\midrule
\multicolumn{4}{c}{\textbf{Monolingual Capabilities}} \\
\midrule
FLORES en$\rightarrow$is & 6.05 & \cellcolor{lightgreen}6.37 & \cellcolor{mediumgreen}11.14 \\
FLORES is$\rightarrow$en & \cellcolor{lightgreen}22.24 & 5.40 & \cellcolor{mediumgreen}25.24 \\
BELEBELE is & \cellcolor{lightgreen}67.00 & 60.00 & \cellcolor{mediumgreen}70.67 \\
SIB200 is & \cellcolor{lightgreen}34.80 & 25.98 & \cellcolor{mediumgreen}38.24 \\
OPUS en$\rightarrow$is & 7.66 & \cellcolor{lightgreen}7.96 & \cellcolor{mediumgreen}13.06 \\
OPUS is$\rightarrow$en & \cellcolor{lightgreen}10.75 & 9.80 & \cellcolor{mediumgreen}16.50 \\
\hdashline
Avg. & \cellcolor{lightgreen}24.75 & 19.25 & \cellcolor{mediumgreen}29.14 \\
\midrule
\multicolumn{4}{c}{\textbf{General Performance}} \\
\midrule
arc challenge & \cellcolor{mediumgreen}80.27 & 64.55 & \cellcolor{lightgreen}76.59 \\
arc easy & \cellcolor{mediumgreen}89.82 & 83.16 & \cellcolor{lightgreen}88.95 \\
hellaswag & \cellcolor{lightgreen}40.86 & 31.59 & \cellcolor{mediumgreen}41.73 \\
piqa & \cellcolor{mediumgreen}80.03 & 60.72 & \cellcolor{lightgreen}74.54 \\
winogrande & \cellcolor{mediumgreen}54.22 & 51.70 & \cellcolor{lightgreen}53.67 \\
mmlu & \cellcolor{mediumgreen}61.66 & 51.99 & \cellcolor{lightgreen}60.94 \\
\hdashline
Avg. & \cellcolor{mediumgreen}67.81 & 57.28 & \cellcolor{lightgreen}66.07 \\
\bottomrule
\addlinespace[10pt]
\end{tabular}
\end{sc}
\caption{Evaluation results comparing different fine-tuning configurations across various tasks for \textbf{Icelandic} for \textsc{Mistral-Nemo-12B}.}
\label{tab:evaluation_results_is_nemo}
\end{table}

\begin{table}[h!]
\small
\centering
\begin{sc}
\begin{tabular}{@{}lccc@{}}
\toprule
Metric & \textbf{Base} & \textbf{Random} & \textbf{Target} \\
\midrule
\multicolumn{4}{c}{\textbf{Monolingual Capabilities}} \\
\midrule
FLORES en$\rightarrow$ka & 2.40 & \cellcolor{mediumgreen}7.97 & \cellcolor{lightgreen}3.83 \\
FLORES ka$\rightarrow$en & 19.32 & \cellcolor{mediumgreen}20.11 & \cellcolor{lightgreen}19.58 \\
BELEBELE ka & \cellcolor{lightgreen}71.89 & 68.56 & \cellcolor{mediumgreen}72.33 \\
SIB200 ka & 30.39 & \cellcolor{mediumgreen}37.25 & \cellcolor{lightgreen}32.35 \\
OPUS en$\rightarrow$ka & 4.68 & \cellcolor{mediumgreen}7.45 & \cellcolor{lightgreen}5.93 \\
OPUS ka$\rightarrow$en & 3.24 & \cellcolor{mediumgreen}5.65 & \cellcolor{lightgreen}3.97 \\
\hdashline
Avg. & 21.99 & \cellcolor{mediumgreen}24.50 & \cellcolor{lightgreen}23.00 \\
\midrule
\multicolumn{4}{c}{\textbf{General Performance}} \\
\midrule
arc challenge & \cellcolor{mediumgreen}80.27 & 69.23 & \cellcolor{lightgreen}77.26 \\
arc easy & \cellcolor{mediumgreen}89.82 & 82.81 & \cellcolor{lightgreen}87.37 \\
hellaswag & \cellcolor{lightgreen}40.86 & \cellcolor{mediumgreen}41.17 & 37.87 \\
piqa & \cellcolor{mediumgreen}80.03 & 70.08 & \cellcolor{lightgreen}72.47 \\
winogrande & \cellcolor{mediumgreen}54.22 & 52.33 & \cellcolor{lightgreen}53.99 \\
mmlu & \cellcolor{mediumgreen}61.66 & 56.63 & \cellcolor{lightgreen}60.48 \\
\hdashline
Avg. & \cellcolor{mediumgreen}67.81 & 62.04 & \cellcolor{lightgreen}64.91 \\
\bottomrule
\addlinespace[10pt]
\end{tabular}
\end{sc}
\caption{Evaluation results comparing different fine-tuning configurations across various tasks for \textbf{Georgian} for \textsc{Mistral-Nemo-12B}.}
\label{tab:evaluation_results_ka_nemo}
\end{table}

\begin{table}[h!]
\small
\centering
\begin{sc}
\begin{tabular}{@{}lccc@{}}
\toprule
Metric & \textbf{Base} & \textbf{Random} & \textbf{Target} \\
\midrule
\multicolumn{4}{c}{\textbf{Monolingual Capabilities}} \\
\midrule
FLORES en$\rightarrow$lt & 5.95 & \cellcolor{lightgreen}7.52 & \cellcolor{mediumgreen}9.46 \\
FLORES lt$\rightarrow$en & 22.84 & \cellcolor{lightgreen}24.02 & \cellcolor{mediumgreen}24.18 \\
BELEBELE lt & \cellcolor{mediumgreen}73.44 & 64.11 & \cellcolor{lightgreen}71.56 \\
SIB200 lt & \cellcolor{lightgreen}33.82 & 25.98 & \cellcolor{mediumgreen}34.80 \\
OPUS en$\rightarrow$lt & 6.97 & \cellcolor{lightgreen}10.58 & \cellcolor{mediumgreen}11.70 \\
OPUS lt$\rightarrow$en & 16.29 & \cellcolor{mediumgreen}20.54 & \cellcolor{lightgreen}19.94 \\
\hdashline
Avg. & \cellcolor{lightgreen}26.55 & 25.46 & \cellcolor{mediumgreen}28.61 \\
\midrule
\multicolumn{4}{c}{\textbf{General Performance}} \\
\midrule
arc challenge & \cellcolor{mediumgreen}80.27 & 68.56 & \cellcolor{lightgreen}76.59 \\
arc easy & \cellcolor{mediumgreen}89.82 & 86.32 & \cellcolor{lightgreen}88.77 \\
hellaswag & \cellcolor{lightgreen}40.85 & 36.47 & \cellcolor{mediumgreen}41.65 \\
piqa & \cellcolor{mediumgreen}80.03 & 67.25 & \cellcolor{lightgreen}72.20 \\
winogrande & \cellcolor{mediumgreen}54.22 & 51.14 & \cellcolor{lightgreen}53.83 \\
mmlu & \cellcolor{mediumgreen}61.66 & 52.25 & \cellcolor{lightgreen}60.16 \\
\hdashline
Avg. & \cellcolor{mediumgreen}67.81 & 60.33 & \cellcolor{lightgreen}65.53 \\
\bottomrule
\addlinespace[10pt]
\end{tabular}
\end{sc}
\caption{Evaluation results comparing different fine-tuning configurations across various tasks for \textbf{Lithuanian} for \textsc{Mistral-Nemo-12B}.}
\label{tab:evaluation_results_lt_nemo}
\end{table}

\begin{table}[h!]
\small
\centering
\begin{sc}
\begin{tabular}{@{}lccc@{}}
\toprule
Metric & \textbf{Base} & \textbf{Random} & \textbf{Target} \\
\midrule
\multicolumn{4}{c}{\textbf{Monolingual Capabilities}} \\
\midrule
FLORES en$\rightarrow$lv & 6.01 & \cellcolor{mediumgreen}14.63 & \cellcolor{lightgreen}7.62 \\
FLORES lv$\rightarrow$en & 25.29 & \cellcolor{mediumgreen}26.96 & \cellcolor{lightgreen}26.65 \\
BELEBELE lv & \cellcolor{mediumgreen}72.89 & 60.44 & \cellcolor{lightgreen}66.00 \\
SIB200 lv & \cellcolor{lightgreen}34.80 & \cellcolor{mediumgreen}46.08 & 30.88 \\
OPUS en$\rightarrow$lv & 6.86 & \cellcolor{mediumgreen}15.63 & \cellcolor{lightgreen}10.67 \\
OPUS lv$\rightarrow$en & 19.44 & \cellcolor{mediumgreen}23.49 & \cellcolor{lightgreen}22.49 \\
\hdashline
Avg. & \cellcolor{lightgreen}27.55 & \cellcolor{mediumgreen}31.21 & 27.39 \\
\midrule
\multicolumn{4}{c}{\textbf{General Performance}} \\
\midrule
arc challenge & \cellcolor{mediumgreen}80.27 & 64.21 & \cellcolor{lightgreen}72.91 \\
arc easy & \cellcolor{mediumgreen}89.82 & 83.68 & \cellcolor{lightgreen}87.54 \\
hellaswag & \cellcolor{mediumgreen}40.86 & 34.50 & \cellcolor{lightgreen}35.25 \\
piqa & \cellcolor{mediumgreen}80.03 & \cellcolor{lightgreen}66.32 & 62.84 \\
winogrande & \cellcolor{mediumgreen}54.22 & 51.22 & \cellcolor{lightgreen}51.93 \\
mmlu & \cellcolor{mediumgreen}61.66 & 53.89 & \cellcolor{lightgreen}58.79 \\
\hdashline
Avg. & \cellcolor{mediumgreen}67.81 & 58.97 & \cellcolor{lightgreen}61.54 \\
\bottomrule
\addlinespace[10pt]
\end{tabular}
\end{sc}
\caption{Evaluation results comparing different fine-tuning configurations across various tasks for \textbf{Latvian} for \textsc{Mistral-Nemo-12B}.}
\label{tab:evaluation_results_lv_nemo}
\end{table}

\begin{table}[h!]
\small
\centering
\begin{sc}
\begin{tabular}{@{}lccc@{}}
\toprule
Metric & \textbf{Base} & \textbf{Random} & \textbf{Target} \\
\midrule
\multicolumn{4}{c}{\textbf{Monolingual Capabilities}} \\
\midrule
FLORES en$\rightarrow$mk & \cellcolor{lightgreen}8.85 & \cellcolor{mediumgreen}16.78 & 8.43 \\
FLORES mk$\rightarrow$en & 32.65 & \cellcolor{mediumgreen}34.40 & \cellcolor{lightgreen}32.79 \\
BELEBELE mk & \cellcolor{lightgreen}76.11 & 73.56 & \cellcolor{mediumgreen}76.67 \\
SIB200 mk & 31.37 & \cellcolor{lightgreen}31.86 & \cellcolor{mediumgreen}44.12 \\
OPUS en$\rightarrow$mk & 8.98 & \cellcolor{mediumgreen}14.63 & \cellcolor{lightgreen}10.12 \\
OPUS mk$\rightarrow$en & 20.96 & \cellcolor{mediumgreen}24.28 & \cellcolor{lightgreen}22.71 \\
\hdashline
Avg. & 29.82 & \cellcolor{mediumgreen}32.59 & \cellcolor{lightgreen}32.47 \\
\midrule
\multicolumn{4}{c}{\textbf{General Performance}} \\
\midrule
arc challenge & \cellcolor{mediumgreen}80.27 & 73.58 & \cellcolor{lightgreen}75.92 \\
arc easy & \cellcolor{mediumgreen}89.82 & 87.02 & \cellcolor{lightgreen}87.89 \\
hellaswag & \cellcolor{lightgreen}40.86 & 37.81 & \cellcolor{mediumgreen}44.30 \\
piqa & \cellcolor{mediumgreen}80.03 & 74.97 & \cellcolor{lightgreen}76.17 \\
winogrande & \cellcolor{mediumgreen}54.22 & 51.93 & \cellcolor{lightgreen}54.06 \\
mmlu & \cellcolor{mediumgreen}61.66 & 57.28 & \cellcolor{lightgreen}59.57 \\
\hdashline
Avg. & \cellcolor{mediumgreen}67.81 & 63.77 & \cellcolor{lightgreen}66.32 \\
\bottomrule
\addlinespace[10pt]
\end{tabular}
\end{sc}
\caption{Evaluation results comparing different fine-tuning configurations across various tasks for \textbf{Macedonian} for \textsc{Mistral-Nemo-12B}.}
\label{tab:evaluation_results_mk_nemo}
\end{table}

\begin{table}[h!]
\small
\centering
\begin{sc}
\begin{tabular}{@{}lccc@{}}
\toprule
Metric & \textbf{Base} & \textbf{Random} & \textbf{Target} \\
\midrule
\multicolumn{4}{c}{\textbf{Monolingual Capabilities}} \\
\midrule
FLORES en$\rightarrow$mt & 3.20 & \cellcolor{lightgreen}19.27 & \cellcolor{mediumgreen}19.41 \\
FLORES mt$\rightarrow$en & 16.24 & \cellcolor{mediumgreen}40.39 & \cellcolor{lightgreen}39.70 \\
BELEBELE mt & 41.33 & \cellcolor{lightgreen}64.67 & \cellcolor{mediumgreen}72.44 \\
SIB200 mt & 30.88 & \cellcolor{mediumgreen}44.61 & \cellcolor{lightgreen}33.33 \\
OPUS en$\rightarrow$mt & 5.19 & \cellcolor{lightgreen}20.71 & \cellcolor{mediumgreen}20.91 \\
OPUS mt$\rightarrow$en & 16.52 & \cellcolor{lightgreen}31.51 & \cellcolor{mediumgreen}33.82 \\
\hdashline
Avg. & 18.89 & \cellcolor{mediumgreen}36.86 & \cellcolor{lightgreen}36.60 \\
\midrule
\multicolumn{4}{c}{\textbf{General Performance}} \\
\midrule
arc challenge & \cellcolor{mediumgreen}80.27 & 68.23 & \cellcolor{lightgreen}74.92 \\
arc easy & \cellcolor{mediumgreen}89.82 & 80.88 & \cellcolor{lightgreen}86.67 \\
hellaswag & \cellcolor{mediumgreen}40.86 & 35.46 & \cellcolor{lightgreen}39.45 \\
piqa & \cellcolor{mediumgreen}80.03 & \cellcolor{lightgreen}72.74 & 65.23 \\
winogrande & \cellcolor{mediumgreen}54.22 & 51.54 & \cellcolor{lightgreen}52.57 \\
mmlu & \cellcolor{mediumgreen}61.66 & 52.45 & \cellcolor{lightgreen}59.24 \\
\hdashline
Avg. & \cellcolor{mediumgreen}67.81 & 60.22 & \cellcolor{lightgreen}63.01 \\
\bottomrule
\addlinespace[10pt]
\end{tabular}
\end{sc}
\caption{Evaluation results comparing different fine-tuning configurations across various tasks for \textbf{Maltese} for \textsc{Mistral-Nemo-12B}.}
\label{tab:evaluation_results_mt_nemo}
\end{table}

\begin{table}[h!]
\small
\centering
\begin{sc}
\begin{tabular}{@{}lccc@{}}
\toprule
Metric & \textbf{Base} & \textbf{Random} & \textbf{Target} \\
\midrule
\multicolumn{4}{c}{\textbf{Monolingual Capabilities}} \\
\midrule
FLORES en$\rightarrow$ne & 1.58 & \cellcolor{mediumgreen}3.21 & \cellcolor{lightgreen}2.90 \\
FLORES ne$\rightarrow$en & \cellcolor{lightgreen}18.36 & 18.36 & \cellcolor{mediumgreen}19.39 \\
BELEBELE ne & \cellcolor{mediumgreen}64.11 & 51.56 & \cellcolor{lightgreen}60.67 \\
SIB200 ne & \cellcolor{mediumgreen}34.31 & 12.25 & \cellcolor{lightgreen}22.55 \\
OPUS en$\rightarrow$ne & 11.35 & \cellcolor{mediumgreen}16.22 & \cellcolor{lightgreen}14.82 \\
OPUS ne$\rightarrow$en & \cellcolor{lightgreen}23.64 & \cellcolor{mediumgreen}26.49 & 21.99 \\
\hdashline
Avg. & \cellcolor{mediumgreen}25.56 & 21.35 & \cellcolor{lightgreen}23.72 \\
\midrule
\multicolumn{4}{c}{\textbf{General Performance}} \\
\midrule
arc challenge & \cellcolor{mediumgreen}80.27 & 51.51 & \cellcolor{lightgreen}72.58 \\
arc easy & \cellcolor{mediumgreen}89.82 & 66.84 & \cellcolor{lightgreen}87.19 \\
hellaswag & \cellcolor{mediumgreen}40.86 & 29.48 & \cellcolor{lightgreen}35.37 \\
piqa & \cellcolor{mediumgreen}80.03 & 63.38 & \cellcolor{lightgreen}76.01 \\
winogrande & \cellcolor{mediumgreen}54.22 & 52.49 & \cellcolor{lightgreen}53.12 \\
mmlu & \cellcolor{mediumgreen}61.66 & 46.18 & \cellcolor{lightgreen}55.45 \\
\hdashline
Avg. & \cellcolor{mediumgreen}67.81 & 51.65 & \cellcolor{lightgreen}63.29 \\
\bottomrule
\addlinespace[10pt]
\end{tabular}
\end{sc}
\caption{Evaluation results comparing different fine-tuning configurations across various tasks for \textbf{Nepali} for \textsc{Mistral-Nemo-12B}.}
\label{tab:evaluation_results_ne_nemo}
\end{table}

\begin{table}[h!]
\small
\centering
\begin{sc}
\begin{tabular}{@{}lccc@{}}
\toprule
Metric & \textbf{Base} & \textbf{Random} & \textbf{Target} \\
\midrule
\multicolumn{4}{c}{\textbf{Monolingual Capabilities}} \\
\midrule
FLORES en$\rightarrow$sk & 7.76 & \cellcolor{mediumgreen}17.36 & \cellcolor{lightgreen}15.05 \\
FLORES sk$\rightarrow$en & \cellcolor{lightgreen}28.93 & 27.71 & \cellcolor{mediumgreen}29.25 \\
BELEBELE sk & \cellcolor{mediumgreen}79.78 & 71.89 & \cellcolor{lightgreen}77.33 \\
SIB200 sk & \cellcolor{lightgreen}34.80 & \cellcolor{mediumgreen}44.12 & 28.43 \\
OPUS en$\rightarrow$sk & 8.73 & \cellcolor{mediumgreen}17.68 & \cellcolor{lightgreen}15.89 \\
OPUS sk$\rightarrow$en & 20.26 & \cellcolor{lightgreen}25.70 & \cellcolor{mediumgreen}26.11 \\
\hdashline
Avg. & 30.04 & \cellcolor{mediumgreen}34.08 & \cellcolor{lightgreen}32.01 \\
\midrule
\multicolumn{4}{c}{\textbf{General Performance}} \\
\midrule
arc challenge & \cellcolor{mediumgreen}80.27 & 70.90 & \cellcolor{lightgreen}75.59 \\
arc easy & \cellcolor{mediumgreen}89.82 & 85.09 & \cellcolor{lightgreen}88.77 \\
hellaswag & \cellcolor{mediumgreen}40.86 & 36.49 & \cellcolor{lightgreen}40.12 \\
piqa & \cellcolor{mediumgreen}80.03 & 70.95 & \cellcolor{lightgreen}78.62 \\
winogrande & \cellcolor{lightgreen}54.22 & 51.62 & \cellcolor{mediumgreen}55.49 \\
mmlu & \cellcolor{mediumgreen}61.66 & 53.36 & \cellcolor{lightgreen}60.42 \\
\hdashline
Avg. & \cellcolor{mediumgreen}67.81 & 61.40 & \cellcolor{lightgreen}66.50 \\
\bottomrule
\addlinespace[10pt]
\end{tabular}
\end{sc}
\caption{Evaluation results comparing different fine-tuning configurations across various tasks for \textbf{Slovak} for \textsc{Mistral-Nemo-12B}.}
\label{tab:evaluation_results_sk_nemo}
\end{table}

\begin{table}[h!]
\small
\centering
\begin{sc}
\begin{tabular}{@{}lccc@{}}
\toprule
Metric & \textbf{Base} & \textbf{Random} & \textbf{Target} \\
\midrule
\multicolumn{4}{c}{\textbf{Monolingual Capabilities}} \\
\midrule
FLORES en$\rightarrow$sl & 6.47 & \cellcolor{lightgreen}10.02 & \cellcolor{mediumgreen}13.19 \\
FLORES sl$\rightarrow$en & \cellcolor{mediumgreen}24.99 & \cellcolor{lightgreen}20.81 & 17.12 \\
BELEBELE sl & \cellcolor{mediumgreen}77.56 & 68.33 & \cellcolor{lightgreen}71.89 \\
SIB200 sl & \cellcolor{mediumgreen}32.84 & \cellcolor{lightgreen}32.35 & 26.96 \\
OPUS en$\rightarrow$sl & 7.18 & \cellcolor{mediumgreen}10.74 & \cellcolor{lightgreen}8.69 \\
OPUS sl$\rightarrow$en & \cellcolor{lightgreen}15.11 & \cellcolor{mediumgreen}17.59 & 11.76 \\
\hdashline
Avg. & \cellcolor{mediumgreen}27.36 & \cellcolor{lightgreen}26.64 & 24.93 \\
\midrule
\multicolumn{4}{c}{\textbf{General Performance}} \\
\midrule
arc challenge & \cellcolor{mediumgreen}80.27 & 68.56 & \cellcolor{lightgreen}72.91 \\
arc easy & \cellcolor{mediumgreen}89.82 & 83.51 & \cellcolor{lightgreen}84.74 \\
hellaswag & \cellcolor{mediumgreen}40.86 & 34.75 & \cellcolor{lightgreen}37.04 \\
piqa & \cellcolor{mediumgreen}80.03 & 64.58 & \cellcolor{lightgreen}75.24 \\
winogrande & \cellcolor{mediumgreen}54.22 & 53.04 & \cellcolor{lightgreen}54.14 \\
mmlu & \cellcolor{mediumgreen}61.66 & 55.13 & \cellcolor{lightgreen}57.41 \\
\hdashline
Avg. & \cellcolor{mediumgreen}67.81 & 59.93 & \cellcolor{lightgreen}63.58 \\
\bottomrule
\addlinespace[10pt]
\end{tabular}
\end{sc}
\caption{Evaluation results comparing different fine-tuning configurations across various tasks for \textbf{Slovenian} for \textsc{Mistral-Nemo-12B}.}
\label{tab:evaluation_results_sl_nemo}
\end{table}

\begin{table}[h!]
\subsection{Aya-Expanse-8B}
\vspace{10pt}
\small
\centering
\begin{sc}
\begin{tabular}{@{}lccc@{}}
\toprule
Metric & \textbf{Base} & \textbf{Random} & \textbf{Target} \\
\midrule
\multicolumn{4}{c}{\textbf{Monolingual Capabilities}} \\
\midrule
FLORES af$\rightarrow$en & 40.36 & \cellcolor{mediumgreen}49.28 & \cellcolor{lightgreen}48.73 \\
FLORES en$\rightarrow$af & 16.28 & \cellcolor{lightgreen}30.87 & \cellcolor{mediumgreen}32.80 \\
BELEBELE af & 76.89 & \cellcolor{lightgreen}81.00 & \cellcolor{mediumgreen}81.56 \\
SIB200 af & 42.16 & \cellcolor{lightgreen}49.51 & \cellcolor{mediumgreen}64.22 \\
OPUS af$\rightarrow$en & \cellcolor{lightgreen}25.06 & \cellcolor{mediumgreen}29.97 & 23.78 \\
OPUS en$\rightarrow$af & \cellcolor{mediumgreen}20.27 & \cellcolor{lightgreen}15.55 & 14.45 \\
\hdashline
Avg. & 36.84 & \cellcolor{lightgreen}42.70 & \cellcolor{mediumgreen}44.25 \\
\midrule
\multicolumn{4}{c}{\textbf{General Performance}} \\
\midrule
arc challenge & \cellcolor{lightgreen}77.26 & \cellcolor{mediumgreen}78.26 & 75.59 \\
arc easy & \cellcolor{lightgreen}90.00 & 88.95 & \cellcolor{mediumgreen}90.53 \\
hellaswag & \cellcolor{mediumgreen}56.15 & \cellcolor{lightgreen}43.21 & 40.11 \\
piqa & 77.48 & \cellcolor{lightgreen}78.35 & \cellcolor{mediumgreen}79.98 \\
winogrande & \cellcolor{mediumgreen}55.72 & \cellcolor{lightgreen}54.22 & 53.43 \\
mmlu & \cellcolor{mediumgreen}88.11 & 83.08 & \cellcolor{lightgreen}85.76 \\
\hdashline
Avg. & \cellcolor{mediumgreen}74.12 & \cellcolor{lightgreen}71.01 & 70.90 \\
\bottomrule
\addlinespace[10pt]
\end{tabular}
\end{sc}
\caption{Evaluation results comparing different fine-tuning configurations across various tasks for \textbf{Afrikaans} for \textsc{Aya-Expanse-8B}.}
\label{tab:evaluation_results_aya_af}
\end{table}

\begin{table}[h!]
\small
\centering
\begin{sc}
\begin{tabular}{@{}lccc@{}}
\toprule
Metric & \textbf{Base} & \textbf{Random} & \textbf{Target} \\
\midrule
\multicolumn{4}{c}{\textbf{Monolingual Capabilities}} \\
\midrule
FLORES cy$\rightarrow$en & 23.47 & \cellcolor{mediumgreen}45.21 & \cellcolor{lightgreen}36.58 \\
FLORES en$\rightarrow$cy & 7.16 & \cellcolor{lightgreen}12.22 & \cellcolor{mediumgreen}26.26 \\
SIB200 cy & 32.35 & \cellcolor{mediumgreen}67.65 & \cellcolor{lightgreen}58.33 \\
OPUS cy$\rightarrow$en & 13.26 & \cellcolor{mediumgreen}19.68 & \cellcolor{lightgreen}15.11 \\
OPUS en$\rightarrow$cy & \cellcolor{mediumgreen}9.39 & 3.46 & \cellcolor{lightgreen}6.87 \\
\hdashline
Avg. & 17.13 & \cellcolor{mediumgreen}29.64 & \cellcolor{lightgreen}28.63 \\
\midrule
\multicolumn{4}{c}{\textbf{General Performance}} \\
\midrule
arc challenge & \cellcolor{mediumgreen}77.26 & 72.24 & \cellcolor{lightgreen}76.25 \\
arc easy & \cellcolor{mediumgreen}90.00 & 87.72 & \cellcolor{lightgreen}88.25 \\
hellaswag & \cellcolor{mediumgreen}56.15 & 38.89 & \cellcolor{lightgreen}55.82 \\
piqa & 77.48 & \cellcolor{lightgreen}77.64 & \cellcolor{mediumgreen}81.23 \\
winogrande & 55.72 & \cellcolor{mediumgreen}56.12 & \cellcolor{lightgreen}56.04 \\
mmlu & \cellcolor{mediumgreen}88.11 & 77.14 & \cellcolor{lightgreen}83.80 \\
\hdashline
Avg. & \cellcolor{mediumgreen}74.12 & 68.29 & \cellcolor{lightgreen}73.56 \\
\bottomrule
\addlinespace[10pt]
\end{tabular}
\end{sc}
\caption{Evaluation results comparing different fine-tuning configurations across various tasks for \textbf{Welsh} for \textsc{Aya-Expanse-8B}.}
\label{tab:evaluation_results_aya_cy}
\end{table}

\begin{table}[h!]
\small
\centering
\begin{sc}
\begin{tabular}{@{}lccc@{}}
\toprule
Metric & \textbf{Base} & \textbf{Random} & \textbf{Target} \\
\midrule
\multicolumn{4}{c}{\textbf{Monolingual Capabilities}} \\
\midrule
FLORES et$\rightarrow$en & 10.84 & \cellcolor{mediumgreen}26.54 & \cellcolor{lightgreen}21.45 \\
FLORES en$\rightarrow$et & 2.72 & \cellcolor{mediumgreen}11.69 & \cellcolor{lightgreen}10.78 \\
BELEBELE et & 54.00 & \cellcolor{mediumgreen}71.78 & \cellcolor{lightgreen}68.78 \\
SIB200 et & 39.71 & \cellcolor{lightgreen}50.49 & \cellcolor{mediumgreen}66.67 \\
OPUS et$\rightarrow$en & 10.85 & \cellcolor{mediumgreen}23.49 & \cellcolor{lightgreen}20.14 \\
OPUS en$\rightarrow$et & 7.46 & \cellcolor{mediumgreen}16.16 & \cellcolor{lightgreen}13.85 \\
\hdashline
Avg. & 20.93 & \cellcolor{lightgreen}33.36 & \cellcolor{mediumgreen}33.61 \\
\midrule
\multicolumn{4}{c}{\textbf{General Performance}} \\
\midrule
arc challenge & \cellcolor{mediumgreen}77.26 & \cellcolor{lightgreen}73.91 & 73.24 \\
arc easy & \cellcolor{mediumgreen}90.00 & \cellcolor{lightgreen}89.12 & 87.54 \\
hellaswag & \cellcolor{mediumgreen}56.15 & \cellcolor{lightgreen}49.79 & 44.84 \\
piqa & \cellcolor{lightgreen}77.48 & 76.17 & \cellcolor{mediumgreen}80.47 \\
winogrande & \cellcolor{lightgreen}55.72 & 55.09 & \cellcolor{mediumgreen}57.30 \\
mmlu & \cellcolor{mediumgreen}88.11 & 82.04 & \cellcolor{lightgreen}84.78 \\
\hdashline
Avg. & \cellcolor{mediumgreen}74.12 & 71.02 & \cellcolor{lightgreen}71.36 \\
\bottomrule
\addlinespace[10pt]
\end{tabular}
\end{sc}
\caption{Evaluation results comparing different fine-tuning configurations across various tasks for \textbf{Estonian} for \textsc{Aya-Expanse-8B}.}
\label{tab:evaluation_results_aya_et}
\end{table}

\begin{table}[h!]
\small
\centering
\begin{sc}
\begin{tabular}{@{}lccc@{}}
\toprule
Metric & \textbf{Base} & \textbf{Random} & \textbf{Target} \\
\midrule
\multicolumn{4}{c}{\textbf{Monolingual Capabilities}} \\
\midrule
FLORES en$\rightarrow$is & 5.92 & \cellcolor{lightgreen}13.29 & \cellcolor{mediumgreen}13.83 \\
FLORES is $\rightarrow$en & 13.74 & \cellcolor{mediumgreen}25.97 & \cellcolor{lightgreen}24.59 \\
BELEBELE is & 53.22 & \cellcolor{lightgreen}70.00 & \cellcolor{mediumgreen}72.78 \\
SIB200 is & 33.33 & \cellcolor{lightgreen}59.80 & \cellcolor{mediumgreen}62.75 \\
OPUS en$\rightarrow$is & 8.98 & \cellcolor{mediumgreen}12.36 & \cellcolor{lightgreen}12.23 \\
OPUS is$\rightarrow$en & 11.21 & \cellcolor{mediumgreen}19.57 & \cellcolor{lightgreen}18.19 \\
\hdashline
Avg. & 21.07 & \cellcolor{lightgreen}33.50 & \cellcolor{mediumgreen}34.06 \\
\midrule
\multicolumn{4}{c}{\textbf{General Performance}} \\
\midrule
arc challenge & \cellcolor{lightgreen}77.26 & 75.59 & \cellcolor{mediumgreen}77.59 \\
arc easy & \cellcolor{mediumgreen}90.00 & 88.77 & \cellcolor{lightgreen}89.82 \\
hellaswag & \cellcolor{mediumgreen}56.15 & 38.54 & \cellcolor{lightgreen}45.24 \\
piqa & \cellcolor{lightgreen}77.48 & 76.55 & \cellcolor{mediumgreen}79.60 \\
winogrande & \cellcolor{mediumgreen}55.72 & \cellcolor{lightgreen}54.22 & 53.75 \\
mmlu & \cellcolor{mediumgreen}88.11 & 85.56 & \cellcolor{lightgreen}87.79 \\
\hdashline
Avg. & \cellcolor{mediumgreen}74.12 & 69.87 & \cellcolor{lightgreen}72.30 \\
\bottomrule
\addlinespace[10pt]
\end{tabular}
\end{sc}
\caption{Evaluation results comparing different fine-tuning configurations across various tasks for \textbf{Icelandic} for \textsc{Aya-Expanse-8B}.}
\label{tab:evaluation_results_aya_is}
\end{table}

\begin{table}[h!]
\small
\centering
\begin{sc}
\begin{tabular}{@{}lccc@{}}
\toprule
Metric & \textbf{Base} & \textbf{Random} & \textbf{Target} \\
\midrule
\multicolumn{4}{c}{\textbf{Monolingual Capabilities}} \\
\midrule
FLORES en$\rightarrow$ka & 3.48 & \cellcolor{lightgreen}5.52 & \cellcolor{mediumgreen}5.70 \\
FLORES ka$\rightarrow$en & 11.23 & \cellcolor{lightgreen}16.93 & \cellcolor{mediumgreen}16.98 \\
BELEBELE ka & 53.56 & \cellcolor{lightgreen}58.44 & \cellcolor{mediumgreen}60.44 \\
SIB200 ka & 48.04 & \cellcolor{mediumgreen}68.63 & \cellcolor{lightgreen}64.71 \\
OPUS en$\rightarrow$ka & 7.48 & \cellcolor{lightgreen}7.96 & \cellcolor{mediumgreen}8.39 \\
OPUS ka$\rightarrow$en & \cellcolor{mediumgreen}6.66 & 5.00 & \cellcolor{lightgreen}5.18 \\
\hdashline
Avg. & 21.74 & \cellcolor{mediumgreen}27.08 & \cellcolor{lightgreen}26.90 \\
\midrule
\multicolumn{4}{c}{\textbf{General Performance}} \\
\midrule
arc challenge & \cellcolor{lightgreen}77.26 & 73.58 & \cellcolor{mediumgreen}77.93 \\
arc easy & \cellcolor{mediumgreen}90.00 & 88.07 & \cellcolor{lightgreen}89.82 \\
hellaswag & \cellcolor{mediumgreen}56.15 & 44.69 & \cellcolor{lightgreen}52.88 \\
piqa & \cellcolor{lightgreen}77.48 & \cellcolor{mediumgreen}81.56 & \cellcolor{mediumgreen}81.56 \\
winogrande & \cellcolor{mediumgreen}55.72 & 52.96 & \cellcolor{lightgreen}54.85 \\
mmlu & \cellcolor{mediumgreen}88.11 & 84.00 & \cellcolor{lightgreen}87.07 \\
\hdashline
Avg. & \cellcolor{mediumgreen}74.12 & 70.81 & \cellcolor{lightgreen}74.02 \\
\bottomrule
\addlinespace[10pt]
\end{tabular}
\end{sc}
\caption{Evaluation results comparing different fine-tuning configurations across various tasks for \textbf{Georgian} for \textsc{Aya-Expanse-8B}.}
\label{tab:evaluation_results_aya_ka}
\end{table}

\begin{table}[h!]
\small
\centering
\begin{sc}
\begin{tabular}{@{}lccc@{}}
\toprule
Metric & \textbf{Base} & \textbf{Random} & \textbf{Target} \\
\midrule
\multicolumn{4}{c}{\textbf{Monolingual Capabilities}} \\
\midrule
FLORES en$\rightarrow$lt & 7.54 & \cellcolor{lightgreen}10.75 & \cellcolor{mediumgreen}13.96 \\
FLORES lt$\rightarrow$en & 17.22 & \cellcolor{lightgreen}21.92 & \cellcolor{mediumgreen}24.08 \\
BELEBELE lt & 70.22 & \cellcolor{lightgreen}74.67 & \cellcolor{mediumgreen}76.22 \\
SIB200 lt & \cellcolor{lightgreen}45.59 & 43.63 & \cellcolor{mediumgreen}46.08 \\
OPUS en$\rightarrow$lt & \cellcolor{lightgreen}12.41 & 11.20 & \cellcolor{mediumgreen}15.35 \\
OPUS lt$\rightarrow$en & 18.36 & \cellcolor{lightgreen}20.52 & \cellcolor{mediumgreen}24.22 \\
\hdashline
Avg. & 28.56 & \cellcolor{lightgreen}30.45 & \cellcolor{mediumgreen}33.32 \\
\midrule
\multicolumn{4}{c}{\textbf{General Performance}} \\
\midrule
arc challenge & \cellcolor{mediumgreen}77.26 & 73.91 & \cellcolor{lightgreen}74.92 \\
arc easy & \cellcolor{mediumgreen}90.00 & 86.84 & \cellcolor{lightgreen}87.54 \\
hellaswag & \cellcolor{mediumgreen}56.15 & 46.50 & \cellcolor{lightgreen}48.36 \\
piqa & 77.48 & \cellcolor{lightgreen}78.78 & \cellcolor{mediumgreen}80.69 \\
winogrande & \cellcolor{lightgreen}55.72 & 53.35 & \cellcolor{mediumgreen}55.88 \\
mmlu & \cellcolor{mediumgreen}88.11 & 82.95 & \cellcolor{lightgreen}86.28 \\
\hdashline
Avg. & \cellcolor{mediumgreen}74.12 & 70.39 & \cellcolor{lightgreen}72.28 \\
\bottomrule
\addlinespace[10pt]
\end{tabular}
\end{sc}
\caption{Evaluation results comparing different fine-tuning configurations across various tasks for \textbf{Lithuanian} for \textsc{Aya-Expanse-8B}.}
\label{tab:evaluation_results_aya_lt}
\end{table}

\begin{table}[h!]
\small
\centering
\begin{sc}
\begin{tabular}{@{}lccc@{}}
\toprule
Metric & \textbf{Base} & \textbf{Random} & \textbf{Target} \\
\midrule
\multicolumn{4}{c}{\textbf{Monolingual Capabilities}} \\
\midrule
FLORES en$\rightarrow$lv & 4.92 & \cellcolor{lightgreen}15.91 & \cellcolor{mediumgreen}16.22 \\
FLORES lv$\rightarrow$en & 11.97 & \cellcolor{mediumgreen}27.16 & \cellcolor{lightgreen}26.43 \\
BELEBELE lv & 56.56 & \cellcolor{mediumgreen}77.33 & \cellcolor{lightgreen}75.78 \\
SIB200 lv & 42.16 & \cellcolor{lightgreen}62.75 & \cellcolor{mediumgreen}67.16 \\
OPUS en$\rightarrow$lv & 9.37 & \cellcolor{mediumgreen}17.79 & \cellcolor{lightgreen}17.28 \\
OPUS lv$\rightarrow$en & 14.25 & \cellcolor{mediumgreen}28.05 & \cellcolor{lightgreen}25.31 \\
\hdashline
Avg. & 23.21 & \cellcolor{mediumgreen}38.17 & \cellcolor{lightgreen}38.03 \\
\midrule
\multicolumn{4}{c}{\textbf{General Performance}} \\
\midrule
arc challenge & \cellcolor{lightgreen}77.26 & 69.23 & \cellcolor{mediumgreen}77.93 \\
arc easy & \cellcolor{mediumgreen}90.00 & 86.14 & \cellcolor{lightgreen}89.47 \\
hellaswag & \cellcolor{mediumgreen}56.15 & 42.04 & \cellcolor{lightgreen}52.23 \\
piqa & \cellcolor{lightgreen}77.48 & 77.26 & \cellcolor{mediumgreen}81.88 \\
winogrande & \cellcolor{mediumgreen}55.72 & \cellcolor{lightgreen}55.09 & 54.85 \\
mmlu & \cellcolor{mediumgreen}88.11 & 81.71 & \cellcolor{lightgreen}87.20 \\
\hdashline
Avg. & \cellcolor{mediumgreen}74.12 & 68.58 & \cellcolor{lightgreen}73.93 \\
\bottomrule
\addlinespace[10pt]
\end{tabular}
\end{sc}
\caption{Evaluation results comparing different fine-tuning configurations across various tasks for \textbf{Latvian} for \textsc{Aya-Expanse-8B}.}
\label{tab:evaluation_results_aya_lv}
\end{table}

\begin{table}[h!]
\small
\centering
\begin{sc}
\begin{tabular}{@{}lccc@{}}
\toprule
Metric & \textbf{Base} & \textbf{Random} & \textbf{Target} \\
\midrule
\multicolumn{4}{c}{\textbf{Monolingual Capabilities}} \\
\midrule
FLORES en$\rightarrow$mk & 6.30 & \cellcolor{lightgreen}21.20 & \cellcolor{mediumgreen}23.28 \\
FLORES mk$\rightarrow$en & 23.74 & \cellcolor{mediumgreen}34.85 & \cellcolor{lightgreen}34.69 \\
BELEBELE mk & 67.33 & \cellcolor{lightgreen}77.22 & \cellcolor{mediumgreen}78.67 \\
SIB200 mk & 43.14 & \cellcolor{lightgreen}66.67 & \cellcolor{mediumgreen}70.10 \\
OPUS en$\rightarrow$mk & 11.13 & \cellcolor{lightgreen}19.50 & \cellcolor{mediumgreen}20.31 \\
OPUS mk$\rightarrow$en & 17.79 & \cellcolor{mediumgreen}26.71 & \cellcolor{lightgreen}26.39 \\
\hdashline
Avg. & 28.24 & \cellcolor{lightgreen}41.03 & \cellcolor{mediumgreen}42.24 \\
\midrule
\multicolumn{4}{c}{\textbf{General Performance}} \\
\midrule
arc challenge & \cellcolor{lightgreen}77.26 & 75.59 & \cellcolor{mediumgreen}78.60 \\
arc easy & \cellcolor{mediumgreen}90.00 & \cellcolor{lightgreen}87.89 & \cellcolor{mediumgreen}90.00 \\
hellaswag & \cellcolor{mediumgreen}56.15 & \cellcolor{lightgreen}48.53 & 43.78 \\
piqa & 77.48 & \cellcolor{lightgreen}78.18 & \cellcolor{mediumgreen}82.05 \\
winogrande & \cellcolor{mediumgreen}55.72 & \cellcolor{lightgreen}55.56 & 54.30 \\
mmlu & \cellcolor{lightgreen}88.11 & 84.39 & \cellcolor{mediumgreen}88.18 \\
\hdashline
Avg. & \cellcolor{mediumgreen}74.12 & 71.69 & \cellcolor{lightgreen}72.82 \\
\bottomrule
\addlinespace[10pt]
\end{tabular}
\end{sc}
\caption{Evaluation results comparing different fine-tuning configurations across various tasks for \textbf{Macedonian} for \textsc{Aya-Expanse-8B}.}
\label{tab:evaluation_results_aya_mk}
\end{table}

\begin{table}[h!]
\small
\centering
\begin{sc}
\begin{tabular}{@{}lccc@{}}
\toprule
Metric & \textbf{Base} & \textbf{Random} & \textbf{Target} \\
\midrule
\multicolumn{4}{c}{\textbf{Monolingual Capabilities}} \\
\midrule
FLORES en$\rightarrow$mt & 4.71 & \cellcolor{lightgreen}21.74 & \cellcolor{mediumgreen}22.90 \\
FLORES mt$\rightarrow$en & 14.45 & \cellcolor{mediumgreen}45.43 & \cellcolor{lightgreen}42.55 \\
BELEBELE mt & 43.56 & \cellcolor{mediumgreen}74.78 & \cellcolor{lightgreen}74.56 \\
SIB200 mt & 45.59 & \cellcolor{lightgreen}59.80 & \cellcolor{mediumgreen}68.63 \\
OPUS en$\rightarrow$mt & 8.93 & \cellcolor{lightgreen}20.44 & \cellcolor{mediumgreen}20.77 \\
OPUS mt$\rightarrow$en & 20.61 & \cellcolor{mediumgreen}40.35 & \cellcolor{lightgreen}37.40 \\
\hdashline
Avg. & 22.97 & \cellcolor{lightgreen}43.76 & \cellcolor{mediumgreen}44.47 \\
\midrule
\multicolumn{4}{c}{\textbf{General Performance}} \\
\midrule
arc challenge & \cellcolor{mediumgreen}77.26 & \cellcolor{lightgreen}76.92 & 74.58 \\
arc easy & \cellcolor{mediumgreen}90.00 & \cellcolor{lightgreen}87.89 & \cellcolor{lightgreen}87.89 \\
hellaswag & \cellcolor{mediumgreen}56.15 & 39.55 & \cellcolor{lightgreen}42.98 \\
piqa & 77.48 & \cellcolor{lightgreen}79.27 & \cellcolor{mediumgreen}79.76 \\
winogrande & \cellcolor{mediumgreen}55.72 & 53.28 & \cellcolor{lightgreen}55.09 \\
mmlu & \cellcolor{mediumgreen}88.11 & 84.32 & \cellcolor{lightgreen}85.50 \\
\hdashline
Avg. & \cellcolor{mediumgreen}74.12 & 70.21 & \cellcolor{lightgreen}70.97 \\
\bottomrule
\addlinespace[10pt]
\end{tabular}
\end{sc}
\caption{Evaluation results comparing different fine-tuning configurations across various tasks for \textbf{Maltese} for \textsc{Aya-Expanse-8B}.}
\label{tab:evaluation_results_aya_mt}
\end{table}

\begin{table}[h!]
\small
\centering
\begin{sc}
\begin{tabular}{@{}lccc@{}}
\toprule
Metric & \textbf{Base} & \textbf{Random} & \textbf{Target} \\
\midrule
\multicolumn{4}{c}{\textbf{Monolingual Capabilities}} \\
\midrule
FLORES en$\rightarrow$ne & 2.86 & \cellcolor{mediumgreen}8.58 & \cellcolor{lightgreen}8.45 \\
FLORES ne$\rightarrow$en & 13.20 & \cellcolor{lightgreen}25.73 & \cellcolor{mediumgreen}25.87 \\
BELEBELE ne & 45.89 & \cellcolor{lightgreen}63.11 & \cellcolor{mediumgreen}66.00 \\
SIB200 ne & 47.55 & \cellcolor{lightgreen}65.20 & \cellcolor{mediumgreen}69.12 \\
OPUS en$\rightarrow$ne & 11.33 & \cellcolor{mediumgreen}13.17 & \cellcolor{lightgreen}13.14 \\
OPUS ne$\rightarrow$en & 22.92 & \cellcolor{mediumgreen}28.42 & \cellcolor{lightgreen}26.62 \\
\hdashline
Avg. & 23.96 & \cellcolor{lightgreen}34.03 & \cellcolor{mediumgreen}34.87 \\
\midrule
\multicolumn{4}{c}{\textbf{General Performance}} \\
\midrule
arc challenge & \cellcolor{lightgreen}77.26 & 73.24 & \cellcolor{mediumgreen}79.26 \\
arc easy & \cellcolor{lightgreen}90.00 & 87.72 & \cellcolor{mediumgreen}90.88 \\
hellaswag & \cellcolor{mediumgreen}56.15 & 41.11 & \cellcolor{lightgreen}49.52 \\
piqa & 77.48 & \cellcolor{lightgreen}81.28 & \cellcolor{mediumgreen}82.64 \\
winogrande & \cellcolor{mediumgreen}55.72 & 53.99 & \cellcolor{lightgreen}54.85 \\
mmlu & \cellcolor{lightgreen}88.11 & 85.50 & \cellcolor{mediumgreen}88.96 \\
\hdashline
Avg. & \cellcolor{lightgreen}74.12 & 70.47 & \cellcolor{mediumgreen}74.35 \\
\bottomrule
\addlinespace[10pt]
\end{tabular}
\end{sc}
\caption{Evaluation results comparing different fine-tuning configurations across various tasks for \textbf{Nepali} for \textsc{Aya-Expanse-8B}.}
\label{tab:evaluation_results}
\end{table}

\begin{table}[h!]
\small
\centering
\begin{sc}
\begin{tabular}{@{}lccc@{}}
\toprule
Metric & \textbf{Base} & \textbf{Random} & \textbf{Target} \\
\midrule
\multicolumn{4}{c}{\textbf{Monolingual Capabilities}} \\
\midrule
FLORES en$\rightarrow$sk & 14.46 & \cellcolor{mediumgreen}19.86 & \cellcolor{lightgreen}18.96 \\
FLORES sk$\rightarrow$en & 30.20 & \cellcolor{lightgreen}31.04 & \cellcolor{mediumgreen}31.82 \\
BELEBELE sk & \cellcolor{lightgreen}80.56 & 79.33 & \cellcolor{mediumgreen}81.00 \\
SIB200 sk & \cellcolor{lightgreen}52.94 & 42.16 & \cellcolor{mediumgreen}54.90 \\
OPUS en$\rightarrow$sk & 17.36 & \cellcolor{lightgreen}18.63 & \cellcolor{mediumgreen}19.39 \\
OPUS sk$\rightarrow$en & 26.00 & \cellcolor{lightgreen}26.64 & \cellcolor{mediumgreen}27.89 \\
\hdashline
Avg. & \cellcolor{lightgreen}36.92 & 36.28 & \cellcolor{mediumgreen}38.99 \\
\midrule
\multicolumn{4}{c}{\textbf{General Performance}} \\
\midrule
arc challenge & \cellcolor{lightgreen}77.26 & \cellcolor{mediumgreen}77.59 & 74.92 \\
arc easy & \cellcolor{mediumgreen}90.00 & 87.72 & \cellcolor{lightgreen}88.77 \\
hellaswag & \cellcolor{mediumgreen}56.15 & \cellcolor{lightgreen}52.13 & 48.54 \\
piqa & 77.48 & \cellcolor{mediumgreen}79.38 & \cellcolor{lightgreen}79.16 \\
winogrande & \cellcolor{mediumgreen}55.72 & 53.99 & \cellcolor{lightgreen}54.70 \\
mmlu & \cellcolor{mediumgreen}88.11 & 83.87 & \cellcolor{lightgreen}86.02 \\
\hdashline
Avg. & \cellcolor{mediumgreen}74.12 & \cellcolor{lightgreen}72.45 & 72.02 \\
\bottomrule
\addlinespace[10pt]
\end{tabular}
\end{sc}
\caption{Evaluation results comparing different fine-tuning configurations across various tasks for \textbf{Slovak} for \textsc{Aya-Expanse-8B}.}
\label{tab:evaluation_results_aya_sk}
\end{table}

\begin{table}[h!]
\small
\centering
\begin{sc}
\begin{tabular}{@{}lccc@{}}
\toprule
Metric & \textbf{Base} & \textbf{Random} & \textbf{Target} \\
\midrule
\multicolumn{4}{c}{\textbf{Monolingual Capabilities}} \\
\midrule
FLORES en$\rightarrow$sl & 8.27 & \cellcolor{lightgreen}15.66 & \cellcolor{mediumgreen}17.50 \\
FLORES sl$\rightarrow$en & 20.22 & \cellcolor{mediumgreen}27.56 & \cellcolor{lightgreen}27.09 \\
BELEBELE sl & 67.22 & \cellcolor{lightgreen}78.56 & \cellcolor{mediumgreen}80.22 \\
SIB200 sl & 40.69 & \cellcolor{lightgreen}49.51 & \cellcolor{mediumgreen}54.90 \\
OPUS en$\rightarrow$sl & 11.25 & \cellcolor{lightgreen}13.93 & \cellcolor{mediumgreen}15.31 \\
OPUS sl$\rightarrow$en & 14.66 & \cellcolor{mediumgreen}21.80 & \cellcolor{lightgreen}20.97 \\
\hdashline
Avg. & 27.05 & \cellcolor{lightgreen}34.50 & \cellcolor{mediumgreen}36.00 \\
\midrule
\multicolumn{4}{c}{\textbf{General Performance}} \\
\midrule
arc challenge & \cellcolor{lightgreen}77.26 & 69.23 & \cellcolor{mediumgreen}78.60 \\
arc easy & \cellcolor{mediumgreen}90.00 & 85.79 & \cellcolor{lightgreen}89.82 \\
hellaswag & \cellcolor{mediumgreen}56.15 & 47.88 & \cellcolor{lightgreen}51.96 \\
piqa & \cellcolor{lightgreen}77.48 & 76.61 & \cellcolor{mediumgreen}79.87 \\
winogrande & \cellcolor{lightgreen}55.72 & 54.93 & \cellcolor{mediumgreen}57.46 \\
mmlu & \cellcolor{mediumgreen}88.11 & 80.54 & \cellcolor{lightgreen}86.81 \\
\hdashline
Avg. & \cellcolor{mediumgreen}74.12 & 69.16 & \cellcolor{lightgreen}74.09 \\
\bottomrule
\addlinespace[10pt]
\end{tabular}
\end{sc}
\caption{Evaluation results comparing different fine-tuning configurations across various tasks for \textbf{Slovenian} for \textsc{Aya-Expanse-8B}.}
\label{tab:evaluation_results_aya_sl}
\end{table}

\clearpage

\section{Generation Prompts}
\label{app:generation}

\subsection{Task Prompts}

We use standardized prompt templates for each task category to ensure consistency across models and datasets. Below we show the representative formats. Placeholders such as \texttt{\{source\_text\}}, \texttt{\{passage\}}, or \texttt{\{question\}} are replaced with the actual content at runtime.

\paragraph{Translation.}
\begin{quote}
\texttt{Translate this sentence into \{target\_name\}: \{source\_text\} Translation:}
\end{quote}

\paragraph{Classification.}
\begin{quote}
\texttt{Classify the following text into one of these topics:}\\
\texttt{A) Science/Technology}\\
\texttt{B) Travel}\\
\texttt{C) Politics}\\
\texttt{D) Sports}\\
\texttt{E) Health}\\
\texttt{F) Entertainment}\\
\texttt{G) Geography}\\

\texttt{Text: \{text\}}\\

\texttt{Answer:}
\end{quote}

\paragraph{Comprehension.}
\begin{quote}
\texttt{Read the following passage and pick the right answer.}\\

\texttt{Passage: \{passage\}}\\

\texttt{Question: \{question\}}\\

\texttt{\{options\}}\\

\texttt{Answer:}
\end{quote}

\paragraph{General Capabilities.}
We adopt dataset-specific templates for MMLU and commonsense reasoning benchmarks:

\begin{itemize}
    \item \textbf{MMLU:}
    \begin{quote}
    \texttt{Answer the following multiple choice question:}\\

    \texttt{\{question\}}\\

    \texttt{\{options\}}\\

    \texttt{Answer:}
    \end{quote}

    \item \textbf{HellaSwag:}
    \begin{quote}
    \texttt{Complete the following scenario by choosing the most likely continuation:}\\

    \texttt{\{ctx\}}\\

    \texttt{\{options\}}\\

    \texttt{Answer:}
    \end{quote}

    \item \textbf{PIQA:}
    \begin{quote}
    \texttt{Choose the most appropriate solution for the following goal:}\\

    \texttt{Goal: \{goal\}}\\

    \texttt{A) \{sol1\}}\\
    \texttt{B) \{sol2\}}\\

    \texttt{Answer:}
    \end{quote}

    \item \textbf{Winogrande:}
    \begin{quote}
    \texttt{Complete the sentence by choosing the correct option:}\\

    \texttt{\{sentence\}}\\

    \texttt{A) \{option1\}}\\
    \texttt{B) \{option2\}}\\

    \texttt{Answer:}
    \end{quote}

    \item \textbf{ARC (Easy/Challenge):}
    \begin{quote}
    \texttt{Answer the following question:}\\

    \texttt{\{question\}}\\

    \texttt{\{options\}}\\

    \texttt{Answer:}
    \end{quote}
\end{itemize}
\clearpage

\subsection{Generation Settings}

The generation is done using the following parameters:

\begin{verbatim}
SamplingParams(
    temperature=0,
    repetition_penalty=1.1,
    max_tokens=128,
    top_p=1.0,
    top_k=-1,
    stop_token_ids=[eos_token_id] if eos_token_id is not None 
            else [],
    stop=[
        "\n", "\n\n",
        "Question:", "Q:",
        "Answer:", "A:",
        "Translation:", "The translation", "Translate:",
        "Summary:", "The summary", "Summarize:",
        "Passage:", "The passage", "Text:",
        "Options:", "The options", "The choices", "Choices:",
        "Context:", "The context", "The problem", "Problem:",
        "\n\nQuestion", "\n\nAnswer",
        "</s>", "<|endoftext|>",
        "Human:", "Assistant:"
    ],
    skip_special_tokens=True
)
\end{verbatim}

\onecolumn
\begin{table*}[t!]
\section{FFN Component Ablations}
\label{app:ffn_ablations}
To understand which parts of the FFN contribute most to language-specific adaptation, we perform ablations by fine-tuning individual or combinations of the up-, gate-, and down-projection matrices of the identified language subnetworks. Table~\ref{tab:ffn_parts} reports the resulting performance across a subset of six languages.

\vspace{25pt}
\small
\centering
\begin{sc}
\begin{tabular}{lcccccc||c}
\toprule
\textbf{FFN component} & \textbf{mt} & \textbf{af} & \textbf{sk} & \textbf{mk} & \textbf{lv} & \textbf{ne} & \textbf{Avg.} \\
{\scriptsize Full subnetwork size} & {\scriptsize 37.0M} & {\scriptsize 31.7M} & {\scriptsize 23.8M} & {\scriptsize 21.4M} & {\scriptsize 39.5M} & {\scriptsize 56.8M} \\
\midrule
base & 30.04 & 37.80 & 33.43 & 33.32 & 29.88 & 27.06 & 31.92 \\
ffn (up+gate+down) & \textbf{39.01} & 42.99 & \textbf{35.56} & 34.98 & 32.20 & \textbf{32.65} & \textbf{36.23} \\
\hdashline
up & 37.17 & 39.98 & 33.48 & 35.27 & 30.83 & 30.69 & 34.57 \\
gate+up & 38.14 & 42.18 & 34.81 & 34.13 & 31.28 & 27.23 & 34.63 \\
down & 34.99 & 40.60 & 35.31 & \textbf{36.57} & 30.36 & 28.66 & 34.42 \\
up+down & 37.13 & \textbf{44.11} & 35.00 & 35.28 & \textbf{33.71} & 29.72 & 35.83 \\
\bottomrule
\addlinespace[10pt]
\end{tabular}
\end{sc}
\caption{Average performance of fine-tuning different FFN components of \textsc{Llama-3.1-8B} across six languages. \emph{up}, \emph{gate}, and \emph{down} denote the identified language-specific components within the FFN up-, gate-, and down-projection weight matrices, respectively. Fine-tuning a single matrix corresponds to one third of the full language subnetwork size, while fine-tuning two matrices corresponds to two thirds.}
\label{tab:ffn_parts}
\end{table*}
\clearpage

\onecolumn
\section{Correlation with Language Resource Availability}
\label{app:resource-correlation}

To examine whether the observed improvements from targeted fine-tuning are systematically related to language resource availability, we analyze correlations using two complementary metrics.

\paragraph{CC-100 Corpus Size.} We use the CC-100 corpus size (in MB) as a proxy for the relative digital resource availability across languages \citep{conneau-etal-2020-unsupervised}. While these values represent web-crawled text from a specific snapshot, they serve as reasonable approximations of the relative proportions of available internet data per language. The ratios between languages remain informative even if absolute quantities have changed over time.

\paragraph{Joshi Resource Class.} We additionally employ the taxonomy proposed by \citet{joshi-etal-2020-state}, which classifies languages into six categories (0--5) based on both labeled and unlabeled data availability. This discrete classification considers multiple factors beyond raw corpus size, including the availability of annotated datasets and representation in NLP research. For our target languages, resource classes range from 1 (Maltese) to 3 (Estonian, Lithuanian, Latvian, Slovak, Slovenian).

\begin{figure*}[h]
    \centering

    \includegraphics[width=0.29\textwidth]{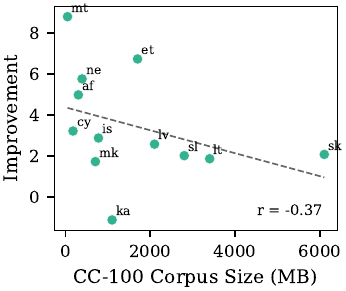}
    \hfill
    \includegraphics[width=0.29\textwidth]{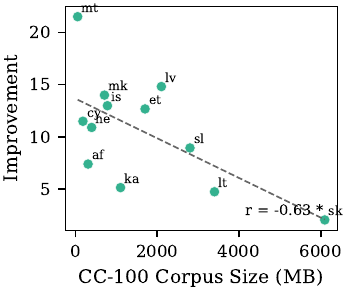}
    \hfill
    \includegraphics[width=0.29\textwidth]{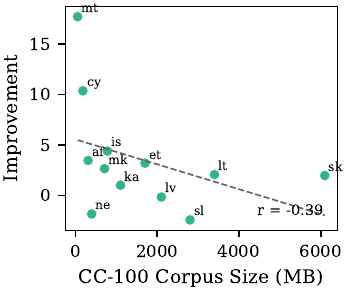}

    \vspace{0.15cm}

    \includegraphics[width=0.29\textwidth]{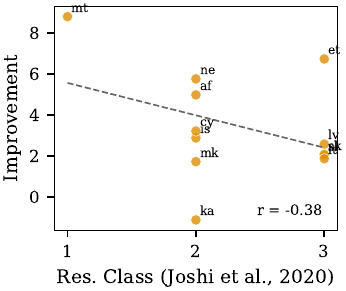}
    \hfill
    \includegraphics[width=0.29\textwidth]{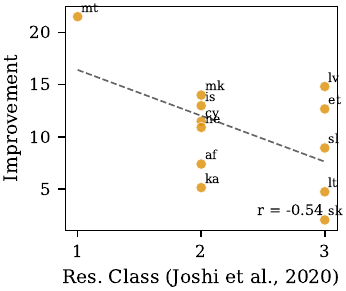}
    \hfill
    \includegraphics[width=0.29\textwidth]{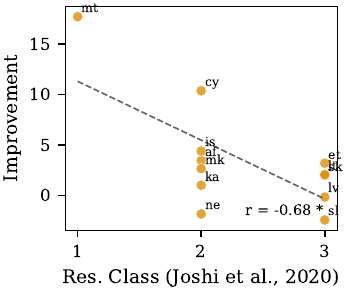}

    \vspace{0.15cm}

    \makebox[0.29\textwidth][c]{\texttt{LLaMA-3.1-8B}}
    \hfill
    \makebox[0.29\textwidth][c]{\texttt{Aya-Expanse-8B}}
    \hfill
    \makebox[0.29\textwidth][c]{\texttt{Mistral-NeMo-12B}}

    \vspace{0.15cm}

    \caption{Correlation between monolingual improvement scores and language resource metrics.
    \textbf{Top row:} CC-100 corpus size in MB, used as a proxy for relative internet data availability \citep{conneau-etal-2020-unsupervised}.
    \textbf{Bottom row:} Resource class from the Joshi et al.\ taxonomy \citep{joshi-etal-2020-state}.
    Pearson correlation coefficients are shown; * indicates $p < 0.05$. Performance gains from targeted fine-tuning are highest for lower-resource languages and diminish as resource availability increases, suggesting limited marginal benefit for languages already well represented during pre-training.}
    \label{fig:resource-correlation}
\end{figure*}

\clearpage
\begin{table*}[t]
\section{Cross-lingual Alignment}

\noindent
We additionally provide detailedper-language cosine similarities between parallel sentences before and after fine-tuning for all models. These results complement Section~\ref{sec:results} by illustrating how targeted subnetwork adaptation affects cross-lingual alignment across languages.

\vspace{10pt}
\label{app:cross-lingual}
\centering
\small
\begin{sc}
\begin{tabular}{@{}lccccccccc@{}}
\toprule
 & ar & de & en & es & fr & it & ja & ru & zh \\
\midrule
Baseline & 85.55 & 92.88 & 90.63 & 91.47 & 91.21 & 91.45 & 86.55 & 88.21 & 87.46 \\
Full & 86.73 & 91.90 & 94.65 & 91.14 & 91.48 & 89.98 & 86.10 & 87.36 & 88.83 \\
MLP & 84.22 & 90.39 & 94.05 & 89.85 & 90.39 & 89.62 & 85.06 & 87.48 & 88.19 \\
Adapter & 84.79 & 95.11 & 92.10 & 93.23 & 93.21 & 93.07 & 86.58 & 88.59 & 87.47 \\
Random & 86.76 & 95.06 & 92.94 & 92.84 & 92.95 & 92.66 & 88.15 & 90.18 & 89.31 \\
Target & \cellcolor{mediumgreen}87.75 & \cellcolor{mediumgreen}96.10 & \cellcolor{mediumgreen}95.03 & \cellcolor{mediumgreen}94.24 & \cellcolor{mediumgreen}94.52 & \cellcolor{mediumgreen}93.89 & \cellcolor{mediumgreen}90.24 & \cellcolor{mediumgreen}91.25 & \cellcolor{mediumgreen}91.46 \\
\bottomrule
\addlinespace[10pt]
\end{tabular}
\end{sc}
\caption{Layer-averaged Cosine similarity (expressed as percentages) between the target language (\textbf{Afrikaans}) and other languages for \textsc{Llama-3.1-8B}.}
\label{tab:cosine_similarity_af}
\end{table*}

\begin{table*}[t]
\centering
\small
\begin{sc}
\begin{tabular}{@{}lccccccccc@{}}
\toprule
 & ar & de & en & es & fr & it & ja & ru & zh \\
\midrule
Baseline & 83.22 & 87.82 & 85.92 & 87.72 & 87.58 & 88.11 & 83.09 & 84.76 & 83.24 \\
Full & 88.11 & 90.77 & 92.58 & 91.70 & 91.88 & 92.55 & 87.44 & 87.37 & 89.85 \\
MLP & 84.88 & 90.89 & 91.98 & 89.65 & 90.63 & 90.65 & 85.80 & 87.55 & 88.96 \\
Adapter & 88.13 & 93.40 & 90.40 & 93.30 & 93.29 & 94.08 & 86.11 & 89.41 & 86.25 \\
Random & 89.94 & 92.71 & 89.41 & 92.48 & 92.39 & 92.13 & 86.22 & 89.49 & 86.88 \\
Target & \cellcolor{mediumgreen}94.35 & \cellcolor{mediumgreen}95.49 & \cellcolor{mediumgreen}93.20 & \cellcolor{mediumgreen}94.90 & \cellcolor{mediumgreen}95.57 & \cellcolor{mediumgreen}95.62 & \cellcolor{mediumgreen}91.70 & \cellcolor{mediumgreen}92.71 & \cellcolor{mediumgreen}91.57 \\
\bottomrule
\addlinespace[10pt]
\end{tabular}
\end{sc}
\caption{Layer-averaged Cosine similarity (expressed as percentages) between the target language (\textbf{Welsh}) and other languages for \textsc{Llama-3.1-8B}.}
\label{tab:cosine_similarity_cy}
\end{table*}

\begin{table*}[t]
\centering
\small
\begin{sc}
\begin{tabular}{@{}lccccccccc@{}}
\toprule
 & ar & de & en & es & fr & it & ja & ru & zh \\
\midrule
Baseline & 85.45 & 91.29 & 89.18 & 90.46 & 90.19 & 90.67 & 86.09 & 88.40 & 86.75 \\
Full & 83.70 & 91.91 & 93.61 & 91.33 & 91.97 & 93.14 & 85.91 & 87.06 & 89.27 \\
MLP & 80.59 & 90.15 & 92.08 & 87.14 & 88.69 & 89.03 & 82.08 & 86.85 & 85.91 \\
Adapter & 86.19 & 93.87 & 90.52 & 92.56 & 92.51 & 92.83 & 86.99 & 89.58 & 87.47 \\
Random & 84.61 & 93.27 & 89.25 & 91.13 & 91.01 & 91.38 & 84.89 & 89.54 & 85.50 \\
Target & \cellcolor{mediumgreen}90.07 & \cellcolor{mediumgreen}95.43 & \cellcolor{mediumgreen}93.87 & \cellcolor{mediumgreen}94.13 & \cellcolor{mediumgreen}93.91 & \cellcolor{mediumgreen}93.73 & \cellcolor{mediumgreen}90.77 & \cellcolor{mediumgreen}92.25 & \cellcolor{mediumgreen}91.84 \\
\bottomrule
\addlinespace[10pt]
\end{tabular}
\end{sc}
\caption{Layer-averaged Cosine similarity (expressed as percentages) between the target language (\textbf{Estonian}) and other languages for \textsc{Llama-3.1-8B}.}
\label{tab:cosine_similarity_es}
\end{table*}

\begin{table*}[t]
\centering
\small
\begin{sc}
\begin{tabular}{@{}lccccccccc@{}}
\toprule
 & ar & de & en & es & fr & it & ja & ru & zh \\
\midrule
Baseline & 82.82 & 88.54 & 85.85 & 87.65 & 87.45 & 87.89 & 83.22 & 85.24 & 83.55 \\
Full & 89.21 & 90.85 & 92.80 & 90.94 & 90.99 & 88.86 & 88.24 & 88.52 & 90.31 \\
MLP & 80.94 & 90.20 & 90.23 & 88.49 & 88.76 & 89.58 & 78.67 & 84.65 & 82.93 \\
Adapter & 85.49 & \cellcolor{mediumgreen}93.79 & 89.50 & 92.61 & 92.24 & 92.57 & 85.22 & 88.89 & 86.21 \\
Random & 87.74 & 91.14 & 89.89 & 91.65 & 90.48 & 90.17 & 85.29 & 91.03 & 87.21 \\
Target & \cellcolor{mediumgreen}91.14 & 93.44 & \cellcolor{mediumgreen}93.11 & \cellcolor{mediumgreen}94.09 & \cellcolor{mediumgreen}94.10 & \cellcolor{mediumgreen}94.25 & \cellcolor{mediumgreen}90.11 & \cellcolor{mediumgreen}91.98 & \cellcolor{mediumgreen}90.65 \\
\bottomrule
\addlinespace[10pt]
\end{tabular}
\end{sc}
\caption{Layer-averaged Cosine similarity (expressed as percentages) between the target language (\textbf{Icelandic}) and other languages for \textsc{Llama-3.1-8B}.}
\label{tab:cosine_similarity_is}
\end{table*}

\begin{table*}[t]
\centering
\small
\begin{sc}
\begin{tabular}{@{}lccccccccc@{}}
\toprule
 & ar & de & en & es & fr & it & ja & ru & zh \\
\midrule
Baseline & 66.64 & 68.23 & 61.90 & 66.87 & 67.40 & 68.37 & 66.58 & 70.69 & 64.28 \\
Full & 64.74 & 64.17 & 57.36 & 64.37 & 63.03 & 64.32 & 63.36 & 66.32 & 61.19 \\
MLP & 67.84 & 69.05 & 61.96 & 67.05 & 67.74 & 68.95 & 67.34 & 69.48 & 65.94 \\
Adapter & 65.52 & 67.76 & 61.55 & 66.33 & 66.81 & 67.67 & 65.41 & 70.47 & 63.65 \\
Random & 67.02 & 69.26 & 62.24 & 67.80 & 68.14 & 69.37 & 66.86 & 71.79 & 65.19 \\
Target & \cellcolor{mediumgreen}69.54 & \cellcolor{mediumgreen}70.67 & \cellcolor{mediumgreen}64.23 & \cellcolor{mediumgreen}69.01 & \cellcolor{mediumgreen}69.66 & \cellcolor{mediumgreen}70.64 & \cellcolor{mediumgreen}69.20 & \cellcolor{mediumgreen}73.41 & \cellcolor{mediumgreen}67.32 \\
\bottomrule
\addlinespace[10pt]
\end{tabular}
\end{sc}
\caption{Layer-averaged Cosine similarity (expressed as percentages) between the target language (\textbf{Georgian}) and other languages for \textsc{Llama-3.1-8B}.}
\label{tab:cosine_similarity_ka}
\end{table*}

\begin{table*}[t]
\centering
\small
\begin{sc}
\begin{tabular}{@{}lccccccccc@{}}
\toprule
 & ar & de & en & es & fr & it & ja & ru & zh \\
\midrule
Baseline & 83.79 & 89.55 & 86.95 & 89.12 & 88.88 & 89.56 & 84.84 & 87.59 & 85.20 \\
Full & 86.30 & 92.68 & \cellcolor{mediumgreen}92.10 & 91.05 & 91.04 & 91.40 & 88.21 & 87.89 & 89.71 \\
MLP & 82.28 & 91.34 & 91.10 & 89.48 & 89.88 & 90.35 & 85.49 & 86.40 & 87.62 \\
Adapter & 83.45 & 91.75 & 88.31 & 91.17 & 90.83 & 91.67 & 84.16 & 88.73 & 84.43 \\
Random & 83.26 & 91.59 & 88.84 & 91.37 & 90.73 & 91.36 & 84.83 & 88.83 & 85.68 \\
Target & \cellcolor{mediumgreen}86.99 & \cellcolor{mediumgreen}93.23 & 91.85 & \cellcolor{mediumgreen}92.78 & \cellcolor{mediumgreen}92.50 & \cellcolor{mediumgreen}92.47 & \cellcolor{mediumgreen}88.56 & \cellcolor{mediumgreen}91.26 & \cellcolor{mediumgreen}90.07 \\
\bottomrule
\addlinespace[10pt]
\end{tabular}
\end{sc}
\caption{Layer-averaged Cosine similarity (expressed as percentages) between the target language (\textbf{Lithuanian}) and other languages for \textsc{Llama-3.1-8B}.}
\label{tab:cosine_similarity_lt}
\end{table*}

\begin{table*}[h!]
\centering
\small
\begin{sc}
\begin{tabular}{@{}lccccccccc@{}}
\toprule
 & ar & de & en & es & fr & it & ja & ru & zh \\
\midrule
Baseline & 82.73 & 88.36 & 85.37 & 87.88 & 87.58 & 88.25 & 83.37 & 86.37 & 83.76 \\
Full & \cellcolor{mediumgreen}86.70 & 89.06 & 90.55 & 90.20 & 89.57 & 88.98 & 87.45 & 84.64 & 89.21 \\
MLP & 80.97 & 88.03 & 88.93 & 86.05 & 86.14 & 86.56 & 82.48 & 84.67 & 86.19 \\
Adapter & 84.62 & 92.14 & 88.45 & 91.85 & 91.36 & 92.08 & 84.51 & 89.09 & 86.13 \\
Random & 82.07 & 91.14 & 87.65 & 90.65 & 90.18 & 90.39 & 83.74 & 87.63 & 84.94 \\
Target & 86.60 & \cellcolor{mediumgreen}93.48 & \cellcolor{mediumgreen}91.73 & \cellcolor{mediumgreen}93.19 & \cellcolor{mediumgreen}93.04 & \cellcolor{mediumgreen}93.14 & \cellcolor{mediumgreen}88.61 & \cellcolor{mediumgreen}90.79 & \cellcolor{mediumgreen}89.61 \\
\bottomrule
\addlinespace[10pt]
\end{tabular}
\end{sc}
\caption{Layer-averaged Cosine similarity (expressed as percentages) between the target language (\textbf{Latvian}) and other languages for \textsc{Llama-3.1-8B}.}
\label{tab:cosine_similarity_lv}
\end{table*}

\begin{table*}[h!]
\centering
\small
\begin{sc}
\begin{tabular}{@{}lccccccccc@{}}
\toprule
 & ar & de & en & es & fr & it & ja & ru & zh \\
\midrule
Baseline & 87.36 & 87.33 & 86.16 & 87.00 & 86.85 & 87.38 & 86.00 & 92.75 & 86.12 \\
Full & 85.37 & 89.55 & 90.10 & 86.95 & 89.35 & 87.79 & 84.21 & 92.15 & 87.69 \\
MLP & 84.71 & 88.09 & 90.33 & 87.73 & 87.95 & 88.06 & 86.25 & 90.71 & 88.38 \\
Adapter & 87.53 & 86.66 & 85.75 & 86.49 & 86.15 & 86.97 & 84.81 & 97.54 & 85.33 \\
Random & \cellcolor{mediumgreen}94.88 & 89.60 & 87.52 & 89.51 & 89.00 & 90.22 & 88.70 & 98.18 & 88.11 \\
Target & 92.74 & \cellcolor{mediumgreen}92.92 & \cellcolor{mediumgreen}92.01 & \cellcolor{mediumgreen}92.99 & \cellcolor{mediumgreen}92.63 & \cellcolor{mediumgreen}93.00 & \cellcolor{mediumgreen}91.52 & \cellcolor{mediumgreen}98.72 & \cellcolor{mediumgreen}92.00 \\
\bottomrule
\addlinespace[10pt]
\end{tabular}
\end{sc}
\caption{Layer-averaged Cosine similarity (expressed as percentages) between the target language (\textbf{Macedonian}) and other languages for \textsc{Llama-3.1-8B}.}
\label{tab:cosine_similarity_mk}
\end{table*}

\begin{table*}[h!]
\centering
\small
\begin{sc}
\begin{tabular}{@{}lccccccccc@{}}
\toprule
 & ar & de & en & es & fr & it & ja & ru & zh \\
\midrule
Baseline & 84.92 & 87.62 & 85.09 & 87.37 & 87.32 & 89.50 & 82.94 & 84.95 & 83.17 \\
Full & 86.03 & 90.98 & 91.14 & 89.55 & 89.71 & 90.60 & 85.43 & 88.60 & 85.49 \\
MLP & 86.29 & 90.72 & 90.80 & 90.55 & 89.50 & 90.07 & 86.65 & 88.56 & 87.35 \\
Adapter & 88.20 & 93.48 & 89.95 & \cellcolor{mediumgreen}93.85 & \cellcolor{mediumgreen}93.66 & \cellcolor{mediumgreen}96.17 & 85.00 & 88.92 & 85.74 \\
Random & 89.54 & 91.84 & 90.38 & 91.49 & 91.17 & 93.10 & 86.12 & 89.70 & 86.50 \\
Target & \cellcolor{mediumgreen}92.23 & \cellcolor{mediumgreen}93.53 & \cellcolor{mediumgreen}92.42 & 92.79 & 92.63 & 94.90 & \cellcolor{mediumgreen}89.06 & \cellcolor{mediumgreen}90.80 & \cellcolor{mediumgreen}89.52 \\
\bottomrule
\addlinespace[10pt]
\end{tabular}
\end{sc}
\caption{Layer-averaged Cosine similarity (expressed as percentages) between the target language (\textbf{Maltese}) and other languages for \textsc{Llama-3.1-8B}.}
\label{tab:cosine_similarity_mt}
\end{table*}

\begin{table*}[h!]
\centering
\small
\begin{sc}
\begin{tabular}{@{}lccccccccc@{}}
\toprule
 & ar & de & en & es & fr & it & ja & ru & zh \\
\midrule
Baseline & 85.43 & 84.39 & 82.60 & 83.91 & 83.85 & 84.81 & 85.77 & 86.33 & 84.49 \\
Full & 88.65 & 88.61 & \cellcolor{mediumgreen}89.58 & 89.29 & 89.63 & 89.65 & 88.12 & 89.33 & 89.39 \\
MLP & 83.47 & 83.90 & 86.27 & 84.17 & 84.67 & 83.77 & 85.09 & 85.00 & 86.68 \\
Adapter & 89.31 & 86.91 & 84.48 & 86.07 & 85.95 & 86.98 & 88.93 & 89.12 & 86.84 \\
Random & 92.35 & 87.82 & 84.56 & 87.06 & 86.96 & 87.95 & 90.76 & 91.25 & 88.24 \\
Target & \cellcolor{mediumgreen}94.93 & \cellcolor{mediumgreen}92.81 & 89.43 & \cellcolor{mediumgreen}91.40 & \cellcolor{mediumgreen}91.43 & \cellcolor{mediumgreen}92.10 & \cellcolor{mediumgreen}94.87 & \cellcolor{mediumgreen}94.62 & \cellcolor{mediumgreen}92.49 \\
\bottomrule
\addlinespace[10pt]
\end{tabular}
\end{sc}
\caption{Layer-averaged Cosine similarity (expressed as percentages) between the target language (\textbf{Nepali}) and other languages for \textsc{Llama-3.1-8B}.}
\label{tab:cosine_similarity_ne}
\end{table*}

\begin{table*}[h!]
\centering
\small
\begin{sc}
\begin{tabular}{@{}lccccccccc@{}}
\toprule
 & ar & de & en & es & fr & it & ja & ru & zh \\
\midrule
Baseline & 87.04 & 91.90 & 90.14 & 91.61 & 91.24 & 91.61 & 87.26 & 90.57 & 87.85 \\
Full & 87.16 & 92.87 & \cellcolor{mediumgreen}93.77 & 91.80 & 91.93 & 91.83 & 88.86 & 88.68 & 89.80 \\
MLP & 84.25 & 91.34 & 93.15 & 90.85 & 90.58 & 90.91 & 84.58 & 88.15 & 87.52 \\
Adapter & 85.71 & 92.33 & 90.44 & 92.23 & 91.83 & 92.12 & 86.20 & 90.18 & 87.07 \\
Random & 87.06 & 92.57 & 91.00 & 92.11 & 91.82 & 91.69 & 87.97 & 91.88 & 88.57 \\
Target & \cellcolor{mediumgreen}87.56 & \cellcolor{mediumgreen}93.88 & 93.35 & \cellcolor{mediumgreen}93.52 & \cellcolor{mediumgreen}93.25 & \cellcolor{mediumgreen}93.33 & \cellcolor{mediumgreen}89.86 & \cellcolor{mediumgreen}92.60 & \cellcolor{mediumgreen}90.45 \\
\bottomrule
\addlinespace[10pt]
\end{tabular}
\end{sc}
\caption{Layer-averaged Cosine similarity (expressed as percentages) between the target language (\textbf{Slovak}) and other languages for \textsc{Llama-3.1-8B}.}
\label{tab:cosine_similarity_sk}
\end{table*}

\begin{table*}[h!]
\centering
\small
\begin{sc}
\begin{tabular}{@{}lccccccccc@{}}
\toprule
 & ar & de & en & es & fr & it & ja & ru & zh \\
\midrule
Baseline & 85.88 & 91.34 & 89.69 & 91.13 & 90.40 & 91.31 & 86.13 & 89.15 & 86.98 \\
Full & 86.41 & 89.81 & 93.60 & 90.56 & 89.94 & 90.11 & 86.23 & 89.84 & 88.16 \\
MLP & 84.26 & 89.02 & 92.18 & 89.38 & 89.64 & 89.61 & 84.05 & 87.33 & 87.56 \\
Adapter & 86.40 & 92.67 & 91.10 & 92.64 & 92.03 & 92.60 & 86.11 & 90.97 & 87.25 \\
Random & 84.48 & 92.45 & 90.52 & 91.82 & 91.10 & 92.14 & 84.45 & 91.15 & 85.49 \\
Target & \cellcolor{mediumgreen}88.57 & \cellcolor{mediumgreen}94.18 & \cellcolor{mediumgreen}93.79 & \cellcolor{mediumgreen}94.25 & \cellcolor{mediumgreen}93.47 & \cellcolor{mediumgreen}94.17 & \cellcolor{mediumgreen}88.72 & \cellcolor{mediumgreen}93.27 & \cellcolor{mediumgreen}90.21 \\
\bottomrule
\addlinespace[10pt]
\end{tabular}
\end{sc}
\caption{Layer-averaged Cosine similarity (expressed as percentages) between the target language (\textbf{Slovenian}) and other languages for \textsc{Llama-3.1-8B}.}
\label{tab:cosine_similarity_sv}
\end{table*}
\clearpage



\end{document}